\documentclass{article}

\PassOptionsToPackage{numbers,sort&compress}{natbib}
\IfFileExists{neurips_2026.sty}{
  \usepackage[main, final]{neurips_2026}
}{\IfFileExists{Styles/neurips_2025.sty}{
  \usepackage[dandb]{neurips_2025}
}{
  \usepackage[margin=1in]{geometry}
  \usepackage[numbers]{natbib}
}}

\usepackage[utf8]{inputenc}
\usepackage[T1]{fontenc}
\usepackage{hyperref}
\usepackage{url}
\usepackage{booktabs}
\usepackage{amsfonts}
\usepackage{nicefrac}
\usepackage{microtype}
\usepackage{xcolor}
\usepackage{amsmath}
\usepackage{graphicx}
\usepackage{array}
\usepackage{rotating}
\usepackage{tabularx}
\usepackage{makecell}
\usepackage{multirow}
\usepackage{siunitx}
\usepackage{float}
\usepackage{wrapfig}
\usepackage{adjustbox}
\usepackage{pifont}
\usepackage{fontawesome5}
\newcommand{\cmark}{\ding{51}}
\newcommand{\xmark}{\ding{55}}

\newcommand{\bad}[1]{\textcolor{red!75!black}{#1}}
\newcommand{\good}[1]{\textbf{#1}}

\title{RadGenome-Anatomy: A Large-Scale Anatomy-Labeled Chest Radiograph Dataset via Physically Grounded Volumetric Projection}

\author{
Shuchang Ye$^{1}$ \quad Mingyuan Meng$^{2}$ \quad \textbf{Hao Wang}$^{3}$ \quad \textbf{Usman Naseem}$^{4}$ \quad \textbf{Jinman Kim}$^{1*}$\vspace{1em}
\\$^1$The University of Sydney   $^2$Zhongguancun Academy \\ $^3$Shanghai Jiao Tong University $^4$Macquarie University
}

\begin{document}

\maketitle

\begin{abstract}
Anatomical structure labels for chest radiographs are essential for medical image segmentation and a broad range of downstream diagnostic tasks. However, annotating anatomy directly on 2D chest radiographs is labor-intensive and intrinsically ambiguous, as 3D anatomical structures are projected onto a single 2D plane where boundaries may overlap, be occluded, or appear only partially visible. Consequently, existing anatomy-labeled chest radiograph datasets remain limited in scale, anatomy coverage, and label reliability. To address these limitations, we introduce \textit{\textbf{RadGenome-Anatomy}} [\href{https://huggingface.co/datasets/EvidenceAIResearch/radgenome-anatomy}{\faDatabase\ Database}], the largest anatomy-labeled chest radiograph dataset, containing over \textbf{10 million} segmentation masks across \textbf{210} anatomical structures in $25{,}692$ studies. It is constructed by projecting large-scale 3D anatomical masks from CT volumes into 2D radiographic space through canonical radiographic geometry. This shifts annotation from directly tracing uncertain 2D boundaries to defining anatomy in volumetric space, where structures that overlap or become partially invisible in radiographs remain spatially separable. As a result, each 2D mask represents the physically grounded projected footprint of a volumetrically defined structure. The scale and broad anatomical coverage of \textit{RadGenome-Anatomy}, including structures that are overlapping, partially visible, or difficult to delineate directly, enable research on geometric measurements as explicit evidence for chest radiograph interpretation. We demonstrate this by training \textit{XAnatomy} [\href{https://github.com/ShuchangYe-bib/XAnatomy}{\faGithub\ Code}] to predict structure-specific masks and derive clinically relevant measurements, achieving diagnostic accuracies of $96.4\%$, $95.6\%$, and $89.2\%$ for cardiomegaly, kyphosis, and scoliosis, respectively.
\end{abstract}

\section{Introduction}

Anatomical structure labels for chest radiographs provide essential supervision for computer-aided radiographic analysis, and have long supported the development of anatomy segmentation methods~\cite{study_on_public,paxray,volume_pseudo,chexmark,clavicle}. Beyond segmentation, such labels also provide structural evidence for downstream tasks, including anatomy-aware report generation and region grounding~\cite{rgrg}, anatomy-driven pathology analysis~\cite{anapath_detect}, and measurement extraction such as cardiothoracic ratio estimation~\cite{cardionet,ctr}.

\begin{figure}[t]
  \centering
  \includegraphics[width=0.95\linewidth]{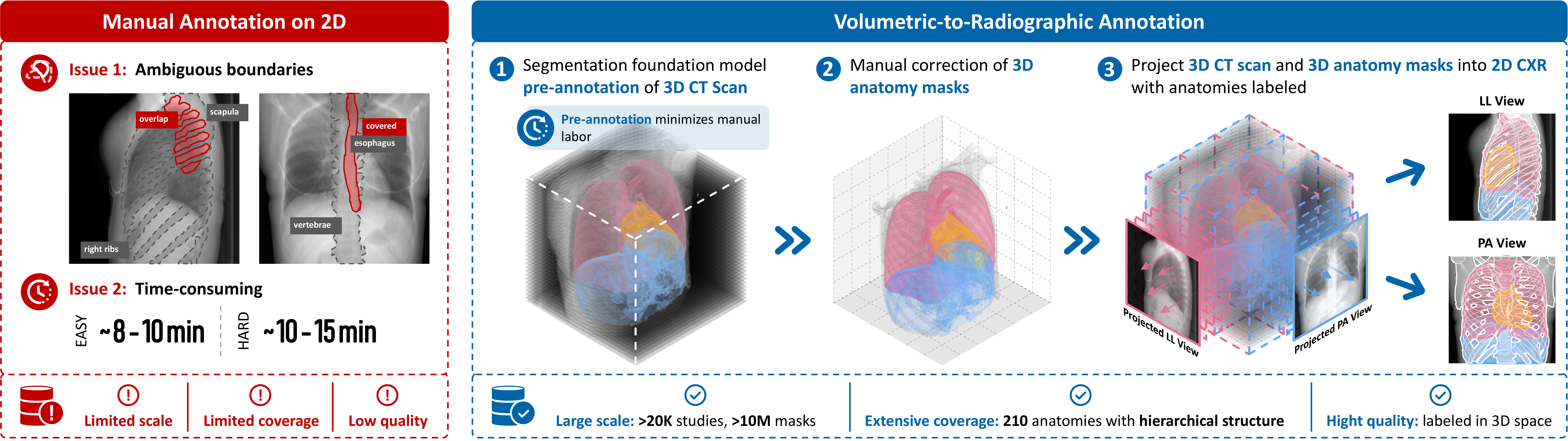}
  \caption{Comparison between conventional 2D manual annotation and our volumetric-to-radiographic annotation paradigm.}
  \label{fig:pipeline}
  \vspace{-10pt}
\end{figure}

The development of anatomy-aware chest-radiograph models is fundamentally constrained by the availability of high-quality anatomical labels. This bottleneck arises from limitations in both dataset scale and label quality, as illustrated in Figure~\ref{fig:pipeline} (left). First, existing anatomy-labeled chest-radiograph datasets remain limited in scale~\cite{study_on_public,jsrt,2datasets}. Fine-grained anatomical masks require expert delineation, and tracing them directly on chest radiographs is slow and labor-intensive. Prior work reports that radiologists require approximately 8--10 minutes for a standard case and 10--15 minutes for a difficult case~\cite{annotime}. This annotation cost makes it difficult to construct large-scale datasets with broad anatomical coverage. Consequently, publicly available expert-labeled chest radiograph datasets have relatively small sample size, and typically cover only a limited set of anatomical structures. Second, label quality is limited by the projection geometry of chest radiography. Because a chest radiograph compresses three-dimensional anatomy into a two-dimensional image, structures frequently overlap, and many boundaries are only partially visible or fully occluded. Annotators therefore cannot rely solely on directly visible image evidence and must often infer hidden anatomy, introducing subjectivity and inconsistency into the resulting labels.

To overcome these limitations, we introduce a volumetric-to-radiographic annotation paradigm for chest radiographs that derives 2D supervision from AI-assisted volumetric CT pre-annotations~\cite{monai_label} rather than direct manual tracing on ambiguous radiographic projections, as shown in Figure~\ref{fig:pipeline} (right). This paradigm addresses the two central limitations of manual radiographic annotation. Instead of requiring experts to trace each anatomical mask directly on chest radiographs, it constructs supervision from volumetric pre-segmentations~\cite{sat} generated by segmentation foundation models~\cite{sam} that only require review and refinement, thereby reducing annotation cost and enabling substantially larger-scale dataset creation. Instead of annotating structures directly on ambiguous 2D projections, where overlapping structures obscure underlying anatomical boundaries, it derives supervision from anatomically resolved 3D volumes in which organs and substructures remain spatially separated and can be delineated more consistently. Since chest radiographs are projections of volumetric anatomy, these 3D labels can be naturally transferred into radiographic views, producing supervision that is potentially more scalable and more anatomically faithful than direct manual annotation.

Building on this annotation paradigm, we reformulate chest-radiograph anatomy labeling as volumetric-to-radiographic supervision transfer under an attenuation-based forward model of radiographic image formation. Each volumetric scan is treated as a discretized attenuation field, from which postero-anterior (PA) and lateral (LL) views are generated by integrating voxel-wise density along the projection axis, consistent with the line-integral formulation of the Beer-Lambert law~\cite{beer} and the X-ray transform. The same operator is applied to the volumetric anatomy masks, ensuring that each projected label is geometrically coupled to image formation rather than drawn directly on an ambiguous 2D projection. The resulting supervision is both view-consistent and physically grounded, with each 2D mask corresponding to the radiographic footprint of a 3D anatomical structure under the same imaging geometry.

In this work, we introduce \textit{\textbf{RadGenome-Anatomy}}, a large-scale anatomy-labeled chest radiograph dataset containing over $10$ million masks spanning $210$ anatomical structures across standard PA and LL radiographs. Our main contributions are as follows.

\begin{itemize}
    \item We introduce RadGenome-Anatomy, the largest anatomy-labeled chest radiograph dataset, substantially expanding the scale of labeled radiographs, the granularity of anatomy coverage and the quality of anatomy labels.
    \item We present a physics-grounded volumetric-to-radiographic annotation paradigm that derives 2D anatomy labels from 3D CT segmentation masks through an attenuation-based forward projection model, replacing ambiguity-prone manual tracing on chest radiographs with anatomically faithful supervision transfer.
    \item We establish a benchmark for anatomy segmentation in chest radiograph, evaluating 19 representative models spanning CNN-, Transformer-, Mamba-, KAN-, and SAM-based architectures.
    \item We demonstrate that \textit{RadGenome-Anatomy} supports measurement-based diagnostic evidence beyond dense segmentation. \textit{XAnatomy}, trained on \textit{RadGenome-Anatomy}, derives geometric measurements from predicted anatomy masks to identify cardiomegaly, kyphosis, and scoliosis, with accuracies of $0.964$, $0.956$, and $0.892$, respectively.
\end{itemize}

\section{Related Work}

\begin{wraptable}[7]{r}{0.5\textwidth}
  \vspace{-40pt}
  \centering
  \caption{Comparison with existing chest radiograph anatomy resources.}
  \label{tab:dataset-comparison}
  \vspace{1mm}
  \scriptsize
  \setlength{\tabcolsep}{3pt}
  \renewcommand{\arraystretch}{0.92}
  \begin{tabularx}{\linewidth}{@{}>{\raggedright\arraybackslash}Xccrrr@{}}
    \toprule
    Dataset & PA & LL & \#Anat. & \#Masks & \#Studies \\
    \midrule
    SCR-JSRT~\cite{study_on_public,jsrt}
    & \cmark & \bad{\xmark}
    & \bad{3}
    & \bad{247}
    & \bad{247} \\
    
    Montgomery~\cite{2datasets}
    & \cmark & \bad{\xmark}
    & \bad{1}
    & \bad{138}
    & \bad{138} \\
    
    VinDr-RibCXR~\cite{vindr}
    & \cmark & \bad{\xmark}
    & \bad{20}
    & \bad{4{,}900}
    & \bad{245} \\
    
    CheXmask~\cite{chexmask} 
    & \cmark & \bad{\xmark}
    & \bad{3}
    & 657{,}566
    & 657{,}566 \\
    
    PAX-Ray++~\cite{volume_pseudo}
    & \cmark & \cmark
    & 157
    & $>$2M
    & 7{,}377 \\
    
    \textbf{RadGenome-Anatomy}
    & \cmark & \cmark
    & \good{210}
    & \good{10{,}790{,}646}
    & \good{25{,}692} \\
    \bottomrule
  \end{tabularx}
  \vspace{-1.0em}
\end{wraptable}

\subsection{Anatomy-labeled Radiograph Datasets}

Chest radiograph anatomy benchmarks remain in small size and limited anatomy coverage because direct 2D annotation is both labour-intensive and ambiguous. SCR-JSRT provides expert masks for only the lungs, heart and clavicles in $247$ PA radiographs~\cite{study_on_public,jsrt}, while Montgomery and Shenzhen provide limited or no comprehensive anatomical segmentation~\cite{2datasets}. In radiographs, projected structures overlap, occlude one another and often have weak boundaries, forcing experts to trace uncertain anatomy manually. 

Recent segmentation foundation models reduce this burden by shifting annotation from manual delineation to model-assisted review and refinement~\cite{sam,sat}. RadGenome-Chest CT applies this strategy to CT-RATE, producing volumetric anatomical labels with substantially less manual drawing than fully expert-traced pipelines~\cite{radgenome-ct,ct-rate}. Volumetric anatomy also resolves the main ambiguity of chest radiograph annotation: structures are defined in 3D, where boundaries are anatomically separable, and then projected into radiographic views as geometry-consistent 2D masks. Although prior work demonstrated this principle at limited scale~\cite{volume_pseudo}, anatomical coverage remained restricted. In this study, we scale this paradigm to \textit{RadGenome-Anatomy}, a chest radiograph anatomy dataset with over 10 million projected masks covering 210 anatomical structures.

\subsection{Anatomy masks for Measurement-based Diagnosis}

Existing AI-assisted chest-radiograph diagnosis~\cite{medgemma,llava-med} is largely built on label-based, anatomy-implicit prediction. Even though recent medical vision-language models can generate explanations or reasoning traces, the evidences are usually expressed as free-text rationales or implicit image-level cues~\cite{med-r1,huatuogpt-o1,maira}. However, in clinical application, the diagnosis is based on explicit anatomical geometry measurements: Cardiomegaly is assessed using the cardiothoracic ratio, while scoliosis and thoracic kyphosis are quantified using spinal curvature measurements such as Cobb-style angles. The gap between generated explanations and clinically grounded evidence creates an opportunity for measurement-based AI-aided diagnosis. \textit{RadGenome-Anatomy} addresses this gap by providing large-scale fine-grained anatomy labels that make the geometry of chest radiographs explicit. We demonstrate that these labels move beyond segmentation supervision by providing diagnostic evidence through named anatomical measurements for structural abnormalities.

\section{RadGenome-Anatomy}
\label{sec:radgenome-anatomy}

\begin{wraptable}[11]{r}{0.5\textwidth}
  \vspace{-60pt}
  \centering
  \caption{Dataset statistics for \emph{RadGenome-Anatomy}.}
  \label{tab:dataset-stats}
  \vspace{2mm}
  \scriptsize
  \setlength{\tabcolsep}{3pt}
  \renewcommand{\arraystretch}{1.05}
  \begin{tabularx}{\linewidth}{@{}>{\raggedright\arraybackslash}Xrrr@{}}
    \toprule
    & Train & Val. & Total \\
    \midrule
    \multicolumn{4}{@{}l}{\textit{Study-level inventory}} \\
    Studies              & 24{,}128       & 1{,}564   & 25{,}692 \\
    PA projections       & 24{,}129       & 1{,}564   & 25{,}693 \\
    LL projections       & 24{,}129       & 1{,}564   & 25{,}693 \\
    \midrule
    \multicolumn{4}{@{}l}{\textit{Anatomy masks, 210 classes per view}} \\
    PA                   & 5{,}066{,}885  & 328{,}438 & 5{,}395{,}323 \\
    LL                   & 5{,}066{,}885  & 328{,}438 & 5{,}395{,}323 \\
    Total                & 10{,}133{,}770 & 656{,}876 & 10{,}790{,}646 \\
    \midrule
    \multicolumn{4}{@{}l}{\textit{Region masks, 10 classes per view}} \\
    PA                   & 241{,}290      & 15{,}640  & 256{,}930 \\
    LL                   & 241{,}290      & 15{,}640  & 256{,}930 \\
    Total                & 482{,}580      & 31{,}280  & 513{,}860 \\
    \bottomrule
  \end{tabularx}
  \vspace{-1.0em}
\end{wraptable}

\subsection{Overview}
\label{sec:radgenome-overview}

\textbf{\emph{RadGenome-Anatomy}} is a large-scale anatomy-labeled chest radiograph dataset constructed from the RadGenome-Chest CT~\cite{radgenome-ct} corpus. As summarized in Table~\ref{tab:dataset-stats}, the dataset contains $25{,}692$ volumetric studies, with $24{,}128$ studies for training and $1{,}564$ for validation. These studies yield $25{,}693$ posteroanterior (PA) projection images and $25{,}693$ lateral (LL) projection images at $384 \times 384$ resolution. Across the two radiographic views, the dataset provides $10{,}790{,}646$ fine-grained anatomy masks over $210$ canonical anatomy classes and $513{,}860$ region masks over $10$ anatomical groups. The training split contains $10{,}133{,}770$ anatomy masks and $482{,}580$ region masks, for $10{,}616{,}350$ mask instances, while the validation split contains $656{,}876$ anatomy masks and $31{,}280$ region masks, for $688{,}156$ mask instances.

\subsection{Data Statistics}
\label{sec:radgenome-statistics}

\begin{figure}[t]
  \centering
  \includegraphics[width=\linewidth]{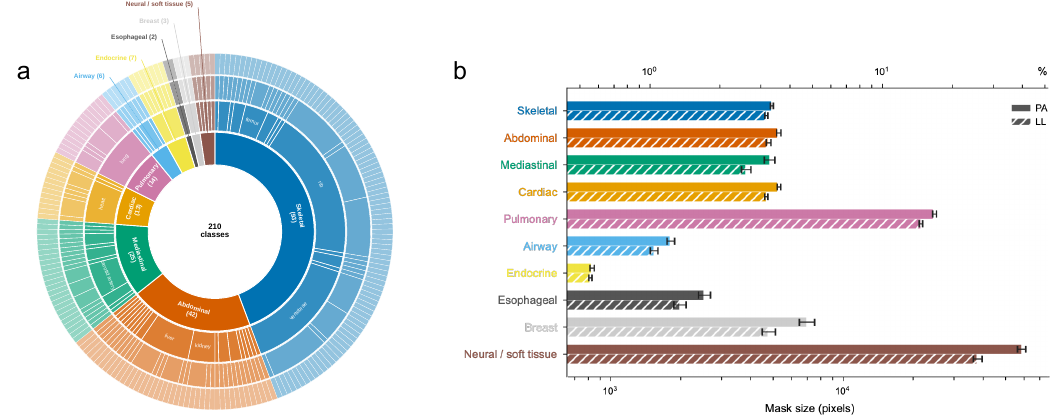}
  \caption{\textbf{Hierarchical label space and anatomy-mask scale in \emph{RadGenome-Anatomy}.} a. Four-level sunburst of the $210$ canonical anatomy classes. The innermost ring shows the 10 body-system groups, while the outer rings progressively expand the hierarchy into organ families, anatomical sub-parts, and the final canonical classes (Detailed label hierarchy is provided in Figure~\ref{fig:label_details}.). b. Anatomy-mask area for each body-system group. Error bars indicate $95\%$ bootstrap CIs.}
  \label{fig:label_mask}
\end{figure}

\paragraph{Label hierarchy} Anatomy defines 210 canonical anatomical classes organized into a four-level hierarchy: \textit{body system} $\rightarrow$ \textit{organ} $\rightarrow$ \textit{substructure} $\rightarrow$ \textit{canonical label} (Fig.~\ref{fig:label_mask}a). Rather than treating anatomical structures as a flat category set, the hierarchy preserves their coarse-to-fine relationships and provides a unified label space across multiple anatomical scales. At the system level, the taxonomy covers 10 major groups: \emph{Skeletal}, \emph{Abdominal}, \emph{Mediastinal}, \emph{Cardiac}, \emph{Pulmonary}, \emph{Airway}, \emph{Endocrine}, \emph{Esophageal}, \emph{Breast}, and \emph{Neural / soft tissue}. These groups are further decomposed into organ- and substructure-level entities, including vertebral levels, ribs, lung lobes, laterality-specific organs, mediastinal vessels, and cardiac chambers. The resulting class distribution is highly non-uniform, reflecting the anatomical granularity available within each system. The largest branches are \emph{Skeletal} with 93 classes, \emph{Abdominal} with 42 classes, and \emph{Mediastinal} with 25 classes, corresponding to detailed coverage of osseous structures, upper-abdominal organs, and thoracic mediastinal anatomy. Smaller branches, including \emph{Esophageal} with 2 classes, \emph{Breast} with 3 classes, and \emph{Neural / soft tissue} with 5 classes, extend the taxonomy beyond dominant thoracic organs to additional structures visible or clinically relevant in chest radiographs. Details of the anatomical label taxonomy, including the mapping from body system to organ, sub-part, and leaf label, are shown in Figure.~\ref{fig:label_details}. Overall, the label space combines broad anatomical coverage with fine-grained structural specificity.

\paragraph{Anatomy Masks} Each anatomy mask in \emph{RadGenome-Anatomy} is a single-channel binary image co-registered with radiographic projection. We quantify its spatial footprint by the number of foreground pixels and summarize the per-group mean footprint for the ten body-system groups in Figure~\ref{fig:label_mask}b and Table~\ref{tab:mask_details}, separately for PA and LL views with $95\%$ bootstrap confidence intervals. Mean mask sizes vary by nearly two orders of magnitude, from small structures such as \emph{Endocrine} masks ($\sim\!8 \times 10^2$ pixels; $\approx\!0.6\%$ of the image) to large \emph{Neural~/~soft tissue} masks ($\sim\!5.9 \times 10^4$ pixels; $\approx\!40\%$). Large-footprint groups are dominated by spatially extensive labels such as skin, muscle, lungs, and lobes, whereas \emph{Airway} and \emph{Endocrine} contain thin tubular or small glandular structures and therefore have substantially smaller masks. PA and LL footprints are broadly consistent across groups: the PA/LL mean ratio ranges from $1.02$ to $1.55$, and remains below $1.3$ for eight of ten groups. The main view asymmetries occur for broad structures whose PA extent exceeds their LL silhouette, including \emph{Neural~/~soft tissue}, \emph{Pulmonary}, and \emph{Breast}. 

\subsection{Volumetric-to-Radiographic Projection}
\label{sec:projection}

Let \(V \in \mathbb{R}^{H \times W \times D}\) denote a CT volume, where \(V[i,j,k]\) is the voxel intensity at spatial index \((i,j,k)\), and let \((s_x, s_y, s_z)\) denote the voxel spacing along the three axes. For each anatomical structure \(c \in \{1,\dots,C\}\), where \(C=210\) in our dataset, let \(M^{(c)} \in \{0,1\}^{H \times W \times D}\) denote its binary 3D mask, with \(M^{(c)}[i,j,k]=1\) indicating that voxel \((i,j,k)\) belongs to anatomy \(c\). 

\begin{figure}[t]
    \centering
    \includegraphics[width=\linewidth]{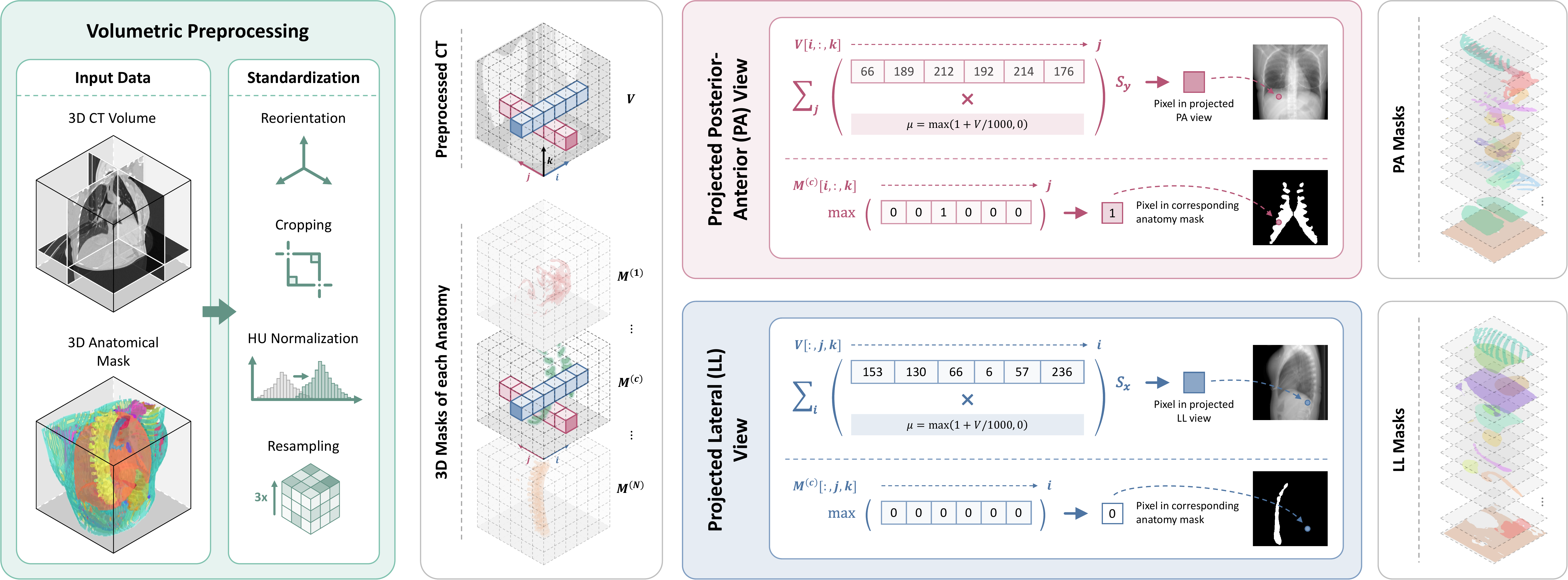}
    \vspace{-10pt}
    \caption{Overview of the volumetric-to-radiographic projection pipeline. A preprocessed 3D CT volume \(V\) and anatomy-specific 3D masks \(\{M^{(c)}\}_{c=1}^{C}\) are projected into postero-anterior (PA) and lateral (LL) views.}
    \label{fig:projection_pipeline}
    \vspace{-10pt}
\end{figure}

Figure~\ref{fig:projection_pipeline} summarizes the overall pipeline. Starting from a preprocessed CT volume and the set of anatomy-specific 3D masks, we construct view-specific 2D image projections by accumulating attenuation-like voxel values along the projection axis, and construct anatomy-specific 2D mask projections by ray-wise occupancy. We further investigate the radiographic transferability of CT-derived chest radiographs and the quality of projected anatomy masks through zero-shot external validation on expert-labeled real chest-radiograph benchmarks in Section~\ref{sec:real_xray_transfer_label_quality}.

\paragraph{Image projection.} X-ray image formation is classically described by the Beer--Lambert law, under which the transmitted intensity along a ray decreases exponentially with the line integral of the linear attenuation coefficient:
\begin{equation}
I = I_0 \exp\!\left(- \int \mu(l)\, dl \right), \qquad -\log\!\left(\frac{I}{I_0}\right) = \int \mu(l)\, dl.
\end{equation}
Motivated by this line-integral form, we approximate radiographic image formation by discrete accumulation of attenuation-like voxel values along the projection axis. Specifically, we convert CT intensities to nonnegative attenuation-like values by
\begin{equation}
\tilde{\mu}[i,j,k] = \max\!\left(1 + \frac{V[i,j,k]}{1000},\, 0\right).
\label{eq:mu_transform}
\end{equation}
The raw PA and LL projections are then defined as
\begin{equation}
P^{\mathrm{PA}}_{\mathrm{raw}}[i,k] = \left(\sum_{j} \tilde{\mu}[i,j,k]\right) s_y, \qquad
P^{\mathrm{LL}}_{\mathrm{raw}}[j,k] = \left(\sum_{i} \tilde{\mu}[i,j,k]\right) s_x.
\label{eq:image_projection}
\end{equation}
Thus, each projected image pixel corresponds to the discrete line integral of attenuation-like values along one viewing ray.

\paragraph{Anatomy-mask projection.} Projection is performed independently for each anatomy-specific mask \(M^{(c)}\). For a given ray, the projected mask pixel is activated if that ray intersects the target structure at any depth. Accordingly, the raw PA and LL mask projections are
\begin{equation}
A^{(c),\mathrm{PA}}_{\mathrm{raw}}[i,k] = \max_{j} M^{(c)}[i,j,k], \qquad
A^{(c),\mathrm{LL}}_{\mathrm{raw}}[j,k] = \max_{i} M^{(c)}[i,j,k].
\label{eq:mask_projection}
\end{equation}
Because \(M^{(c)}\) is binary, the max operator is equivalent to testing whether any voxel of anatomy \(c\) lies along the corresponding ray, yielding an anatomy-specific radiographic footprint under the chosen projection geometry.

\paragraph{Image-plane resampling and output generation.}
Each projected image is resampled according to the physical spacing of the retained image-plane axes to compensate for anisotropic CT voxel spacing. Grayscale projections are resampled with bilinear interpolation, whereas anatomy masks are resampled with nearest-neighbor interpolation and re-binarized to preserve discrete labels. The outputs are then mapped to a canonical view-specific orientation by flips and rotations. Finally, each grayscale projection is normalized image-wise to the 8-bit range:
\begin{equation}
P^{v}[u,v'] = 255 \, \frac{P^{v}_{\mathrm{proc}}[u,v'] - P^{v}_{\min}}{P^{v}_{\max} - P^{v}_{\min}},
\qquad v \in \{\mathrm{PA}, \mathrm{LL}\},
\label{eq:image_normalization}
\end{equation}
where \(P^{v}_{\mathrm{proc}}\) denotes the processed projection and \(P^{v}_{\min}, P^{v}_{\max}\) are its image-wise minimum and maximum intensities. Anatomy masks are stored as binary 8-bit maps after nearest-neighbor resampling.

\begin{table}[t]
  \centering
  \scriptsize
  \caption{Overall benchmark of 19 segmentation models on \textit{RadGenome-Anatomy}. Each cell is reported as mean $\pm$ 95\% CI half-width from bootstrap resampling over classes. $\uparrow$ indicates higher is better, and $\downarrow$ indicates lower is better. Best results are shown in bold.}
  \label{tab:benchmark}
  \setlength{\tabcolsep}{2.2pt}
  \resizebox{\linewidth}{!}{%
  \begin{tabular}{@{}llrrrrrrrr@{}}
    \toprule
    Family & Model & Dice ($\uparrow$) & IoU ($\uparrow$) & HD95 ($\downarrow$) & ASD ($\downarrow$) & NSD ($\uparrow$) & Det. P ($\uparrow$) & Det. R ($\uparrow$) & Det. F1 ($\uparrow$) \\
    \midrule
    \multirow{7}{*}{CNN-based} & U-Net & $58.01 \pm 4.01$ & $48.00 \pm 3.66$ & $15.49 \pm 3.05$ & $3.79 \pm 0.89$ & $55.46 \pm 2.15$ & $73.66 \pm 2.88$ & $1.63 \pm 0.17$ & $3.15 \pm 0.32$ \\
     & DeepLabV3+ & $60.14 \pm 3.91$ & $50.07 \pm 3.55$ & $15.34 \pm 2.96$ & $3.86 \pm 0.84$ & $56.34 \pm 2.14$ & $78.65 \pm 2.72$ & $1.72 \pm 0.17$ & $3.34 \pm 0.32$ \\
     & U-Net++ & $58.96 \pm 3.96$ & $49.07 \pm 3.61$ & $14.63 \pm 2.82$ & $3.56 \pm 0.82$ & $57.29 \pm 2.21$ & $73.21 \pm 2.69$ & $1.66 \pm 0.16$ & $3.22 \pm 0.31$ \\
     & PointRend & $58.03 \pm 3.85$ & $47.81 \pm 3.53$ & $16.13 \pm 2.89$ & $4.37 \pm 1.00$ & $55.38 \pm 2.07$ & $76.91 \pm 2.56$ & $1.64 \pm 0.15$ & $3.19 \pm 0.30$ \\
     & nnU-Net & $58.80 \pm 3.85$ & $48.81 \pm 3.57$ & $15.58 \pm 2.99$ & $4.36 \pm 1.27$ & $56.13 \pm 2.27$ & $60.66 \pm 3.10$ & $1.70 \pm 0.17$ & $3.27 \pm 0.31$ \\
     & UNeXt & $53.43 \pm 3.99$ & $43.44 \pm 3.69$ & $16.84 \pm 2.95$ & $3.69 \pm 0.68$ & $51.93 \pm 2.23$ & $79.03 \pm 2.86$ & $1.44 \pm 0.15$ & $2.81 \pm 0.28$ \\
     & PIDNet & $59.94 \pm 3.90$ & $49.88 \pm 3.46$ & $14.77 \pm 2.65$ & $3.68 \pm 0.75$ & $56.38 \pm 2.12$ & $77.84 \pm 2.91$ & $1.69 \pm 0.16$ & $3.29 \pm 0.31$ \\
    \midrule
    \multirow{7}{*}{Transformer-based} & UPerNet & $61.91 \pm 3.71$ & $51.74 \pm 3.35$ & $\mathbf{14.07 \pm 2.57}$ & $3.65 \pm 0.77$ & $58.37 \pm 1.94$ & $80.31 \pm 2.30$ & $\mathbf{1.83 \pm 0.18}$ & $\mathbf{3.56 \pm 0.34}$ \\
     & TransUNet & $57.63 \pm 3.92$ & $47.70 \pm 3.69$ & $15.70 \pm 3.06$ & $3.57 \pm 0.78$ & $55.18 \pm 2.41$ & $74.92 \pm 2.85$ & $1.60 \pm 0.16$ & $3.10 \pm 0.30$ \\
     & SegFormer & $\mathbf{61.98 \pm 3.76}$ & $\mathbf{51.99 \pm 3.39}$ & $14.24 \pm 2.61$ & $3.41 \pm 0.67$ & $58.52 \pm 2.03$ & $\mathbf{81.96 \pm 2.28}$ & $1.76 \pm 0.16$ & $3.41 \pm 0.30$ \\
     & Swin-UNet & $58.42 \pm 3.84$ & $48.41 \pm 3.52$ & $15.87 \pm 2.99$ & $4.01 \pm 0.84$ & $54.96 \pm 2.26$ & $43.81 \pm 2.80$ & $1.73 \pm 0.16$ & $3.29 \pm 0.30$ \\
     & TopFormer & $55.66 \pm 3.93$ & $45.55 \pm 3.48$ & $16.17 \pm 2.96$ & $3.70 \pm 0.72$ & $53.30 \pm 1.96$ & $78.25 \pm 2.53$ & $1.53 \pm 0.15$ & $2.98 \pm 0.30$ \\
     & Mask2Former & $61.75 \pm 3.85$ & $51.74 \pm 3.41$ & $14.29 \pm 2.62$ & $3.45 \pm 0.72$ & $\mathbf{58.86 \pm 1.96}$ & $81.38 \pm 2.21$ & $1.75 \pm 0.17$ & $3.40 \pm 0.30$ \\
     & SeaFormer & $48.93 \pm 3.98$ & $39.10 \pm 3.53$ & $20.48 \pm 3.42$ & $5.27 \pm 1.44$ & $47.45 \pm 2.27$ & $71.56 \pm 3.57$ & $1.36 \pm 0.14$ & $2.65 \pm 0.27$ \\
    \midrule
    \multirow{2}{*}{Mamba-based} & VMUNet & $58.08 \pm 3.90$ & $48.13 \pm 3.61$ & $15.88 \pm 3.01$ & $3.92 \pm 0.87$ & $55.43 \pm 2.35$ & $64.54 \pm 3.12$ & $1.65 \pm 0.16$ & $3.19 \pm 0.31$ \\
     & U-Mamba & $57.91 \pm 3.95$ & $47.93 \pm 3.42$ & $15.65 \pm 2.86$ & $3.58 \pm 0.73$ & $55.62 \pm 2.31$ & $64.60 \pm 2.92$ & $1.65 \pm 0.16$ & $3.19 \pm 0.31$ \\
    \midrule
    \multirow{1}{*}{KAN-based} & U-KAN & $57.57 \pm 4.01$ & $47.56 \pm 3.42$ & $15.42 \pm 3.00$ & $3.51 \pm 0.73$ & $55.50 \pm 2.05$ & $79.53 \pm 2.61$ & $1.58 \pm 0.16$ & $3.08 \pm 0.29$ \\
    \midrule
    \multirow{2}{*}{SAM-based} & Med-SAM-A & $56.03 \pm 3.91$ & $46.22 \pm 3.64$ & $15.09 \pm 3.07$ & $\mathbf{3.36 \pm 0.74}$ & $56.54 \pm 2.12$ & $37.00 \pm 3.34$ & $1.69 \pm 0.16$ & $3.15 \pm 0.30$ \\
     & S-SAM & $58.07 \pm 3.85$ & $48.13 \pm 3.53$ & $16.88 \pm 3.67$ & $4.87 \pm 2.09$ & $55.57 \pm 2.33$ & $64.24 \pm 3.06$ & $1.65 \pm 0.17$ & $3.18 \pm 0.30$ \\
    \bottomrule
  \end{tabular}%
  }
\end{table}

\begin{table}[t]
  \centering
  \scriptsize
  \caption{The performance of best evaluation-selected models under each hierarchy, view, and anatomical-group partition.}
  \label{tab:multigranularity}
  \setlength{\tabcolsep}{2.2pt}
  \resizebox{\linewidth}{!}{%
  \begin{tabular}{@{}llrrrrrrrr@{}}
    \toprule
    Partition & Model & Dice ($\uparrow$) & IoU ($\uparrow$) & HD95 ($\downarrow$) & ASD ($\downarrow$) & NSD ($\uparrow$) & Det. P ($\uparrow$) & Det. R ($\uparrow$) & Det. F1 ($\uparrow$) \\
    \midrule
    \multicolumn{10}{l}{\textit{Hierarchy}} \\
    \cmidrule(lr){1-10}
    L1 & SegFormer & $68.13 \pm 5.36$ & $57.95 \pm 5.10$ & $13.81 \pm 4.57$ & $3.00 \pm 1.00$ & $60.44 \pm 2.92$ & $83.47 \pm 3.09$ & $1.50 \pm 0.20$ & $2.94 \pm 0.38$ \\
    L2 & SegFormer & $61.61 \pm 8.13$ & $52.56 \pm 7.57$ & $10.18 \pm 3.03$ & $2.37 \pm 0.75$ & $60.14 \pm 3.81$ & $84.15 \pm 5.32$ & $1.96 \pm 0.30$ & $3.81 \pm 0.57$ \\
    L3 & SegFormer & $62.69 \pm 5.24$ & $51.26 \pm 4.76$ & $11.28 \pm 2.49$ & $2.68 \pm 0.74$ & $59.76 \pm 2.24$ & $82.57 \pm 3.29$ & $1.97 \pm 0.32$ & $3.81 \pm 0.59$ \\
    Pathology & UPerNet & $14.86 \pm 8.58$ & $10.36 \pm 6.30$ & $59.38 \pm 19.85$ & $19.16 \pm 6.09$ & $26.55 \pm 8.08$ & $49.89 \pm 10.63$ & $1.13 \pm 0.76$ & $2.19 \pm 1.45$ \\
    \midrule
    \multicolumn{10}{l}{\textit{View}} \\
    \cmidrule(lr){1-10}
    PA & SegFormer & $62.85 \pm 3.82$ & $52.97 \pm 3.54$ & $15.00 \pm 2.81$ & $3.69 \pm 0.79$ & $59.16 \pm 2.13$ & $82.62 \pm 2.28$ & $1.83 \pm 0.17$ & $3.54 \pm 0.32$ \\
    LL & SegFormer & $61.10 \pm 3.88$ & $51.01 \pm 3.59$ & $13.31 \pm 2.52$ & $3.10 \pm 0.63$ & $58.03 \pm 1.89$ & $81.86 \pm 2.43$ & $1.70 \pm 0.16$ & $3.31 \pm 0.31$ \\
    \midrule
    \multicolumn{10}{l}{\textit{Anatomical group}} \\
    \cmidrule(lr){1-10}
    Lung & SegFormer & $71.08 \pm 14.46$ & $64.06 \pm 13.30$ & $14.28 \pm 9.71$ & $4.06 \pm 3.05$ & $55.06 \pm 7.97$ & $80.53 \pm 9.07$ & $0.90 \pm 0.24$ & $1.77 \pm 0.46$ \\
    Cardiac & SegFormer & $84.81 \pm 4.40$ & $75.16 \pm 5.90$ & $4.32 \pm 2.41$ & $0.74 \pm 0.37$ & $65.71 \pm 2.67$ & $94.59 \pm 4.05$ & $1.71 \pm 0.42$ & $3.35 \pm 0.81$ \\
    Vascular & UPerNet & $45.66 \pm 10.27$ & $35.45 \pm 8.64$ & $16.01 \pm 7.73$ & $4.47 \pm 2.06$ & $52.39 \pm 5.84$ & $78.83 \pm 7.31$ & $2.15 \pm 0.44$ & $4.16 \pm 0.85$ \\
    Bone & SegFormer & $66.90 \pm 4.39$ & $55.70 \pm 4.13$ & $11.57 \pm 2.75$ & $2.47 \pm 0.69$ & $63.87 \pm 1.84$ & $81.77 \pm 3.21$ & $1.86 \pm 0.24$ & $3.61 \pm 0.45$ \\
    Mediastinum & SegFormer & $62.54 \pm 14.18$ & $50.67 \pm 12.71$ & $6.66 \pm 3.13$ & $1.41 \pm 0.80$ & $61.22 \pm 6.85$ & $86.85 \pm 8.42$ & $3.05 \pm 1.03$ & $5.85 \pm 1.95$ \\
    Abdomen & SegFormer & $50.81 \pm 9.31$ & $41.22 \pm 8.38$ & $26.63 \pm 9.52$ & $6.41 \pm 2.49$ & $47.24 \pm 5.52$ & $80.39 \pm 5.40$ & $1.65 \pm 0.37$ & $3.22 \pm 0.72$ \\
    Other & Mask2Former & $57.85 \pm 18.44$ & $51.29 \pm 17.85$ & $14.20 \pm 8.21$ & $3.54 \pm 2.08$ & $59.47 \pm 9.13$ & $78.35 \pm 7.49$ & $1.19 \pm 0.33$ & $2.33 \pm 0.63$ \\
    \bottomrule
  \end{tabular}%
  }
\end{table}

\section{Image Segmentation Benchmark Performances}
\label{sec:radgenome_anatomy_benchmark}

\subsection{Experimental Setup}

\paragraph{Models} We evaluate \textit{RadGenome-Anatomy} with 19 representative segmentation models spanning five architecture families: CNN, Transformer, Mamba, Kan and SAM. The CNN-based group includes U-Net~\cite{unet}, DeepLabV3+~\cite{deeplabv3p}, U-Net++~\cite{unetpp}, PointRend~\cite{pointrend}, nnU-Net~\cite{nnunet}, UNeXt~\cite{unext}, and PIDNet~\cite{pidnet}, covering classical encoder--decoder segmentation, atrous spatial context modeling, point-based boundary refinement, self-configuring medical segmentation, and lightweight real-time designs. The Transformer-based group includes UPerNet~\cite{upernet}, TransUNet~\cite{transunet}, SegFormer~\cite{segformer}, Swin-UNet~\cite{swinunet}, TopFormer~\cite{topformer}, Mask2Former~\cite{mask2former}, and SeaFormer~\cite{seaformer}, covering pyramid parsing, hybrid convolution--Transformer encoding, hierarchical attention, windowed self-attention, token-efficient segmentation, mask-query prediction, and mobile-efficient attention. We further include VMUNet~\cite{vm-unet} and U-Mamba~\cite{umamba} as Mamba-based models, U-KAN~\cite{ukan} as a KAN-based model, and MedSAM-Adaptor~\cite{medsam_adapter} and S-SAM~\cite{s-sam} as SAM-based models. 

\paragraph{Evaluation Metrics} We comprehensively evaluate segmentation quality using three main categories of metrics: region overlap, boundary fidelity, and component-level detection. For region overlap, we report Dice~\cite{dice} and IoU~\cite{miou}. For boundary fidelity, we report HD95~\cite{hd95}, ASD~\cite{asd}, and NSD~\cite{nsd}, where HD95 and ASD measure boundary distance and NSD measures the fraction of boundary points within an accepted tolerance. For component-level detection, we report connected-component precision, recall, and F1 to assess whether predicted masks recover small, repeated, or disconnected anatomical structures. 

\begin{figure}[t]
  \centering
  \includegraphics[width=\linewidth]{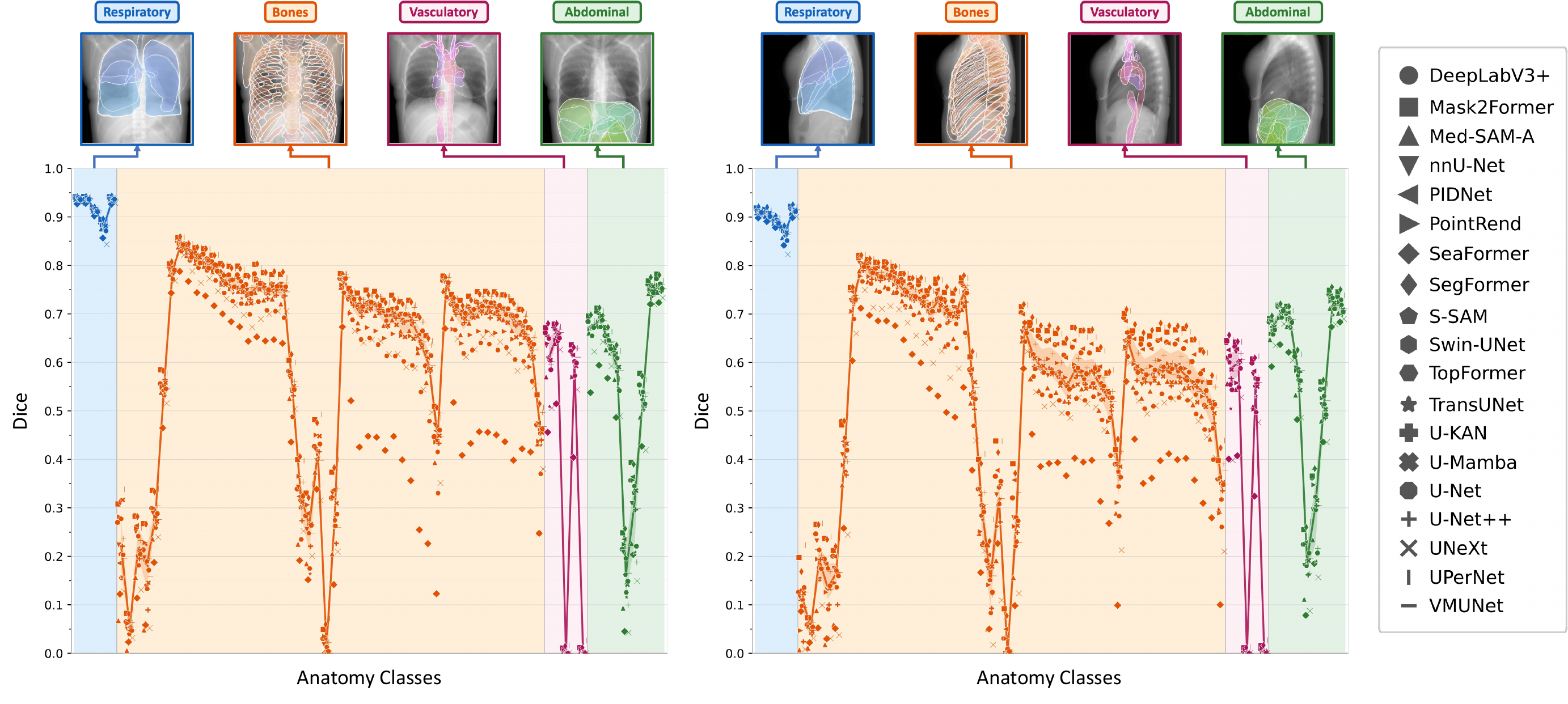}
  \vspace{-18pt}
  \caption{Per-class Dice performance for selected L3 anatomical classes in PA and LL views (see Table~\ref{tab:hierarchical-dice-detail} for exact values). The left panel reports the PA view, and the right panel reports the LL view. In each panel, the x-axis indexes anatomical classes, grouped by anatomical system, and the y-axis shows Dice score. For each class, points denote model performance values, the central curve shows the mean Dice across models, and the upper and lower curves show mean $\pm$ one standard deviation.}
  \label{fig:classes_performance}
  \vspace{-5pt}
\end{figure}

\subsection{Results and Discussion}
\label{sec:benchmark_results_discussion}

\paragraph{Benchmark Analysis and Model Behavior.} RadGenome-Anatomy enables evaluation of chest radiograph anatomy segmentation across anatomical scale, view geometry, boundary quality, and component preservation. Table~\ref{tab:benchmark} shows that hierarchical Transformer-based models form the leading performance cluster, indicating the importance of multi-scale context, long-range anatomical layout, and repeated-instance reasoning for projected radiographic anatomy. SegFormer achieves the strongest class-balanced overlap, UPerNet obtains the best HD95 and connected-component recall/F1, and Mask2Former achieves the highest NSD. These complementary optima show that the dataset separates different notions of anatomical quality, including region overlap, surface fidelity, component recovery, and boundary agreement.

Figure~\ref{fig:classes_performance} further shows that the dataset supports structure-level diagnosis. Respiratory structures are the most stable across views, while bone classes show larger variation due to repeated instances, class adjacency, and small-structure localization. Vascular and abdominal structures obtain lower and more dispersed Dice, particularly in lateral views, reflecting thin geometry, weak contrast, truncation, and stronger anatomical superposition. These trends show that high aggregate Dice is mainly driven by large, spatially regular anatomy, while the remaining gap is concentrated in thin, repeated, occluded, or weakly contrasted structures.

\paragraph{Group-wise Analysis.}
The hierarchical structure of anatomical labels in RadGenome-Anatomy enables a stratified evaluation across granularity levels, view projections, and anatomical-group partitions. Table~\ref{tab:multigranularity} shows that performance is highest for coarse anatomical groups and decreases toward organs, substructures, and fine-grained leaf classes, where adjacent anatomy, repeated structures, and weak projected boundaries become harder to separate. The paired PA and LL projections reveal a consistent PA advantage, reflecting stronger superposition and narrower projected silhouettes in lateral views. The pathology partition yields substantially lower performance, showing that disease-related shape variation, heterogeneous appearance, sparse labels, and unstable mask extents remain major challenges. Across anatomical systems, cardiac and pulmonary structures are segmented more reliably, while vascular and abdominal structures remain difficult because of thin geometry, truncation, and overlap with high-attenuation thoracic anatomy.

\section{Anatomy Masks as Measurement-Based Diagnostic Evidence}
\label{sec:measurement_diagnosis}

\begin{wrapfigure}[19]{r}{0.4\textwidth}
  \vspace{-6pt}
  \centering
  \includegraphics[width=\linewidth]{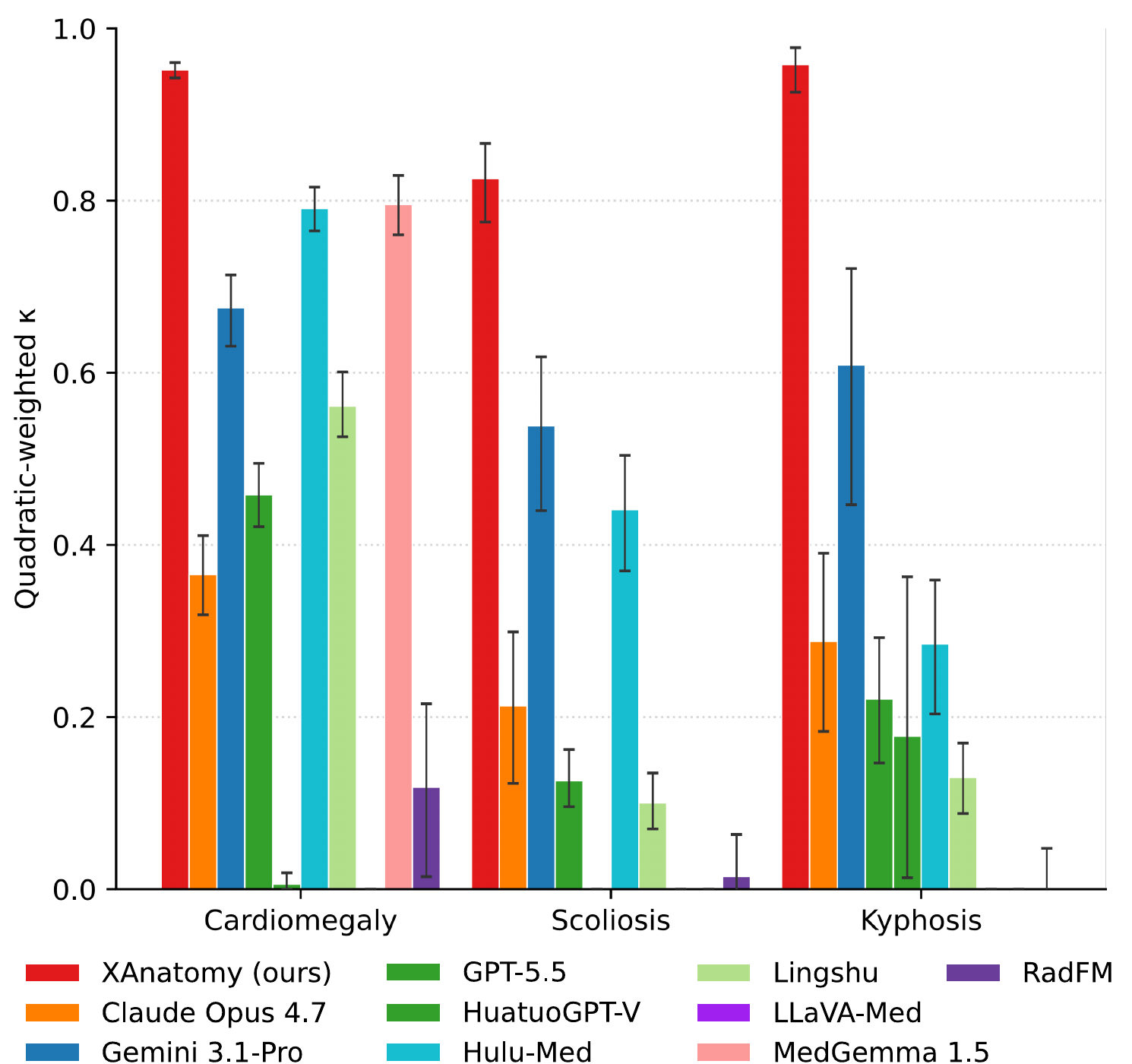}
  \vspace{-10pt}
  \caption{
  Ordinal severity agreement among medical VLMs for cardiomegaly, scoliosis, and kyphosis, measured by quadratic-weighted $\kappa$.
  }
  \label{fig:measurement_qwk}
  \vspace{-10pt}
\end{wrapfigure}

RadGenome-Anatomy supports fine-grained anatomy-based measurement on chest radiographs by providing masks for overlapping structures that are difficult to delineate in 2D. The projected masks convert anatomical geometry into measurable quantities such as structure size, position, curvature, and inter-structure relationships. We demonstrate this capability using \textit{XAnatomy}, an anatomy segmentation model trained on \textit{RadGenome-Anatomy}, to generate structure-specific masks from chest radiographs. Implementation details of the measurement protocol are provided in Appendix~\ref{app:measurement_protocol}.

\paragraph{Dataset}
We evaluate on chest radiographs with paired radiology reports from \textsc{MIMIC-CXR}~\cite{mimic_cxr}. For each target condition, we mine report sentences for positive and negative mentions together with available severity modifiers. The resulting labels are mapped into an ordinal four-grade scale: negative, mild, moderate, and severe. The final evaluable cohorts contain $1{,}287$ cardiomegaly studies, $228$ kyphosis studies, and $809$ scoliosis studies.

\begin{figure*}[t]
  \centering
  \includegraphics[width=\linewidth]{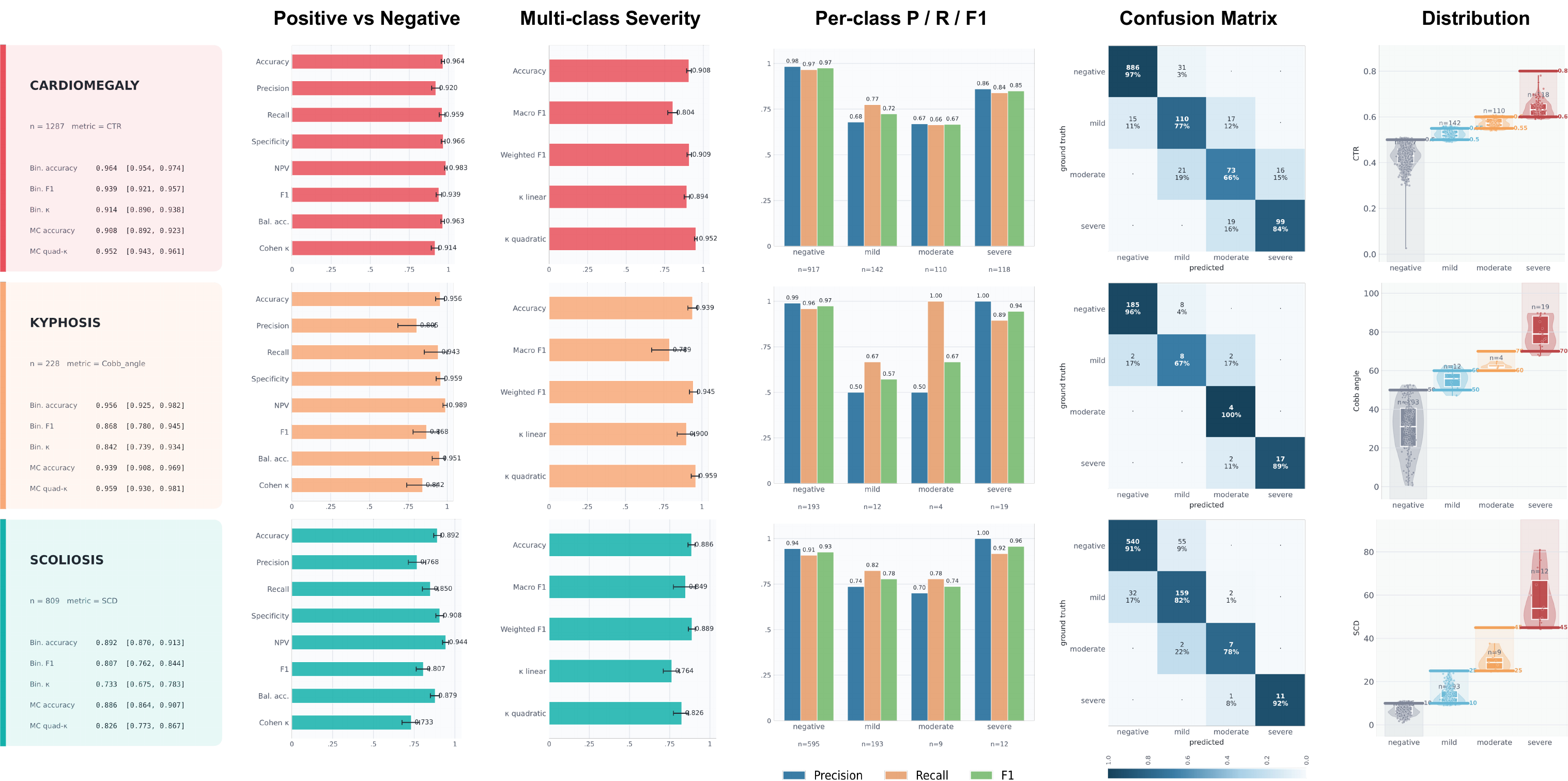}
  \caption{
  Quantitative evaluation of mask-derived measurement-based screening for cardiomegaly, kyphosis, and scoliosis. Each row corresponds to one condition. From left to right, the panels show the cohort summary and headline metrics, binary positive vs. negative performance, ordinal severity performance, per-class precision/recall/$F_1$, row-normalised confusion matrix, and the distribution of the underlying continuous measurement across ground-truth severity grades.
  }
  \label{fig:measurement_scorecard}
\end{figure*}

\begin{figure*}[t]
  \centering
  \includegraphics[width=\linewidth]{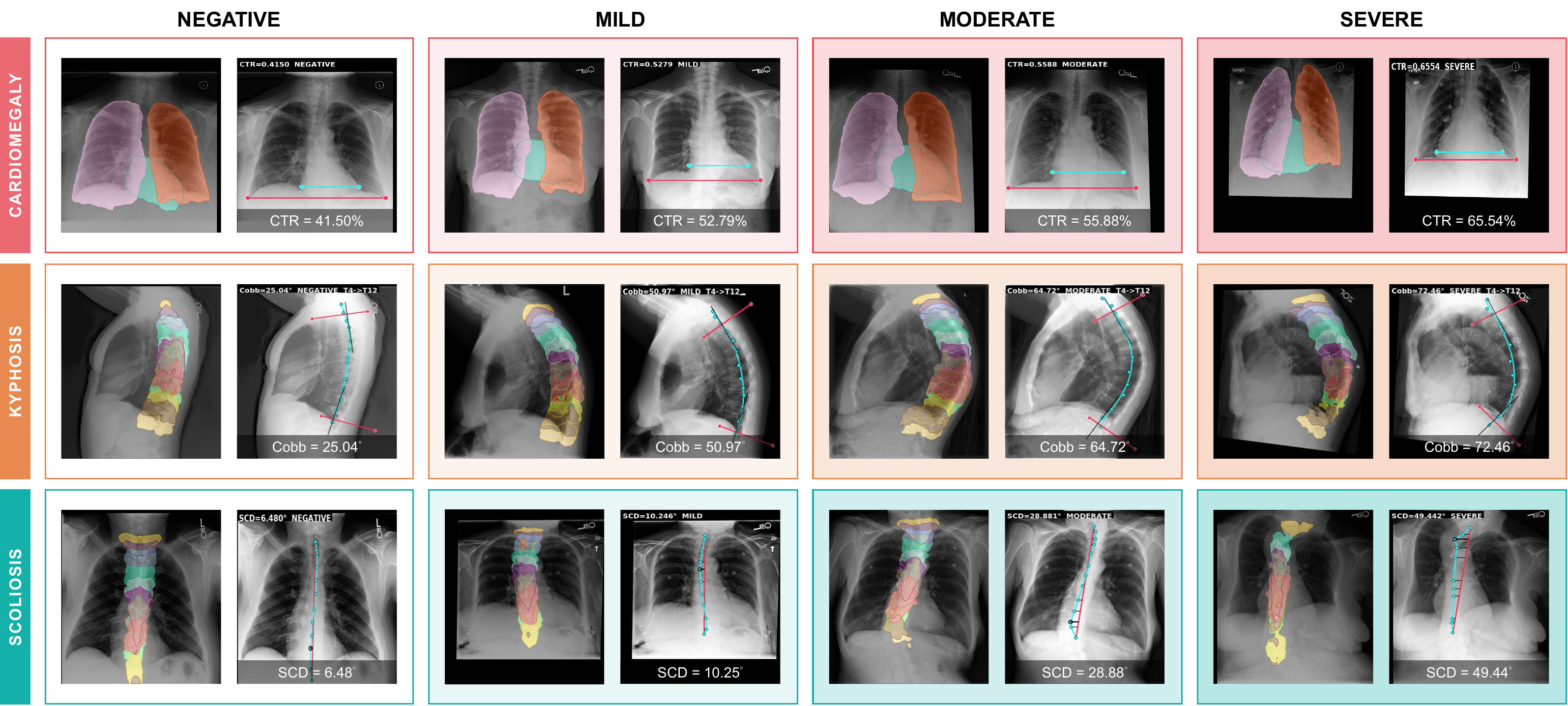}
  \vspace{-10pt}
  \caption{
  Qualitative examples of diagnostic measurements derived from \textit{XAnatomy}-predicted anatomy masks. Columns present cases spanning negative, mild, moderate, and severe categories, while rows correspond to cardiomegaly, thoracic kyphosis, and scoliosis.
  }
  \label{fig:measurement_cases}
  \vspace{-10pt}
\end{figure*}

\paragraph{Anatomical measurements support radiology diagnosis.}
Figure~\ref{fig:measurement_qwk} shows that measurements derived from \textit{XAnatomy} masks achieve stronger ordinal agreement than general vision-language baselines. This outperforming highlights the importance of explicit anatomical measurement for findings whose severity depends on organ size, spatial layout, or spinal curvature. Figure~\ref{fig:measurement_scorecard} further shows that these measurements provide effective negative/positive diagnosis and severity estimation across the three radiology findings. Cardiomegaly obtains the strongest performance, with accuracy $0.964$, indicating that the predicted heart and thoracic masks recover clinically meaningful CTR-based enlargement. Kyphosis also achieves strong binary performance, with accuracy $0.956$, suggesting that lateral-view vertebral geometry can be reliably recovered from the anatomy masks. Scoliosis is more challenging but remains clinically informative, reaching binary accuracy $0.892$ and $F_1=0.807$. This lower performance is consistent with the greater ambiguity of mild spinal curvature on routine chest radiographs, where the clinical boundary between negative and mild cases is inherently soft. The qualitative examples in Figure~\ref{fig:measurement_cases} further show that the predicted anatomy masks expose the structures used to compute each measurement and reveal progressive geometric changes across severity levels. 

\paragraph{Disease Severity}
The distribution panels in Figure~\ref{fig:measurement_scorecard} show a consistent trend across the evaluated findings: measurements derived from \textit{XAnatomy} masks vary systematically with clinical severity and largely align with the threshold ranges used for diagnosis. This indicates that the proposed anatomy-based pipeline recovers severity-relevant geometric evidence, rather than relying only on implicit image-level prediction. Each decision can be traced back to measurable anatomical quantities, such as relative organ size, spatial layout, or curvature, which are directly comparable with clinical criteria. 

\section{Conclusion}
\label{sec:conclusion}

We introduced a new annotation paradigm for chest radiographs that replaces ambiguous and labor-intensive 2D manual tracing with anatomically grounded volumetric-to-radiographic projection. Built on this paradigm, \textit{RadGenome-Anatomy} provides the largest anatomy-labeled chest radiograph resource to date, covering 210 anatomical structures across both PA and lateral views. Beyond scaling dense anatomical supervision, our results suggest that anatomy masks can serve as explicit geometric evidence for clinically relevant measurements, moving AI-assisted chest-radiograph diagnosis from implicit label prediction toward more interpretable and auditable decision support. 
\paragraph{Limitations and Future Work.} While the current measurement-based analysis demonstrates this potential, its clinical utility remains bounded by anatomy-specific segmentation accuracy and requires further validation in clinically grounded settings.

\bibliographystyle{plainnat}
\bibliography{reference}

\newpage

\appendix

\newcommand{\appendixfigures}{
    \renewcommand{\thefigure}{A\arabic{figure}}
    \setcounter{figure}{\value{appendixfig}}
}
\newcounter{appendixfig}

\newcommand{\appendixtables}{
    \renewcommand{\thetable}{A\arabic{table}}
    \setcounter{table}{\value{appendixtab}}
}
\newcounter{appendixtab}

\newcommand{\appendixequations}{
    \renewcommand{\theequation}{A\arabic{equation}}
    \setcounter{equation}{\value{appendixequation}}
}
\newcounter{appendixequation}

\appendixfigures
\appendixtables

\section{Radiographic Transferability and Label Quality}
\label{sec:real_xray_transfer_label_quality}

We assess whether anatomy supervision from \textit{RadGenome-Anatomy} transfers from CT-derived radiographic projections to clinically acquired chest radiographs. \textit{XAnatomy}, trained on \textit{RadGenome-Anatomy}, is evaluated in a zero-shot setting on real chest-radiograph anatomy benchmarks, without using their labels for training. This experiment probes two dataset properties: the radiographic transferability of the projected images and the quality of the projected anatomy masks as segmentation supervision. Comparable recovery of expert-annotated structures in clinical radiographs would indicate that the projected images and labels preserve anatomy that remains meaningful beyond the synthetic projection domain.

We evaluate on SCR-JSRT~\cite{study_on_public,jsrt}, Montgomery~\cite{2datasets}, and CheXmask~\cite{chexmask}. SCR-JSRT provides expert masks for the left lung, right lung, and heart in $247$ PA chest radiographs under the shared three-class setting. Montgomery provides lung annotations for $138$ frontal chest radiographs. CheXmask provides left-lung, right-lung, and heart masks for real chest radiographs. We compare \textit{XAnatomy} with MedSAM as a zero-shot segmentation foundation-model baseline.

\begin{table*}[b]
\centering
\scriptsize
\setlength{\tabcolsep}{3pt}
\renewcommand{\arraystretch}{1.05}
\caption{Zero-shot external validation on real chest-radiograph anatomy benchmarks. \textit{XAnatomy} is trained on \textit{RadGenome-Anatomy} and evaluated directly on real chest-radiograph annotations. MedSAM is evaluated as a zero-shot segmentation foundation-model baseline.}
\label{tab:external_real_radiograph_validation}
\vspace{2mm}
\resizebox{\textwidth}{!}{
\begin{tabular}{lllcccccccccc}
\toprule
\multirow{2}{*}{Dataset} 
& \multirow{2}{*}{Class} 
& \multirow{2}{*}{$n$}
& \multicolumn{2}{c}{Dice $\uparrow$}
& \multicolumn{2}{c}{IoU $\uparrow$}
& \multicolumn{2}{c}{Sens. $\uparrow$}
& \multicolumn{2}{c}{Spec. $\uparrow$}
& \multicolumn{2}{c}{HD95 $\downarrow$} \\
\cmidrule(lr){4-5}
\cmidrule(lr){6-7}
\cmidrule(lr){8-9}
\cmidrule(lr){10-11}
\cmidrule(lr){12-13}
& & 
& XAnat. & MedSAM
& XAnat. & MedSAM
& XAnat. & MedSAM
& XAnat. & MedSAM
& XAnat. & MedSAM \\
\midrule

\multirow{4}{*}{SCR-JSRT}
& Left lung 
& 247
& 0.772 & 0.781
& 0.643 & 0.661
& 0.948 & 0.875
& 0.926 & 0.941
& 78.8 & 73.7 \\

& Right lung 
& 247
& 0.795 & 0.885
& 0.673 & 0.802
& 0.958 & 0.922
& 0.916 & 0.969
& 87.0 & 25.0 \\

& Heart 
& 247
& 0.765 & 0.673
& 0.633 & 0.519
& 0.945 & 0.673
& 0.952 & 0.970
& 126.8 & 82.6 \\

& Macro
& 247
& 0.777 & 0.779
& 0.650 & 0.660
& 0.950 & 0.823
& 0.931 & 0.960
& 97.5 & 60.4 \\

\midrule

Montgomery
& Lung 
& 138
& 0.769 & 0.700
& 0.625 & 0.594
& 0.993 & 0.795
& 0.797 & 0.855
& 364.0 & 404.9 \\

\midrule

\multirow{4}{*}{CheXmask}
& Left lung 
& 4{,}999
& 0.701 & 0.588
& 0.546 & 0.448
& 0.958 & 0.718
& 0.904 & 0.926
& 142.6 & 110.3 \\

& Right lung 
& 4{,}999
& 0.708 & 0.582
& 0.553 & 0.457
& 0.977 & 0.649
& 0.883 & 0.944
& 119.4 & 99.7 \\

& Heart 
& 4{,}999
& 0.814 & 0.669
& 0.692 & 0.519
& 0.906 & 0.652
& 0.969 & 0.975
& 74.4 & 77.3 \\

& Macro
& 4{,}999
& 0.741 & 0.613
& 0.597 & 0.475
& 0.947 & 0.673
& 0.919 & 0.948
& 112.2 & 95.7 \\

\bottomrule
\end{tabular}
}
\end{table*}

Table~\ref{tab:external_real_radiograph_validation} shows that supervision learned from \textit{RadGenome-Anatomy} transfers to real chest radiographs. On SCR-JSRT, \textit{XAnatomy} achieves Dice scores of $0.772$, $0.795$, and $0.765$ for the left lung, right lung, and heart, respectively, with a three-class macro Dice of $0.777$, closely matching MedSAM's macro Dice of $0.779$. On Montgomery, \textit{XAnatomy} obtains a Dice score of $0.769$ for lung segmentation and a sensitivity of $0.993$, showing that the projected-mask supervision recovers nearly all lung regions in a clinical radiograph domain. On CheXmask, \textit{XAnatomy} obtains a macro Dice of $0.741$, macro IoU of $0.597$, and macro sensitivity of $0.947$, with strong performance across the left lung, right lung, and heart.

The comparison with MedSAM provides an external reference for assessing projected-mask quality against expert-labeled real chest-radiograph masks. MedSAM represents a strong zero-shot medical segmentation prior and achieves comparable performance on SCR-JSRT, with similar macro overlap, higher specificity, and lower HD95 for several structures. \textit{XAnatomy} achieves stronger sensitivity across all three datasets and stronger overlap on Montgomery and CheXmask, especially for heart segmentation. Since \textit{XAnatomy} is trained only with CT-derived radiographic projections and projected anatomy masks, its comparable real-domain performance indicates that the projected masks provide supervision aligned with expert-labeled anatomical masks in real chest radiographs. These results support the label quality of \textit{RadGenome-Anatomy} and show that its projected anatomy masks reach a standard suitable for training models that generalize to manually annotated clinical radiographs.

Together, these external results demonstrate that CT-derived radiographic projections in \textit{RadGenome-Anatomy} preserve anatomical information that transfers to clinically acquired chest radiographs. The projected masks train a model that recovers corresponding expert-annotated structures in real radiographs, with comparable overlap to MedSAM and consistently high sensitivity for major thoracic anatomy. This real-domain transfer supports both the radiographic transferability of the projected images and the anatomical quality of the labels, establishing \textit{RadGenome-Anatomy} as a scalable source of geometry-consistent supervision for chest-radiograph anatomy segmentation.

\section{Measurement-Based Diagnostic Protocol}
\label{app:measurement_protocol}

This section documents how predicted anatomical masks are converted into
clinically interpretable geometric measurements for structural diagnosis.
The protocol is deliberately deterministic: no disease-specific classifier,
calibration layer, or learned severity head is trained. Instead, each
diagnostic output is obtained by selecting task-relevant anatomy masks,
extracting geometric primitives from those masks, computing a continuous
measurement, and assigning a severity grade using fixed clinical cut-points.
This design ensures that every prediction can be traced back to a named
anatomical structure and an explicit measurement rule.

\paragraph{Mask notation and coordinate system.}
For an input chest radiograph \(I\), the anatomy model predicts a set of
binary masks
\[
\mathcal{M}(I)=\{\hat{M}_{k}\}_{k=1}^{C},
\]
where \(C=210\) is the number of anatomical classes and
\(\hat{M}_{k}\in\{0,1\}^{H\times W}\) denotes the predicted mask for class
\(k\). We use image coordinates \((x,y)\), where \(x\) indexes the
left--right image axis and \(y\) indexes the superior--inferior axis.
For frontal-view measurements, the masks are taken from the PA projection;
for sagittal spinal measurements, the masks are taken from the lateral
projection.

\paragraph{Mask cleaning and geometric primitives.}
Before measurement, each task uses only the anatomical masks required for
the corresponding clinical quantity. Small isolated components are removed,
and the relevant connected component or union of components is retained
for measurement. For compact structures such as the cardiac silhouette, we
use the foreground extent of the predicted mask. For skeletal structures,
we extract vertebral centroids from the predicted vertebral masks. Given a
binary mask \(\hat{M}\), its foreground centroid is computed as
\[
\mathbf{c}(\hat{M})
=
\frac{1}{|\Omega(\hat{M})|}
\sum_{(x,y)\in\Omega(\hat{M})}
(x,y),
\quad
\Omega(\hat{M})=\{(x,y):\hat{M}(x,y)=1\}.
\]
Cases are excluded from the measurement protocol when the required anatomy
is not sufficiently recovered, for example when the cardiac silhouette is
fragmented or when too few contiguous vertebral masks are detected to form
a stable spinal curve.

\paragraph{Cardiomegaly: cardiothoracic ratio.}
Cardiomegaly is assessed using the cardiothoracic ratio (CTR), which
compares the transverse cardiac width with the thoracic width on a frontal
chest radiograph. From the predicted anatomy masks, we first obtain the
cardiac silhouette mask \(\hat{M}_{\mathrm{heart}}\). We then obtain a
thoracic reference mask \(\hat{M}_{\mathrm{thorax}}\) from the predicted
thoracic boundary structures, using the lung and rib-cage region to
estimate the maximum inner thoracic span. For each image row \(y\), the
horizontal width of a mask is defined as the distance between its leftmost
and rightmost foreground pixels:
\[
w(\hat{M},y)
=
\max\{x:\hat{M}(x,y)=1\}
-
\min\{x:\hat{M}(x,y)=1\}.
\]
The maximum cardiac width is computed by scanning over all rows containing
heart-mask foreground pixels, and the maximum thoracic width is computed
analogously from the thoracic reference mask. The CTR is then
\[
\mathrm{CTR}
=
\frac{
\max_y w(\hat{M}_{\mathrm{heart}},y)
}{
\max_y w(\hat{M}_{\mathrm{thorax}},y)
}.
\]
This produces a scale-normalised measurement, making the grading robust to
image resizing and resolution differences.

\paragraph{Scoliosis: frontal spinal curvature deviation.}
Scoliosis is evaluated from frontal-view vertebral masks. We first collect
the predicted thoracic and visible spine-related vertebral masks and compute
one centroid for each detected vertebra. The centroids are sorted along the
superior--inferior axis to form an ordered spinal centreline:
\[
\mathcal{C}=\{\mathbf{c}_{1},\mathbf{c}_{2},\ldots,\mathbf{c}_{n}\},
\quad
\mathbf{c}_{i}=(x_i,y_i), \quad y_1<y_2<\cdots<y_n.
\]
The upper and lower endpoints are defined as the most superior and most
inferior reliable vertebral centroids, denoted by
\(\mathbf{c}_{\mathrm{top}}\) and \(\mathbf{c}_{\mathrm{bottom}}\). We then
construct the endpoint chord connecting these two points. The apex
vertebra is selected as the centroid with the largest perpendicular
distance from this chord:
\[
\mathbf{c}_{\mathrm{apex}}
=
\arg\max_{\mathbf{c}_{i}\in\mathcal{C}}
d_{\perp}
\left(
\mathbf{c}_{i},
\overline{
\mathbf{c}_{\mathrm{top}}\mathbf{c}_{\mathrm{bottom}}
}
\right).
\]
The spinal curvature deviation (SCD) is computed as the angle induced by
the apex relative to the two endpoint directions:
\[
\mathrm{SCD}
=
\angle
\left(
\mathbf{c}_{\mathrm{top}}-\mathbf{c}_{\mathrm{apex}},
\mathbf{c}_{\mathrm{bottom}}-\mathbf{c}_{\mathrm{apex}}
\right).
\]
A straight spine yields a small SCD, while a laterally deviated spine yields
a larger apex angle. This provides a mask-derived geometric proxy for
frontal spinal curvature severity.

\paragraph{Thoracic kyphosis: sagittal Cobb-style angle.}
Thoracic kyphosis is measured on the lateral chest radiograph using
predicted vertebral masks. As in the scoliosis protocol, we first extract
vertebral centroids and sort them along the superior--inferior axis.
Because individual vertebral masks may be slightly noisy, we do not measure
the angle from a single pair of local components directly. Instead, we fit a
smooth thoracic spine curve through the ordered centroids and compute the
angle between upper and lower tangent directions. Let
\[
\gamma(t)=(x(t),y(t)), \quad t\in[0,1],
\]
denote the fitted spine curve parameterised by normalised arc position. The
unit tangent direction is
\[
\hat{T}(t)
=
\frac{\gamma'(t)}{\|\gamma'(t)\|_2}.
\]
We identify upper and lower thoracic reference positions
\(t_{\mathrm{upper}}\) and \(t_{\mathrm{lower}}\), corresponding to the
upper and lower thoracic vertebral region, and compute a Cobb-style
sagittal angle as
\[
\mathrm{Cobb}
=
\angle
\left(
\hat{T}(t_{\mathrm{upper}}),
\hat{T}(t_{\mathrm{lower}})
\right).
\]
This smooth-curve formulation reduces sensitivity to local mask noise while
preserving the clinically relevant quantity: the angular curvature of the
thoracic spine in the sagittal plane.

\paragraph{Severity grading.}
After the continuous measurements are computed, they are discretised into
four ordinal grades: negative, mild, moderate, and severe. The thresholds
used for each condition are summarised in Table~\ref{tab:measurement_grading}.
Because the grading step is threshold-based, cases near a boundary may move
between adjacent grades after small changes in a mask boundary or vertebral
centroid. We therefore interpret adjacent-grade disagreements as
borderline measurement cases rather than arbitrary classification errors.

\begin{table}[H]
  \centering
  \footnotesize
  \caption{Measurement definitions and severity thresholds used for anatomy-mask-derived structural diagnosis. Each condition is reduced to a clinically interpretable geometric measurement computed from predicted anatomy masks, and the resulting continuous value is discretised into negative, mild, moderate, and severe grades using fixed clinical cut-points.}
  \label{tab:measurement_grading}
  \setlength{\tabcolsep}{3pt}
  \renewcommand{\arraystretch}{1.18}
  \begin{tabularx}{\linewidth}{@{}
    >{\raggedright\arraybackslash}p{0.16\linewidth}
    >{\raggedright\arraybackslash}p{0.20\linewidth}
    >{\raggedright\arraybackslash}X
    >{\raggedright\arraybackslash}p{0.27\linewidth}
    @{}}
    \toprule
    Condition & Required anatomy / view & Mask-derived measurement & Severity grading rule \\
    \midrule

    Cardiomegaly
    &
    Heart and thoracic boundaries in frontal view
    &
    Cardiothoracic ratio (CTR), computed as the maximum cardiac width divided by the maximum thoracic width.
    &
    Negative: $\mathrm{CTR} \leq 0.50$ \newline
    Mild: $0.50 < \mathrm{CTR} \leq 0.55$ \newline
    Moderate: $0.55 < \mathrm{CTR} \leq 0.60$ \newline
    Severe: $\mathrm{CTR} > 0.60$
    \\

    \midrule

    Scoliosis
    &
    Frontal-view vertebral masks
    &
    Spinal curvature deviation (SCD), computed from vertebral centroids as the angular deviation induced by the apex vertebra relative to the upper and lower spinal endpoints.
    &
    Negative: $\mathrm{SCD} < 10^\circ$ \newline
    Mild: $10^\circ \leq \mathrm{SCD} < 25^\circ$ \newline
    Moderate: $25^\circ \leq \mathrm{SCD} < 45^\circ$ \newline
    Severe: $\mathrm{SCD} \geq 45^\circ$
    \\

    \midrule

    Thoracic kyphosis
    &
    Lateral-view vertebral masks
    &
    Cobb-style sagittal angle, computed by fitting a smooth thoracic spine curve to vertebral centroids and measuring the angle between upper and lower thoracic tangents.
    &
    Negative: $\mathrm{Cobb} < 50^\circ$ \newline
    Mild: $50^\circ \leq \mathrm{Cobb} < 60^\circ$ \newline
    Moderate: $60^\circ \leq \mathrm{Cobb} < 70^\circ$ \newline
    Severe: $\mathrm{Cobb} \geq 70^\circ$
    \\

    \bottomrule
  \end{tabularx}
\end{table}

\section{Pairwise Statistical Analysis}

We further examine whether the model rankings are stable across anatomical categories by conducting paired comparisons over the shared set of $210$ anatomy classes. For each model pair, we apply the Wilcoxon signed-rank test to class-wise scores and report paired-bootstrap significance, Cohen's $d$, and rank-biserial correlation. Bonferroni correction is applied over all $171$ pairwise comparisons. This analysis is intended to complement the aggregate benchmark results: while mean performance summarises overall accuracy, the paired tests evaluate whether one model consistently improves over another across the anatomy vocabulary.

\begin{table}[H]
  \centering
  \tiny
  \setlength{\tabcolsep}{1.6pt}
  \renewcommand{\arraystretch}{0.78}
  \caption{Complete pairwise Wilcoxon signed-rank comparisons across the $210$ anatomical classes. To fit the full set of $171$ comparisons on one page, the table is arranged as two side-by-side blocks. For each model pair, we report the Wilcoxon statistic $W$, Bonferroni-corrected $p$-value $p_{\mathrm{Bonf}}$, Cohen's $d$, rank-biserial correlation $r_{\mathrm{rb}}$, and Bonferroni significance. Effect sizes are signed as the first model minus the second model. Positive values favour the first model.}
  \label{tab:pairwise_wilcoxon_compact_full}
  \begin{adjustbox}{max width=\linewidth,max totalheight=0.70\textheight}
  \begin{tabular}{@{}lrrrrc@{\hspace{0.75em}}lrrrrc@{}}
    \toprule
    Pair & $W$ & $p_{\mathrm{Bonf}}$ & $d$ & $r_{\mathrm{rb}}$ & Sig. &
    Pair & $W$ & $p_{\mathrm{Bonf}}$ & $d$ & $r_{\mathrm{rb}}$ & Sig. \\
    \midrule
    DeepLabV3+--Mask2Former & $378$ & $2.60\times 10^{-31}$ & $-0.950$ & $-0.966$ & \checkmark & PointRend--U-KAN & $8668$ & $1.000$ & $0.205$ & $0.189$ & -- \\
    DeepLabV3+--Med-SAM-A & $12$ & $1.43\times 10^{-33}$ & $1.087$ & $0.999$ & \checkmark & PointRend--U-Mamba & $10270$ & $1.000$ & $0.054$ & $0.038$ & -- \\
    DeepLabV3+--nnU-Net & $955$ & $6.66\times 10^{-28}$ & $0.828$ & $0.911$ & \checkmark & PointRend--U-Net & $9816$ & $1.000$ & $0.009$ & $-0.078$ & -- \\
    DeepLabV3+--PIDNet & $8434$ & $0.872$ & $0.186$ & $0.224$ & -- & PointRend--U-Net++ & $5984$ & $8.18\times 10^{-6}$ & $-0.301$ & $-0.436$ & \checkmark \\
    DeepLabV3+--PointRend & $949$ & $6.16\times 10^{-28}$ & $1.002$ & $0.910$ & \checkmark & PointRend--UNeXt & $161$ & $2.64\times 10^{-32}$ & $1.116$ & $0.984$ & \checkmark \\
    DeepLabV3+--SeaFormer & $0$ & $1.20\times 10^{-33}$ & $1.293$ & $1.000$ & \checkmark & PointRend--UPerNet & $0$ & $1.20\times 10^{-33}$ & $-1.264$ & $-1.000$ & \checkmark \\
    DeepLabV3+--SegFormer & $380$ & $1.79\times 10^{-31}$ & $-0.934$ & $-0.966$ & \checkmark & PointRend--VMUNet & $10263$ & $1.000$ & $-0.025$ & $-0.037$ & -- \\
    DeepLabV3+--S-SAM & $110$ & $5.85\times 10^{-33}$ & $1.104$ & $0.989$ & \checkmark & SeaFormer--SegFormer & $0$ & $1.75\times 10^{-33}$ & $-1.386$ & $-1.000$ & \checkmark \\
    DeepLabV3+--Swin-UNet & $127$ & $7.46\times 10^{-33}$ & $1.142$ & $0.987$ & \checkmark & SeaFormer--S-SAM & $0$ & $3.58\times 10^{-32}$ & $-1.154$ & $-1.000$ & \checkmark \\
    DeepLabV3+--TopFormer & $0$ & $1.20\times 10^{-33}$ & $1.263$ & $1.000$ & \checkmark & SeaFormer--Swin-UNet & $0$ & $3.72\times 10^{-33}$ & $-1.241$ & $-1.000$ & \checkmark \\
    DeepLabV3+--TransUNet & $310$ & $1.00\times 10^{-31}$ & $0.903$ & $0.970$ & \checkmark & SeaFormer--TopFormer & $0$ & $1.68\times 10^{-32}$ & $-1.178$ & $-1.000$ & \checkmark \\
    DeepLabV3+--U-KAN & $2$ & $1.23\times 10^{-33}$ & $1.212$ & $1.000$ & \checkmark & SeaFormer--TransUNet & $49$ & $7.51\times 10^{-32}$ & $-1.069$ & $-0.993$ & \checkmark \\
    DeepLabV3+--U-Mamba & $258$ & $4.81\times 10^{-32}$ & $1.082$ & $0.976$ & \checkmark & SeaFormer--U-KAN & $0$ & $5.22\times 10^{-32}$ & $-1.115$ & $-1.000$ & \checkmark \\
    DeepLabV3+--U-Net & $282$ & $6.75\times 10^{-32}$ & $0.958$ & $0.972$ & \checkmark & SeaFormer--U-Mamba & $0$ & $2.46\times 10^{-32}$ & $-1.183$ & $-1.000$ & \checkmark \\
    DeepLabV3+--U-Net++ & $3472$ & $2.97\times 10^{-15}$ & $0.529$ & $0.678$ & \checkmark & SeaFormer--U-Net & $35$ & $8.89\times 10^{-32}$ & $-1.105$ & $-0.991$ & \checkmark \\
    DeepLabV3+--UNeXt & $0$ & $1.20\times 10^{-33}$ & $1.363$ & $1.000$ & \checkmark & SeaFormer--U-Net++ & $1$ & $2.49\times 10^{-32}$ & $-1.136$ & $-0.999$ & \checkmark \\
    DeepLabV3+--UPerNet & $88$ & $4.27\times 10^{-33}$ & $-0.922$ & $-0.991$ & \checkmark & SeaFormer--UNeXt & $1211$ & $1.67\times 10^{-23}$ & $-0.785$ & $-0.853$ & \checkmark \\
    DeepLabV3+--VMUNet & $248$ & $4.18\times 10^{-32}$ & $1.118$ & $0.976$ & \checkmark & SeaFormer--UPerNet & $0$ & $1.20\times 10^{-33}$ & $-1.380$ & $-1.000$ & \checkmark \\
    Mask2Former--Med-SAM-A & $0$ & $2.55\times 10^{-33}$ & $1.209$ & $1.000$ & \checkmark & SeaFormer--VMUNet & $0$ & $7.62\times 10^{-32}$ & $-1.154$ & $-1.000$ & \checkmark \\
    Mask2Former--nnU-Net & $0$ & $2.55\times 10^{-33}$ & $1.351$ & $1.000$ & \checkmark & SegFormer--S-SAM & $0$ & $1.75\times 10^{-33}$ & $1.363$ & $1.000$ & \checkmark \\
    Mask2Former--PIDNet & $306$ & $9.47\times 10^{-32}$ & $1.153$ & $0.972$ & \checkmark & SegFormer--Swin-UNet & $0$ & $1.75\times 10^{-33}$ & $1.393$ & $1.000$ & \checkmark \\
    Mask2Former--PointRend & $16$ & $3.22\times 10^{-33}$ & $1.385$ & $0.998$ & \checkmark & SegFormer--TopFormer & $0$ & $1.75\times 10^{-33}$ & $1.405$ & $1.000$ & \checkmark \\
    Mask2Former--SeaFormer & $0$ & $2.55\times 10^{-33}$ & $1.352$ & $1.000$ & \checkmark & SegFormer--TransUNet & $0$ & $1.75\times 10^{-33}$ & $1.214$ & $1.000$ & \checkmark \\
    Mask2Former--SegFormer & $5816$ & $1.67\times 10^{-6}$ & $-0.169$ & $-0.458$ & \checkmark & SegFormer--U-KAN & $0$ & $1.75\times 10^{-33}$ & $1.403$ & $1.000$ & \checkmark \\
    Mask2Former--S-SAM & $0$ & $2.55\times 10^{-33}$ & $1.345$ & $1.000$ & \checkmark & SegFormer--U-Mamba & $0$ & $1.75\times 10^{-33}$ & $1.360$ & $1.000$ & \checkmark \\
    Mask2Former--Swin-UNet & $2$ & $2.63\times 10^{-33}$ & $1.382$ & $0.999$ & \checkmark & SegFormer--U-Net & $0$ & $1.75\times 10^{-33}$ & $1.308$ & $1.000$ & \checkmark \\
    Mask2Former--TopFormer & $0$ & $2.55\times 10^{-33}$ & $1.364$ & $1.000$ & \checkmark & SegFormer--U-Net++ & $2$ & $1.80\times 10^{-33}$ & $1.051$ & $1.000$ & \checkmark \\
    Mask2Former--TransUNet & $0$ & $2.55\times 10^{-33}$ & $1.193$ & $1.000$ & \checkmark & SegFormer--UNeXt & $0$ & $1.75\times 10^{-33}$ & $1.472$ & $1.000$ & \checkmark \\
    Mask2Former--U-KAN & $0$ & $2.55\times 10^{-33}$ & $1.371$ & $1.000$ & \checkmark & SegFormer--UPerNet & $7116$ & $0.002$ & $0.027$ & $0.353$ & \checkmark \\
    Mask2Former--U-Mamba & $0$ & $2.55\times 10^{-33}$ & $1.336$ & $1.000$ & \checkmark & SegFormer--VMUNet & $0$ & $1.75\times 10^{-33}$ & $1.385$ & $1.000$ & \checkmark \\
    Mask2Former--U-Net & $0$ & $2.55\times 10^{-33}$ & $1.310$ & $1.000$ & \checkmark & S-SAM--Swin-UNet & $7875$ & $0.275$ & $-0.271$ & $-0.253$ & -- \\
    Mask2Former--U-Net++ & $0$ & $2.55\times 10^{-33}$ & $1.050$ & $1.000$ & \checkmark & S-SAM--TopFormer & $1163$ & $2.33\times 10^{-25}$ & $0.880$ & $0.873$ & \checkmark \\
    Mask2Former--UNeXt & $0$ & $2.55\times 10^{-33}$ & $1.436$ & $1.000$ & \checkmark & S-SAM--TransUNet & $8081$ & $1.000$ & $0.232$ & $0.198$ & -- \\
    Mask2Former--UPerNet & $9317$ & $1.000$ & $-0.092$ & $0.145$ & -- & S-SAM--U-KAN & $3934$ & $2.40\times 10^{-11}$ & $0.424$ & $0.598$ & \checkmark \\
    Mask2Former--VMUNet & $0$ & $2.55\times 10^{-33}$ & $1.373$ & $1.000$ & \checkmark & S-SAM--U-Mamba & $7684$ & $0.665$ & $0.131$ & $0.229$ & -- \\
    Med-SAM-A--nnU-Net & $132$ & $1.20\times 10^{-31}$ & $-0.879$ & $-0.982$ & \checkmark & S-SAM--U-Net & $9878$ & $1.000$ & $0.042$ & $0.024$ & -- \\
    Med-SAM-A--PIDNet & $56$ & $2.69\times 10^{-33}$ & $-1.072$ & $-0.994$ & \checkmark & S-SAM--U-Net++ & $3460$ & $9.18\times 10^{-14}$ & $-0.473$ & $-0.656$ & \checkmark \\
    Med-SAM-A--PointRend & $2920$ & $2.78\times 10^{-17}$ & $-0.601$ & $-0.720$ & \checkmark & S-SAM--UNeXt & $161$ & $4.02\times 10^{-31}$ & $1.238$ & $0.982$ & \checkmark \\
    Med-SAM-A--SeaFormer & $197$ & $7.24\times 10^{-30}$ & $1.028$ & $0.968$ & \checkmark & S-SAM--UPerNet & $0$ & $1.20\times 10^{-33}$ & $-1.153$ & $-1.000$ & \checkmark \\
    Med-SAM-A--SegFormer & $0$ & $1.75\times 10^{-33}$ & $-1.256$ & $-1.000$ & \checkmark & S-SAM--VMUNet & $8749$ & $1.000$ & $-0.017$ & $0.124$ & -- \\
    Med-SAM-A--S-SAM & $799$ & $3.97\times 10^{-27}$ & $-0.844$ & $-0.915$ & \checkmark & Swin-UNet--TopFormer & $162$ & $3.95\times 10^{-32}$ & $1.112$ & $0.981$ & \checkmark \\
    Med-SAM-A--Swin-UNet & $364$ & $7.14\times 10^{-31}$ & $-0.798$ & $-0.962$ & \checkmark & Swin-UNet--TransUNet & $7521$ & $0.061$ & $0.316$ & $0.284$ & -- \\
    Med-SAM-A--TopFormer & $7758$ & $0.643$ & $0.139$ & $0.233$ & -- & Swin-UNet--U-KAN & $3131$ & $4.24\times 10^{-16}$ & $0.566$ & $0.697$ & \checkmark \\
    Med-SAM-A--TransUNet & $2142$ & $2.26\times 10^{-19}$ & $-0.659$ & $-0.770$ & \checkmark & Swin-UNet--U-Mamba & $5607$ & $9.98\times 10^{-7}$ & $0.399$ & $0.464$ & \checkmark \\
    Med-SAM-A--U-KAN & $2049$ & $7.38\times 10^{-20}$ & $-0.665$ & $-0.765$ & \checkmark & Swin-UNet--U-Net & $8356$ & $1.000$ & $0.210$ & $0.209$ & -- \\
    Med-SAM-A--U-Mamba & $1388$ & $7.09\times 10^{-24}$ & $-0.685$ & $-0.852$ & \checkmark & Swin-UNet--U-Net++ & $6643$ & $7.12\times 10^{-4}$ & $-0.237$ & $-0.366$ & \checkmark \\
    Med-SAM-A--U-Net & $1388$ & $1.78\times 10^{-23}$ & $-0.738$ & $-0.843$ & \checkmark & Swin-UNet--UNeXt & $5$ & $4.00\times 10^{-33}$ & $1.233$ & $0.999$ & \checkmark \\
    Med-SAM-A--U-Net++ & $355$ & $4.70\times 10^{-30}$ & $-0.924$ & $-0.957$ & \checkmark & Swin-UNet--UPerNet & $3$ & $1.25\times 10^{-33}$ & $-1.171$ & $-0.999$ & \checkmark \\
    Med-SAM-A--UNeXt & $374$ & $1.58\times 10^{-28}$ & $1.141$ & $0.955$ & \checkmark & Swin-UNet--VMUNet & $7233$ & $0.016$ & $0.260$ & $0.313$ & \checkmark \\
    Med-SAM-A--UPerNet & $0$ & $1.20\times 10^{-33}$ & $-1.134$ & $-1.000$ & \checkmark & TopFormer--TransUNet & $3216$ & $4.67\times 10^{-15}$ & $-0.590$ & $-0.676$ & \checkmark \\
    Med-SAM-A--VMUNet & $658$ & $1.28\times 10^{-27}$ & $-0.788$ & $-0.920$ & \checkmark & TopFormer--U-KAN & $2138$ & $4.96\times 10^{-20}$ & $-0.735$ & $-0.778$ & \checkmark \\
    nnU-Net--PIDNet & $2240$ & $5.40\times 10^{-21}$ & $-0.688$ & $-0.792$ & \checkmark & TopFormer--U-Mamba & $1154$ & $2.07\times 10^{-25}$ & $-0.875$ & $-0.871$ & \checkmark \\
    nnU-Net--PointRend & $5649$ & $8.39\times 10^{-7}$ & $0.358$ & $0.465$ & \checkmark & TopFormer--U-Net & $2129$ & $4.46\times 10^{-20}$ & $-0.733$ & $-0.776$ & \checkmark \\
    nnU-Net--SeaFormer & $0$ & $1.68\times 10^{-32}$ & $1.190$ & $1.000$ & \checkmark & TopFormer--U-Net++ & $979$ & $1.97\times 10^{-26}$ & $-0.867$ & $-0.894$ & \checkmark \\
    nnU-Net--SegFormer & $0$ & $1.75\times 10^{-33}$ & $-1.319$ & $-1.000$ & \checkmark & TopFormer--UNeXt & $966$ & $1.65\times 10^{-26}$ & $0.789$ & $0.893$ & \checkmark \\
    nnU-Net--S-SAM & $3488$ & $1.21\times 10^{-13}$ & $0.578$ & $0.641$ & \checkmark & TopFormer--UPerNet & $0$ & $1.20\times 10^{-33}$ & $-1.307$ & $-1.000$ & \checkmark \\
    nnU-Net--Swin-UNet & $7273$ & $0.019$ & $0.290$ & $0.301$ & \checkmark & TopFormer--VMUNet & $983$ & $2.08\times 10^{-26}$ & $-0.879$ & $-0.889$ & \checkmark \\
    nnU-Net--TopFormer & $345$ & $2.72\times 10^{-30}$ & $1.012$ & $0.959$ & \checkmark & TransUNet--U-KAN & $7969$ & $1.000$ & $0.030$ & $0.206$ & -- \\
    nnU-Net--TransUNet & $3969$ & $7.31\times 10^{-12}$ & $0.514$ & $0.598$ & \checkmark & TransUNet--U-Mamba & $8879$ & $1.000$ & $-0.130$ & $-0.124$ & -- \\
    nnU-Net--U-KAN & $913$ & $8.03\times 10^{-27}$ & $0.762$ & $0.904$ & \checkmark & TransUNet--U-Net & $7088$ & $0.074$ & $-0.326$ & $-0.273$ & -- \\
    nnU-Net--U-Mamba & $3177$ & $5.17\times 10^{-15}$ & $0.574$ & $0.674$ & \checkmark & TransUNet--U-Net++ & $632$ & $1.63\times 10^{-28}$ & $-0.878$ & $-0.933$ & \checkmark \\
    nnU-Net--U-Net & $3942$ & $9.43\times 10^{-12}$ & $0.478$ & $0.602$ & \checkmark & TransUNet--UNeXt & $6$ & $3.92\times 10^{-32}$ & $1.232$ & $0.998$ & \checkmark \\
    nnU-Net--U-Net++ & $7023$ & $0.026$ & $-0.085$ & $-0.311$ & \checkmark & TransUNet--UPerNet & $0$ & $1.20\times 10^{-33}$ & $-1.048$ & $-1.000$ & \checkmark \\
    nnU-Net--UNeXt & $0$ & $1.68\times 10^{-32}$ & $1.251$ & $1.000$ & \checkmark & TransUNet--VMUNet & $8194$ & $1.000$ & $-0.238$ & $-0.183$ & -- \\
    nnU-Net--UPerNet & $0$ & $1.20\times 10^{-33}$ & $-1.068$ & $-1.000$ & \checkmark & U-KAN--U-Mamba & $6680$ & $0.007$ & $-0.226$ & $-0.333$ & \checkmark \\
    nnU-Net--VMUNet & $3159$ & $4.29\times 10^{-15}$ & $0.575$ & $0.673$ & \checkmark & U-KAN--U-Net & $6052$ & $2.82\times 10^{-4}$ & $-0.246$ & $-0.381$ & \checkmark \\
    PIDNet--PointRend & $622$ & $7.58\times 10^{-30}$ & $1.064$ & $0.941$ & \checkmark & U-KAN--U-Net++ & $1924$ & $3.78\times 10^{-21}$ & $-0.685$ & $-0.806$ & \checkmark \\
    PIDNet--SeaFormer & $0$ & $1.20\times 10^{-33}$ & $1.288$ & $1.000$ & \checkmark & U-KAN--UNeXt & $8$ & $5.90\times 10^{-32}$ & $1.149$ & $0.997$ & \checkmark \\
    PIDNet--SegFormer & $280$ & $4.42\times 10^{-32}$ & $-1.091$ & $-0.975$ & \checkmark & U-KAN--UPerNet & $0$ & $1.20\times 10^{-33}$ & $-1.204$ & $-1.000$ & \checkmark \\
    PIDNet--S-SAM & $188$ & $1.78\times 10^{-32}$ & $1.039$ & $0.982$ & \checkmark & U-KAN--VMUNet & $4440$ & $3.52\times 10^{-9}$ & $-0.400$ & $-0.542$ & \checkmark \\
    PIDNet--Swin-UNet & $703$ & $2.28\times 10^{-29}$ & $0.992$ & $0.932$ & \checkmark & U-Mamba--U-Net & $9726$ & $1.000$ & $-0.050$ & $-0.021$ & -- \\
    PIDNet--TopFormer & $1$ & $1.22\times 10^{-33}$ & $1.231$ & $1.000$ & \checkmark & U-Mamba--U-Net++ & $3411$ & $5.63\times 10^{-14}$ & $-0.498$ & $-0.659$ & \checkmark \\
    PIDNet--TransUNet & $368$ & $2.26\times 10^{-31}$ & $0.841$ & $0.965$ & \checkmark & U-Mamba--UNeXt & $4$ & $2.61\times 10^{-32}$ & $1.193$ & $0.999$ & \checkmark \\
    PIDNet--U-KAN & $0$ & $1.20\times 10^{-33}$ & $1.132$ & $1.000$ & \checkmark & U-Mamba--UPerNet & $0$ & $1.20\times 10^{-33}$ & $-1.197$ & $-1.000$ & \checkmark \\
    PIDNet--U-Mamba & $433$ & $5.60\times 10^{-31}$ & $0.974$ & $0.959$ & \checkmark & U-Mamba--VMUNet & $7664$ & $0.615$ & $-0.142$ & $-0.227$ & -- \\
    PIDNet--U-Net & $583$ & $4.45\times 10^{-30}$ & $0.857$ & $0.945$ & \checkmark & U-Net--U-Net++ & $1633$ & $1.64\times 10^{-22}$ & $-0.784$ & $-0.832$ & \checkmark \\
    PIDNet--U-Net++ & $4886$ & $1.00\times 10^{-9}$ & $0.431$ & $0.549$ & \checkmark & U-Net--UNeXt & $6$ & $5.72\times 10^{-32}$ & $1.201$ & $0.998$ & \checkmark \\
    PIDNet--UNeXt & $0$ & $1.20\times 10^{-33}$ & $1.358$ & $1.000$ & \checkmark & U-Net--UPerNet & $0$ & $1.20\times 10^{-33}$ & $-1.102$ & $-1.000$ & \checkmark \\
    PIDNet--UPerNet & $49$ & $2.43\times 10^{-33}$ & $-0.905$ & $-0.995$ & \checkmark & U-Net--VMUNet & $9449$ & $1.000$ & $-0.052$ & $-0.061$ & -- \\
    PIDNet--VMUNet & $400$ & $3.54\times 10^{-31}$ & $1.000$ & $0.962$ & \checkmark & U-Net++--UNeXt & $0$ & $3.58\times 10^{-32}$ & $1.287$ & $1.000$ & \checkmark \\
    PointRend--SeaFormer & $1$ & $2.59\times 10^{-33}$ & $1.268$ & $1.000$ & \checkmark & U-Net++--UPerNet & $10$ & $1.39\times 10^{-33}$ & $-0.869$ & $-0.999$ & \checkmark \\
    PointRend--SegFormer & $1$ & $1.77\times 10^{-33}$ & $-1.428$ & $-1.000$ & \checkmark & U-Net++--VMUNet & $3401$ & $5.09\times 10^{-14}$ & $0.494$ & $0.660$ & \checkmark \\
    PointRend--S-SAM & $9924$ & $1.000$ & $-0.019$ & $-0.068$ & -- & UNeXt--UPerNet & $0$ & $1.20\times 10^{-33}$ & $-1.399$ & $-1.000$ & \checkmark \\
    PointRend--Swin-UNet & $7136$ & $0.007$ & $-0.246$ & $-0.331$ & \checkmark & UNeXt--VMUNet & $15$ & $9.58\times 10^{-32}$ & $-1.220$ & $-0.997$ & \checkmark \\
    PointRend--TopFormer & $611$ & $1.50\times 10^{-29}$ & $0.947$ & $0.940$ & \checkmark & UPerNet--VMUNet & $0$ & $1.20\times 10^{-33}$ & $1.174$ & $1.000$ & \checkmark \\
    PointRend--TransUNet & $10307$ & $1.000$ & $0.129$ & $0.033$ & -- &  &  &  &  &  &  \\
    \bottomrule
  \end{tabular}
  \end{adjustbox}
\end{table}

The pairwise results reveal that the benchmark induces a strongly ordered, but not fully total, model hierarchy. Out of $171$ comparisons, $148$ remain significant after Bonferroni correction, indicating that most observed performance differences are not isolated to a small number of anatomical classes. SegFormer has the strongest dominance profile, with significant Wilcoxon advantages over all other models and no significant losses. However, the magnitude of this dominance is not uniform. Its advantage over most lower-ranked models is accompanied by large signed effect sizes and rank-biserial correlations close to one, suggesting broad class-wise superiority. In contrast, its comparison with UPerNet shows only a very small Cohen's $d$ and is not supported by the paired-bootstrap test, despite remaining significant under the corrected Wilcoxon test. We therefore interpret SegFormer as the strongest model under the rank-based paired test, but the practical margin between SegFormer and UPerNet should be regarded as limited.

Mask2Former and UPerNet form the most competitive group below SegFormer. Both models significantly outperform nearly all other baselines, and their direct comparison is not significant after correction. This suggests that their relative ordering is not reliably distinguishable at the class-wise level. The main difference between them lies in their relation to SegFormer: Mask2Former is significantly below SegFormer with a small-to-moderate signed effect, whereas UPerNet shows a weaker and less bootstrap-stable difference. Thus, the top-performing region of the benchmark is better interpreted as a compact group consisting of SegFormer, UPerNet, and Mask2Former, rather than as a set of widely separated models.

DeepLabV3+ and PIDNet form a second high-performing group. Both models significantly outperform most remaining methods and are significantly weaker than SegFormer, UPerNet, and Mask2Former. Their direct comparison is not significant after Bonferroni correction, indicating that the benchmark does not provide reliable evidence to distinguish them. This result is informative because the two models have different architectural motivations, yet their class-wise anatomy segmentation performance is statistically comparable under this evaluation. Their shared failure relative to the top group suggests that the main performance gap is not merely between convolutional and transformer-based designs, but between models that can consistently preserve fine-grained anatomical boundaries across many structures and those whose gains are less uniform across the label space.

The middle region of the benchmark is less strictly ordered. Models including PointRend, S-SAM, Swin-UNet, TransUNet, U-KAN, U-Mamba, U-Net, and VMUNet exhibit multiple non-significant pairwise differences. This indicates that their aggregate scores should not be over-interpreted as a stable ranking. Several of these models are statistically distinguishable from clearly stronger or weaker methods, but they are not consistently separable from one another. In practical terms, this means that small differences in their mean benchmark scores may reflect class-specific trade-offs rather than systematic superiority across the full anatomy vocabulary. For dense radiographic anatomy segmentation, such behaviour is expected: different structures vary substantially in size, contrast, overlap, and boundary visibility, so a model may improve on large organs while offering limited gains for small or weakly visible structures.

The lower-performing group is also clearly identified. SeaFormer is significantly worse than every other model, while UNeXt is significantly better than SeaFormer but significantly worse than all remaining methods. Med-SAM-A and TopFormer are also dominated by most baselines and are not significantly different from each other. These results suggest that direct adaptation or lightweight segmentation designs may be insufficient for this setting when evaluated over a large and fine-grained thoracic anatomy vocabulary. The failure is not simply low mean performance; rather, the signed rank statistics show that these models are consistently inferior across many anatomical classes.

Overall, the pairwise analysis supports three conclusions. First, RadGenome-Anatomy provides enough class-level granularity to statistically resolve performance differences among segmentation architectures. Second, the strongest methods are those that maintain consistent gains across the anatomy vocabulary, not merely those that improve the mean score on a subset of easy structures. Third, statistical significance must be interpreted together with effect size and bootstrap stability. Some comparisons are significant because the paired test aggregates evidence over $210$ classes, but their practical differences are small. Therefore, the most reliable conclusions are the separation between the top group and the rest of the benchmark, the statistical equivalence within several adjacent model groups, and the consistent underperformance of the lowest-ranked methods.

\section{Licenses for Existing Assets}
\label{app:licenses_existing_assets}

This work builds on existing medical-imaging datasets and public model implementations, and all reused assets are credited through their original publications or official repositories. The volumetric source data are derived from RadGenome-Chest CT, which is released under a CC BY 4.0 license, and from its underlying CT-RATE resource, which is distributed under a CC BY-NC-SA 4.0 license with gated access and additional terms restricting use to academic, research, and educational purposes. We comply with these terms by using the data only for research, by citing the original dataset creators, and by not redistributing the original CT volumes or reports.

The measurement-based diagnostic evaluation uses MIMIC-CXR, which is distributed through PhysioNet under the PhysioNet Credentialed Health Data License 1.5.0. Access to MIMIC-CXR requires user credentialing, completion of the required training, and acceptance of the PhysioNet data use agreement. We use MIMIC-CXR only for research evaluation, do not attempt to re-identify individuals, and do not redistribute any original MIMIC-CXR images or reports.

The segmentation benchmark uses public model architectures, including CNN-, Transformer-, Mamba-, KAN-, and SAM-based methods, each cited in the main paper. These methods are used for comparative research evaluation, and their original authors retain ownership of the corresponding code, models, and documentation. When public implementations or pretrained weights are used, we follow the licenses and usage terms specified by the corresponding repositories. Our released code and dataset documentation will specify the licenses of all reused software dependencies and clarify that users remain responsible for obtaining restricted medical datasets through their original access procedures.

\section{Appendix Figures}

\begin{figure}[H]
  \centering
  \includegraphics[height=0.85\textheight,keepaspectratio]{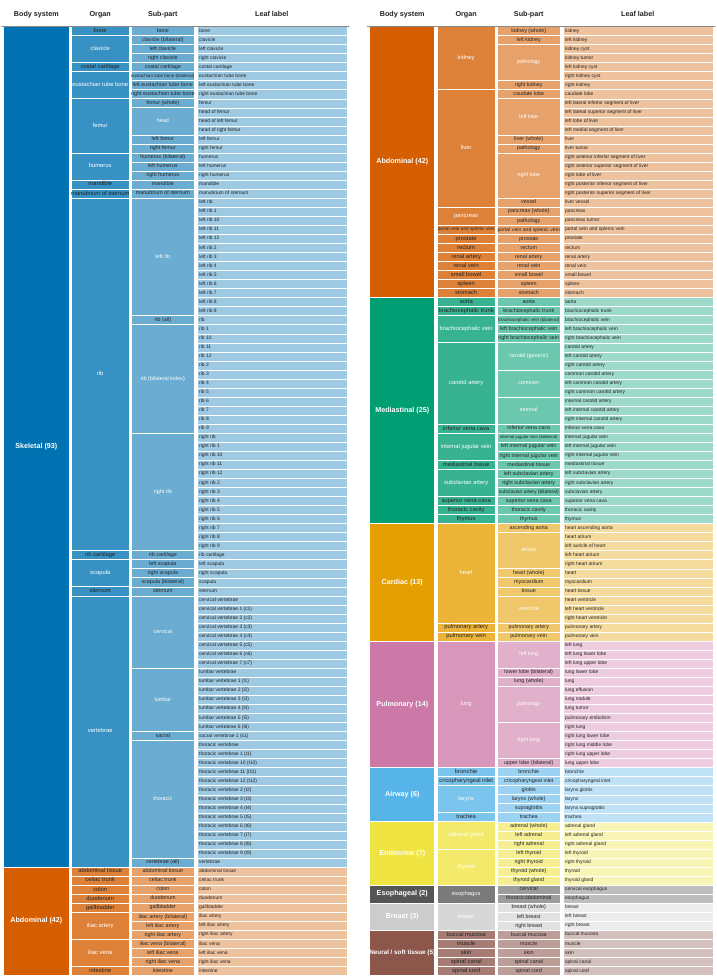}
  \caption{Four-level anatomical label hierarchy of \textit{RadGenome-Anatomy}. The taxonomy organizes 210 leaf-level anatomical labels into a structured hierarchy from body system to organ, anatomical sub-part, and final leaf label.}
  \label{fig:label_details}
\end{figure}

\begin{figure}[H]
  \centering
  \includegraphics[height=0.9\textheight,keepaspectratio]{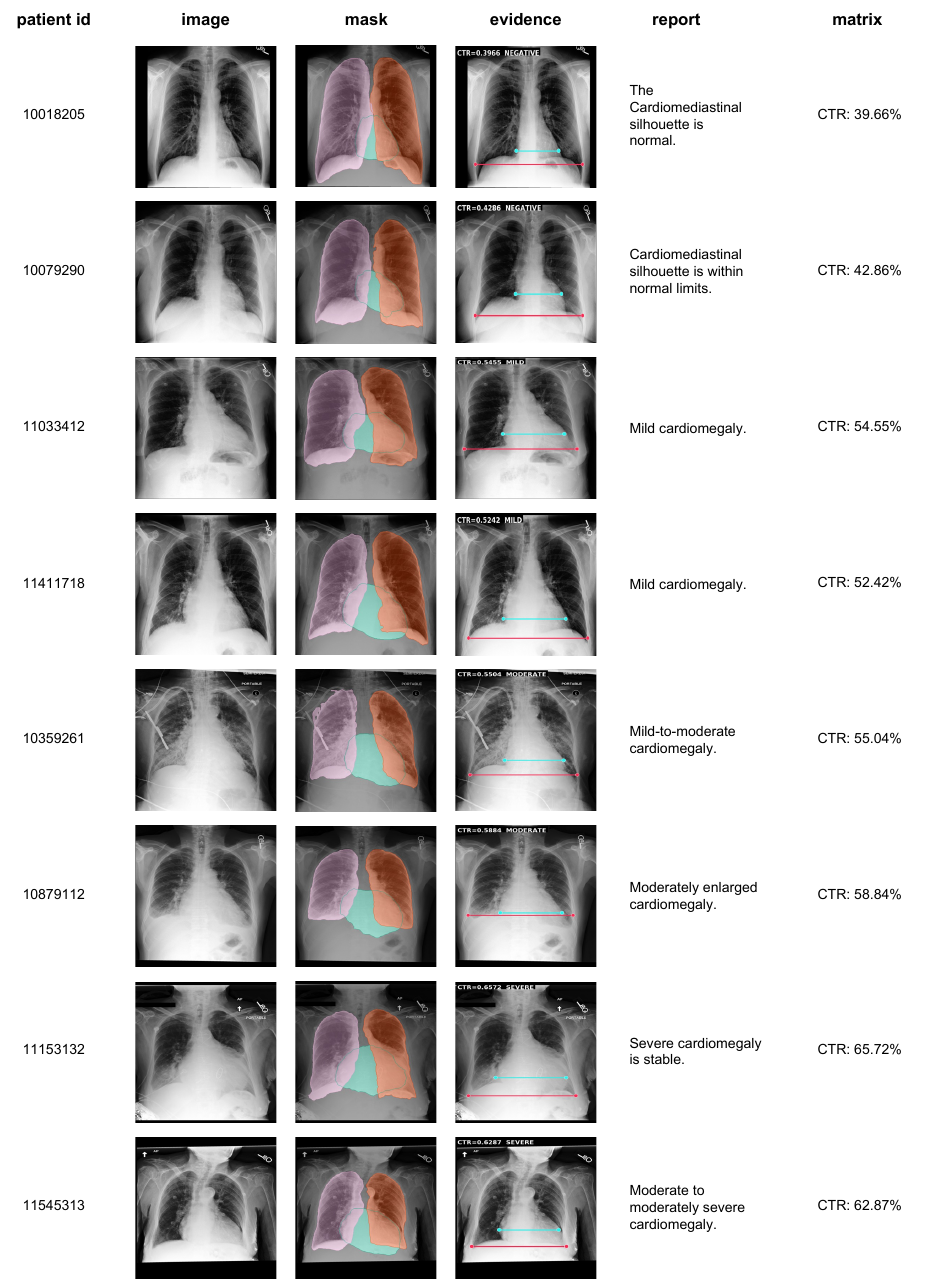}
  \caption{
    Qualitative examples of cardiomegaly measurements. For each study, columns show the patient identifier, PA chest radiograph, predicted cardiothoracic anatomy-mask overlay, measurement evidence, corresponding radiology-report description, and the computed cardiothoracic ratio (CTR). Rows span normal cardiomediastinal silhouette and mild-to-severe cardiomegaly, illustrating how predicted heart and thoracic-cage masks support explicit cardiothoracic measurement from PA chest radiographs.
    }
  \label{fig:label_details}
\end{figure}

\begin{figure}[H]
  \centering
  \includegraphics[height=0.9\textheight,keepaspectratio]{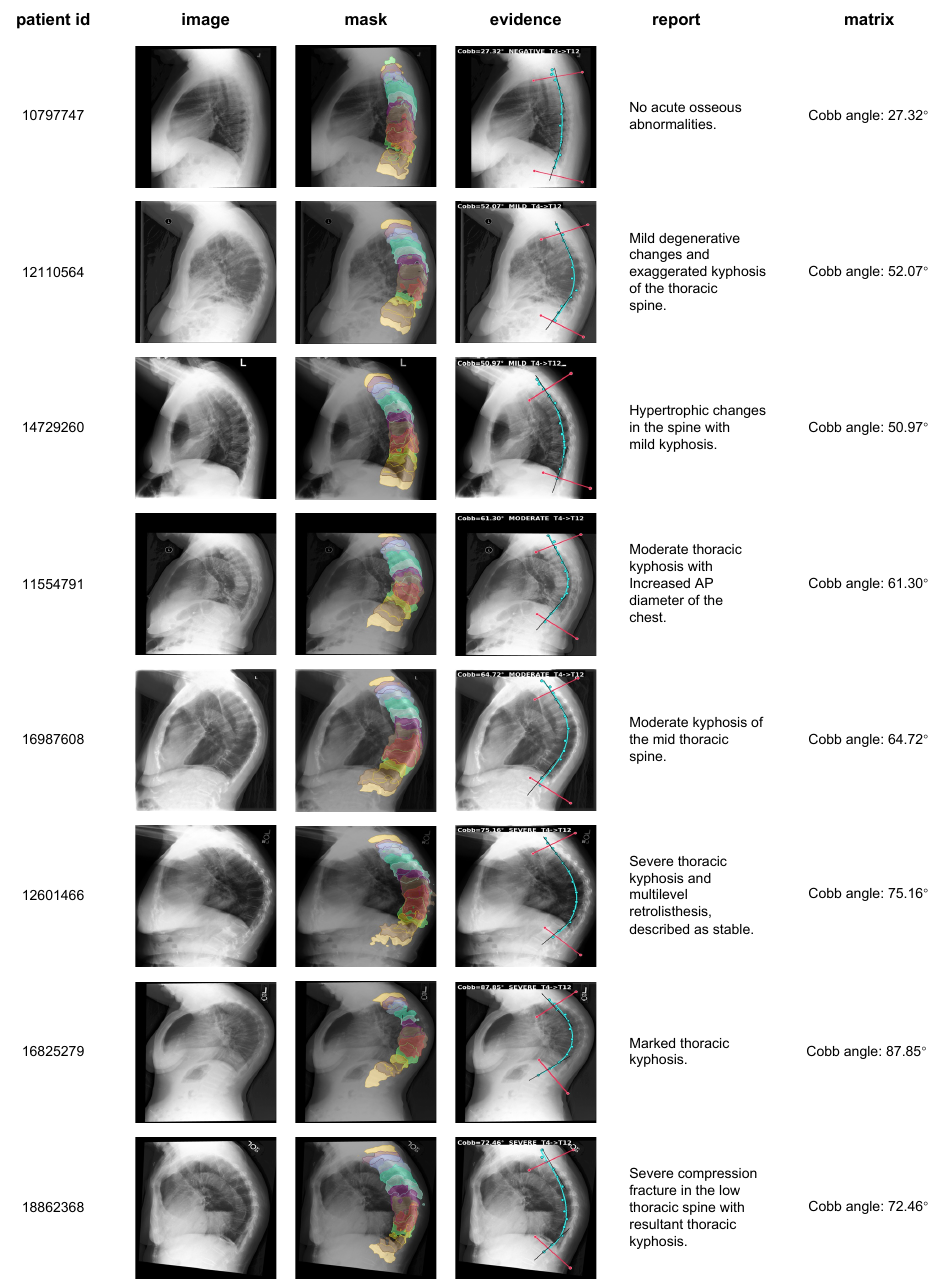}
  \caption{
    Qualitative examples of thoracic kyphosis measurements. For each study, columns show the patient identifier, lateral chest radiograph, predicted spinal anatomy-mask overlay, measurement evidence, corresponding radiology-report description, and the computed Cobb angle. Rows include normal or no-acute-osseous-abnormality cases and mild, moderate, marked, and severe thoracic kyphosis, illustrating how predicted vertebral anatomy supports explicit curvature estimation from lateral chest radiographs.
    }
  \label{fig:label_details}
\end{figure}

\begin{figure}[H]
  \centering
  \includegraphics[height=0.9\textheight,keepaspectratio]{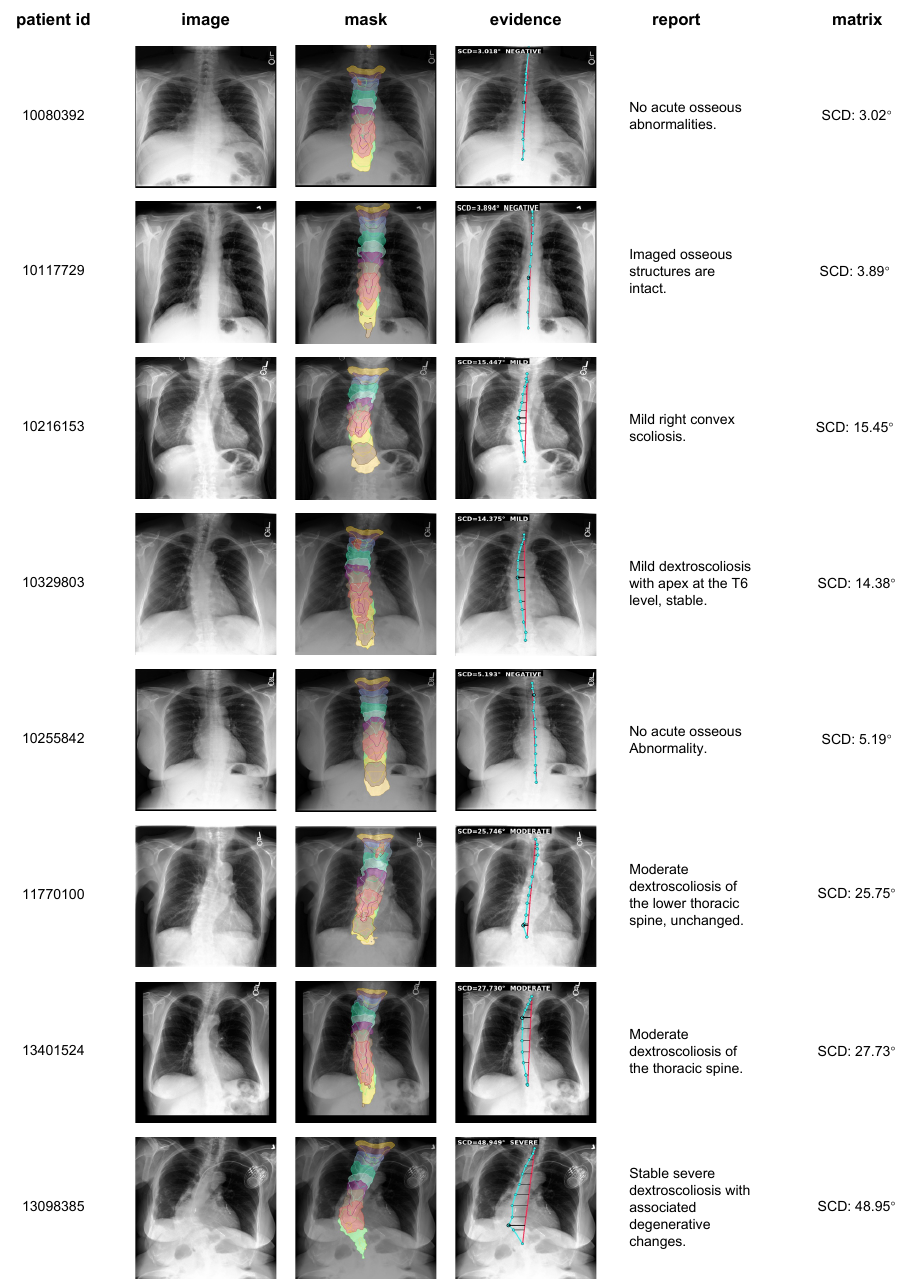}
  \caption{
    Qualitative examples of scoliosis measurements. For each study, columns show the patient identifier, PA chest radiograph, predicted anatomy-mask overlay, measurement evidence, corresponding radiology-report description, and the computed spine-curvature degree (SCD). Rows include normal or no-acute-osseous-abnormality cases, mild dextroscoliosis, moderate dextroscoliosis, and severe dextroscoliosis, illustrating how the predicted spinal anatomy supports explicit curvature estimation from chest radiographs.
    }
  \label{fig:label_details}
\end{figure}

\begin{figure}[H]
  \centering
  \includegraphics[height=0.85\textheight,keepaspectratio]{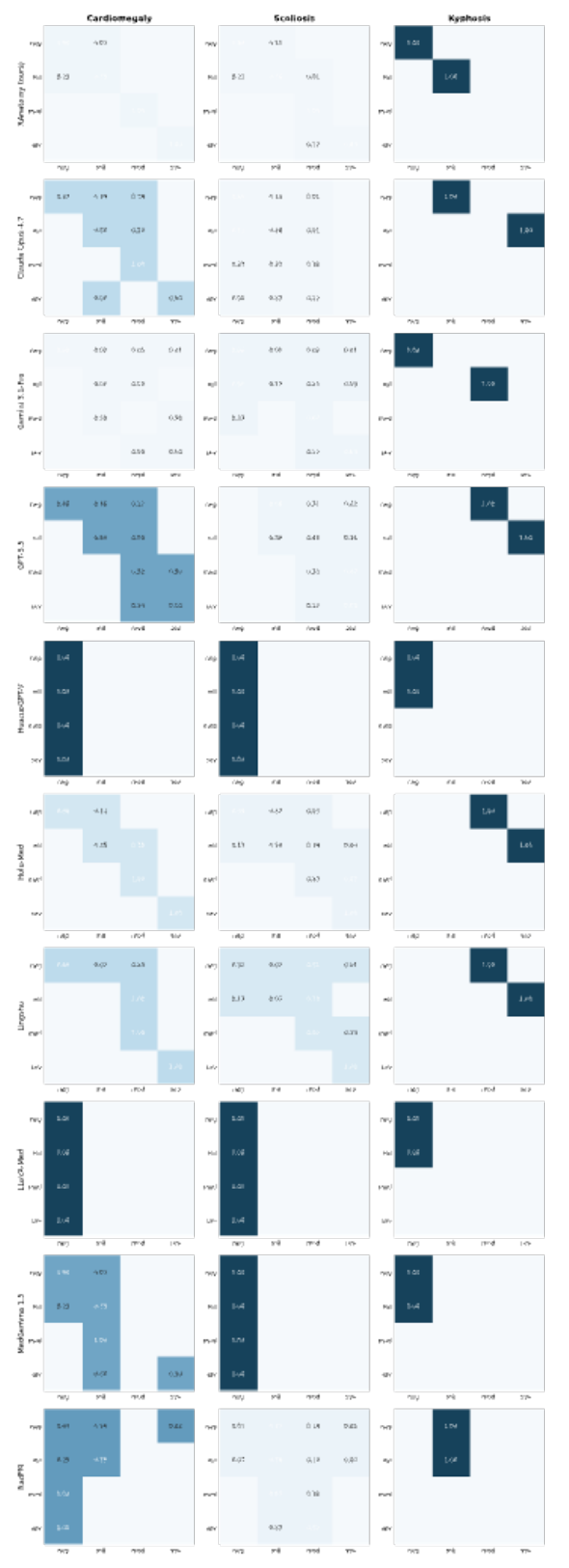}
  \caption{Severity-level confusion matrices for measurement-based diagnosis using \textit{XAnatomy}. The figure reports model behavior across cardiomegaly, scoliosis, and kyphosis, showing how predicted severity categories align with report-derived severity labels.}
  \label{fig:severity_confusion}
\end{figure}

\begin{figure}[H]
  \centering
  \includegraphics[width=\textwidth,keepaspectratio]{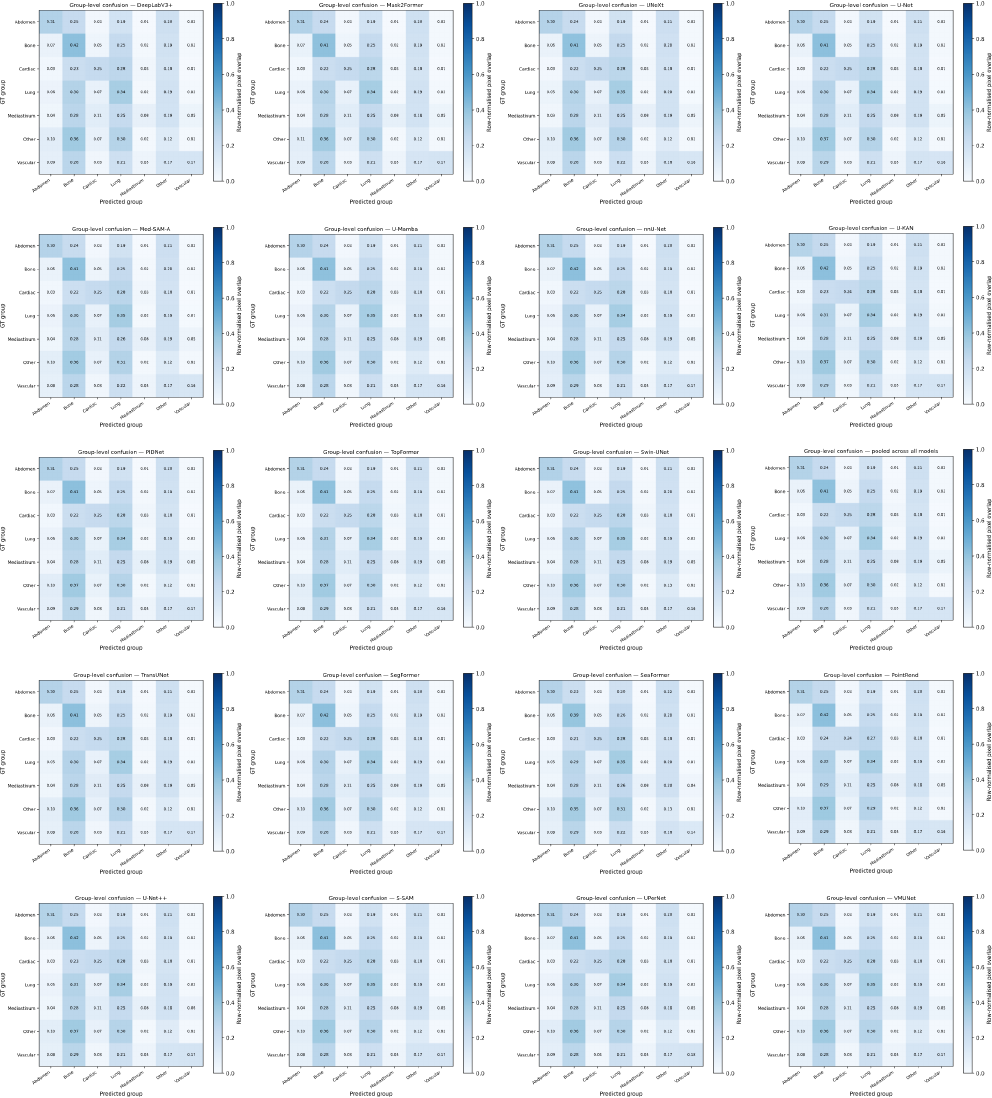}
  \caption{Grouped anatomical confusion matrices across segmentation models. The matrices summarize prediction patterns among major anatomical groups, providing a structured view of cross-group confusions beyond leaf-level segmentation scores.}
  \label{fig:group_confusion}
\end{figure}

\begin{figure}[H]
  \centering
  \includegraphics[width=\textwidth,keepaspectratio]{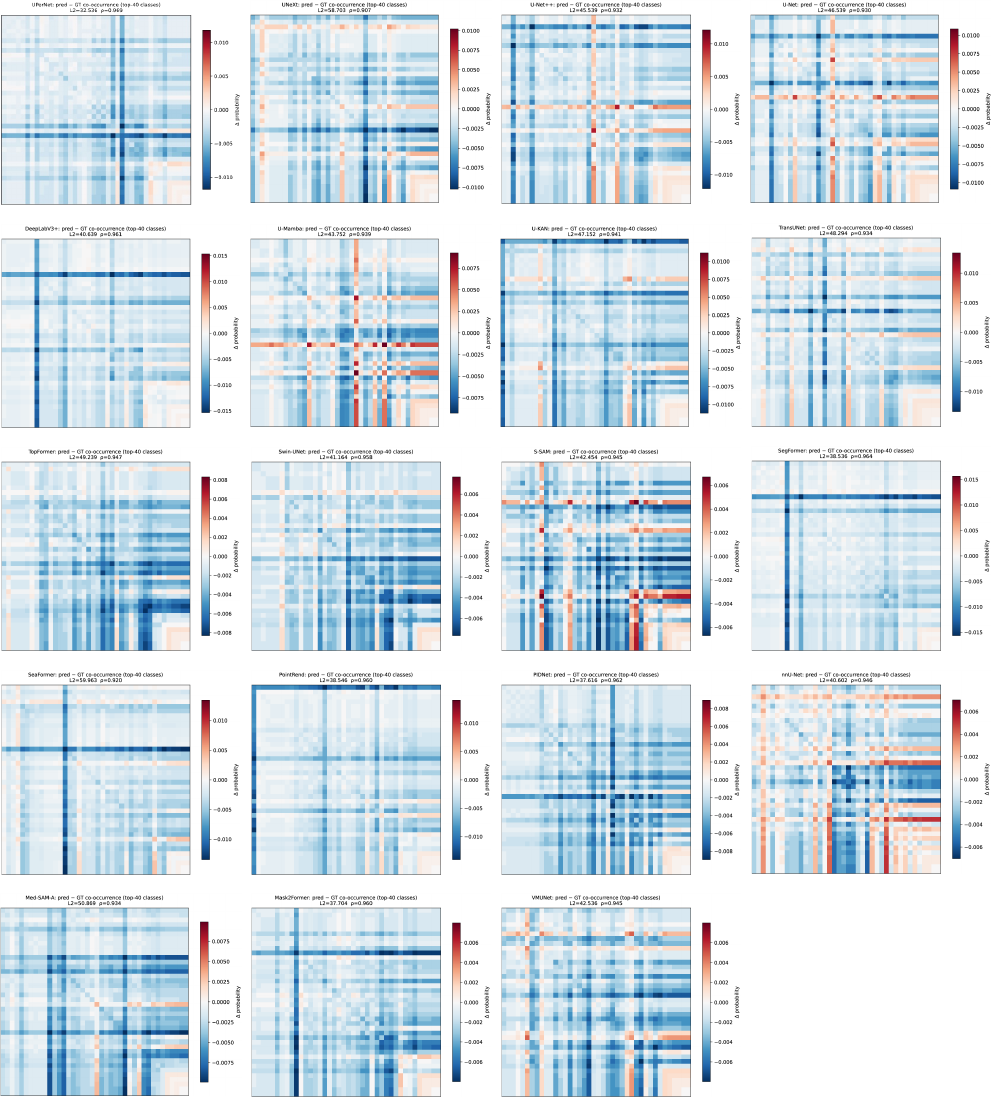}
  \caption{Anatomical label co-occurrence patterns across segmentation models. The heatmaps visualize pairwise co-occurrence differences among anatomical labels, revealing whether model predictions preserve the structural dependencies present in the projected anatomy annotations.}
  \label{fig:co_occurrence}
\end{figure}

\section{Appendix Tables}

\begin{table}[H]
  \centering
  \small
  \caption{Anatomy-mask size per body-system group. For each of the ten level-1 body-system groups we report the mean foreground-pixel count of every positive-area mask in the PA and LL projections, the matching $95\%$ bootstrap confidence interval on the mean, the fraction of the $384 \times 384$ projection image that the mean represents, and the PA/LL ratio. All values are computed from a seeded uniform sample of $500$ studies ($\approx\!1{,}000$ observations per label per view).}
  \label{tab:mask_details}
  \begin{tabular}{@{}lrrrrrr@{}}
    \toprule
                     & \multicolumn{2}{c}{PA}           & \multicolumn{2}{c}{LL}           &               &         \\
    \cmidrule(lr){2-3}\cmidrule(lr){4-5}
    Body system      & Mean (px) & $95\%$ CI            & Mean (px) & $95\%$ CI            & \% of image   & PA/LL   \\
    \midrule
    Neural / soft tissue & 58{,}652 & [56{,}279, 61{,}181] & 37{,}923 & [36{,}280, 39{,}465] & 39.8 / 25.7   & 1.55 \\
    Pulmonary            & 24{,}761 & [24{,}287, 25{,}230] & 21{,}608 & [21{,}214, 21{,}976] & 16.8 / 14.7   & 1.15 \\
    Breast               &  6{,}999 & [6{,}472,  7{,}518]  &  4{,}799 & [4{,}495,  5{,}103]  &  4.7 /  3.3   & 1.46 \\
    Cardiac              &  5{,}292 & [5{,}205,  5{,}380]  &  4{,}670 & [4{,}590,  4{,}743]  &  3.6 /  3.2   & 1.13 \\
    Abdominal            &  5{,}279 & [5{,}168,  5{,}399]  &  4{,}777 & [4{,}676,  4{,}883]  &  3.6 /  3.2   & 1.11 \\
    Skeletal             &  4{,}931 & [4{,}857,  5{,}010]  &  4{,}669 & [4{,}599,  4{,}738]  &  3.3 /  3.2   & 1.06 \\
    Mediastinal          &  4{,}818 & [4{,}569,  5{,}066]  &  3{,}832 & [3{,}661,  4{,}011]  &  3.3 /  2.6   & 1.26 \\
    Esophageal           &  2{,}537 & [2{,}382,  2{,}689]  &  1{,}988 & [1{,}876,  2{,}105]  &  1.7 /  1.3   & 1.28 \\
    Airway               &  1{,}814 & [1{,}741,  1{,}886]  &  1{,}536 & [1{,}479,  1{,}592]  &  1.2 /  1.0   & 1.18 \\
    Endocrine            &     831  & [813,      849]      &     817  & [804,      829]      &  0.56/ 0.55   & 1.02 \\
    \bottomrule
  \end{tabular}
\end{table}

\begin{table}[H]
\centering
\caption{Detailed multiclass breakdown of Figure~\ref{fig:measurement_qwk}. Rows indicate metrics and columns indicate models, with separate blocks for each disease. We report multiclass accuracy, off-by-one accuracy, macro-F1, weighted-F1, and linear- and quadratic-weighted Cohen's $\kappa$. Each cell shows the point estimate and 95\% bootstrap confidence interval. 
}
\label{tab:barplot-own-coverage-detail}
\vspace{4mm}
\setlength{\tabcolsep}{3pt}
\renewcommand{\arraystretch}{1.05}
\resizebox{\linewidth}{!}{%
\begin{tabular}{@{}lllllllllll@{}}
\toprule
Metric & \textbf{Ours} & Claude & Gemini & GPT & HuatuoGPT & Hulu-Med & Lingshu & LLaVA-Med & MedGemma & RadFM \\
\midrule
\multicolumn{11}{@{}l@{}}{\textbf{Cardiomegaly} \hfill \scriptsize Ours: $n$=1287 \, Claude: $n$=1287 \, Gemini: $n$=1287 \, GPT: $n$=1287 \, HuatuoGPT: $n$=1287 \, Hulu-Med: $n$=1287 \, Lingshu: $n$=1287 \, LLaVA-Med: $n$=152 \, MedGemma: $n$=1285 \, RadFM: $n$=801} \\
\addlinespace[1pt]
Acc. & \textbf{0.91 [0.89, 0.92]} & 0.38 [0.36, 0.41] & 0.72 [0.70, 0.75] & 0.30 [0.28, 0.33] & 0.71 [0.69, 0.74] & 0.67 [0.65, 0.70] & 0.54 [0.51, 0.57] & 0.77 [0.70, 0.84] & 0.82 [0.80, 0.84] & 0.29 [0.26, 0.32] \\
Off-by-1 & \textbf{1.00} & 0.83 & 0.89 & 0.81 & 0.82 & 0.95 & 0.78 & 0.83 & 0.95 & 0.93 \\
Macro-F1 & \textbf{0.80 [0.77, 0.83]} & 0.30 [0.27, 0.33] & 0.46 [0.42, 0.49] & 0.32 [0.29, 0.34] & 0.83 [0.82, 0.85] & 0.54 [0.51, 0.57] & 0.41 [0.39, 0.44] & 0.87 [0.83, 0.91] & 0.54 [0.51, 0.71] & 0.30 [0.25, 0.35] \\
W-F1 & \textbf{0.91 [0.89, 0.92]} & 0.46 [0.43, 0.48] & 0.72 [0.69, 0.74] & 0.37 [0.34, 0.40] & 0.59 [0.56, 0.63] & 0.72 [0.70, 0.74] & 0.61 [0.58, 0.63] & 0.67 [0.58, 0.76] & 0.80 [0.78, 0.82] & 0.35 [0.31, 0.39] \\
Linear $\kappa$ & \textbf{0.89 [0.87, 0.91]} & 0.24 [0.21, 0.28] & 0.57 [0.53, 0.60] & 0.30 [0.27, 0.33] & 0.00 [0.00, 0.01] & 0.64 [0.61, 0.67] & 0.44 [0.41, 0.47] & 0.00 [0.00, 0.00] & 0.70 [0.67, 0.74] & 0.02 [-0.04, 0.08] \\
Quad.\ $\kappa$ & \textbf{0.95 [0.94, 0.96]} & 0.37 [0.32, 0.41] & 0.68 [0.63, 0.71] & 0.46 [0.42, 0.49] & 0.01 [0.00, 0.02] & 0.79 [0.76, 0.82] & 0.56 [0.53, 0.60] & 0.00 [0.00, 0.00] & 0.80 [0.76, 0.83] & 0.12 [0.01, 0.22] \\
\midrule
\multicolumn{11}{@{}l@{}}{\textbf{Scoliosis} \hfill \scriptsize Ours: $n$=809 \, Claude: $n$=809 \, Gemini: $n$=809 \, GPT: $n$=809 \, HuatuoGPT: $n$=809 \, Hulu-Med: $n$=809 \, Lingshu: $n$=809 \, LLaVA-Med: $n$=541 \, MedGemma: $n$=808 \, RadFM: $n$=415} \\
\addlinespace[1pt]
Acc. & \textbf{0.89 [0.86, 0.91]} & 0.69 [0.66, 0.72] & 0.77 [0.74, 0.79] & 0.11 [0.09, 0.13] & 0.74 [0.71, 0.76] & 0.57 [0.54, 0.60] & 0.24 [0.21, 0.27] & 0.72 [0.68, 0.76] & 0.74 [0.71, 0.77] & 0.20 [0.15, 0.23] \\
Off-by-1 & \textbf{1.00} & 0.98 & 0.98 & 0.74 & 0.97 & 0.95 & 0.54 & 0.98 & 0.98 & 0.76 \\
Macro-F1 & \textbf{0.85 [0.77, 0.91]} & 0.41 [0.39, 0.58] & 0.49 [0.41, 0.55] & 0.14 [0.10, 0.24] & 0.85 [0.83, 0.87] & 0.41 [0.35, 0.55] & 0.27 [0.21, 0.32] & 0.84 [0.81, 0.86] & 0.85 [0.83, 0.87] & 0.13 [0.11, 0.19] \\
W-F1 & \textbf{0.89 [0.87, 0.91]} & 0.67 [0.63, 0.70] & 0.72 [0.69, 0.76] & 0.06 [0.05, 0.07] & 0.62 [0.59, 0.66] & 0.63 [0.60, 0.66] & 0.34 [0.30, 0.38] & 0.60 [0.56, 0.66] & 0.62 [0.59, 0.66] & 0.11 [0.08, 0.15] \\
Linear $\kappa$ & \textbf{0.76 [0.72, 0.81]} & 0.17 [0.10, 0.23] & 0.41 [0.34, 0.48] & 0.03 [0.01, 0.05] & 0.00 [0.00, 0.00] & 0.31 [0.26, 0.36] & 0.07 [0.04, 0.09] & 0.00 [0.00, 0.00] & 0.00 [0.00, 0.00] & 0.01 [-0.01, 0.04] \\
Quad.\ $\kappa$ & \textbf{0.83 [0.78, 0.87]} & 0.21 [0.12, 0.30] & 0.54 [0.44, 0.62] & 0.13 [0.10, 0.16] & 0.00 [0.00, 0.00] & 0.44 [0.37, 0.50] & 0.10 [0.07, 0.13] & 0.00 [0.00, 0.00] & 0.00 [0.00, 0.00] & 0.01 [-0.03, 0.06] \\
\midrule
\multicolumn{11}{@{}l@{}}{\textbf{Kyphosis} \hfill \scriptsize Ours: $n$=191 \, Claude: $n$=191 \, Gemini: $n$=191 \, GPT: $n$=191 \, HuatuoGPT: $n$=191 \, Hulu-Med: $n$=191 \, Lingshu: $n$=191 \, LLaVA-Med: $n$=2 \, MedGemma: $n$=191 \, RadFM: $n$=186} \\
\addlinespace[1pt]
Acc. & \textbf{0.93 [0.89, 0.96]} & 0.13 [0.08, 0.17] & 0.70 [0.63, 0.76] & 0.09 [0.06, 0.14] & 0.77 [0.71, 0.83] & 0.13 [0.08, 0.17] & 0.10 [0.06, 0.15] & 0.50 [0.00, 1.00] & 0.82 [0.76, 0.87] & 0.28 [0.23, 0.35] \\
Off-by-1 & \textbf{1.00} & 0.80 & 0.85 & 0.56 & 0.86 & 0.59 & 0.17 & 1.00 & 0.88 & 0.86 \\
Macro-F1 & \textbf{0.79 [0.67, 0.88]} & 0.19 [0.12, 0.30] & 0.56 [0.49, 0.77] & 0.22 [0.16, 0.41] & 0.89 [0.85, 0.92] & 0.26 [0.22, 0.43] & 0.33 [0.26, 0.75] & 0.67 [0.67, 1.00] & 0.90 [0.87, 0.93] & 0.28 [0.22, 0.34] \\
W-F1 & \textbf{0.93 [0.90, 0.96]} & 0.15 [0.09, 0.22] & 0.76 [0.70, 0.82] & 0.06 [0.03, 0.10] & 0.73 [0.65, 0.80] & 0.11 [0.06, 0.17] & 0.06 [0.03, 0.10] & 0.33 [0.00, 1.00] & 0.73 [0.66, 0.81] & 0.35 [0.27, 0.43] \\
Linear $\kappa$ & \textbf{0.90 [0.83, 0.94]} & 0.15 [0.08, 0.22] & 0.48 [0.34, 0.59] & 0.13 [0.08, 0.18] & 0.14 [0.01, 0.29] & 0.17 [0.11, 0.23] & 0.08 [0.05, 0.11] & 0.00 [0.00, 0.00] & 0.00 [0.00, 0.00] & -0.02 [-0.08, 0.04] \\
Quad.\ $\kappa$ & \textbf{0.96 [0.93, 0.98]} & 0.29 [0.18, 0.39] & 0.61 [0.45, 0.72] & 0.22 [0.15, 0.29] & 0.18 [0.01, 0.36] & 0.29 [0.20, 0.36] & 0.13 [0.09, 0.17] & 0.00 [0.00, 0.00] & 0.00 [0.00, 0.00] & -0.06 [-0.16, 0.05] \\
\bottomrule
\end{tabular}%
}
\end{table}

\begin{table*}
\centering
\caption{Per-class mean Dice and mIoU for the 69 L3 sub-segment anatomy classes across 19 segmentation models, decomposed by view. Anatomy is broken into three hierarchy columns (L1 organ system $\rightarrow$ L2 sub-region $\rightarrow$ L3 leaf segment); rows are grouped per L3 leaf, with a Dice row and an mIoU row each. Columns under \textbf{PA} use frontal-only cases; columns under \textbf{LL} use lateral-only cases. Anatomy order follows the x-axis of the hierarchical metric Figures~\ref{fig:classes_performance}. ``--'' marks classes with no positive cases for that view.}
\label{tab:hierarchical-dice-detail}
\vspace{4mm}
\setlength{\tabcolsep}{2pt}
\renewcommand{\arraystretch}{1.0}
\resizebox{\linewidth}{!}{%
\begin{tabular}{@{}lll l ccccccccccccccccccc ccccccccccccccccccc@{}}
\toprule
 &  &  &  & \multicolumn{19}{c}{\textbf{PA (frontal)}} & \multicolumn{19}{c}{\textbf{LL (lateral)}} \\
\cmidrule(lr){5-23} \cmidrule(lr){24-42}
L1 & L2 & L3 & Metric & DLv3+ & M2F & MSA & nnU & PID & PR & SeaF & SegF & SSAM & SwU & TopF & TrU & UKAN & UMa & UNet & UN++ & UNX & UPN & VMU & DLv3+ & M2F & MSA & nnU & PID & PR & SeaF & SegF & SSAM & SwU & TopF & TrU & UKAN & UMa & UNet & UN++ & UNX & UPN & VMU \\
\midrule
\multicolumn{42}{l}{\textit{Respiratory}} \\
lung & left lung & left lung upper lobe & Dice & 0.939 & 0.941 & 0.937 & 0.938 & 0.938 & 0.936 & 0.927 & 0.942 & 0.938 & 0.940 & 0.934 & 0.937 & 0.936 & 0.937 & 0.935 & 0.939 & 0.928 & 0.941 & 0.938 & 0.917 & 0.919 & 0.909 & 0.911 & 0.915 & 0.913 & 0.898 & 0.921 & 0.911 & 0.913 & 0.908 & 0.911 & 0.910 & 0.911 & 0.910 & 0.912 & 0.902 & 0.917 & 0.912 \\
 &  &  & mIoU & 0.887 & 0.891 & 0.884 & 0.885 & 0.885 & 0.883 & 0.866 & 0.893 & 0.885 & 0.890 & 0.879 & 0.883 & 0.883 & 0.884 & 0.881 & 0.887 & 0.868 & 0.891 & 0.886 & 0.850 & 0.853 & 0.837 & 0.840 & 0.846 & 0.843 & 0.819 & 0.857 & 0.840 & 0.844 & 0.834 & 0.840 & 0.839 & 0.841 & 0.837 & 0.842 & 0.825 & 0.849 & 0.841 \\
 &  & left lung lower lobe & Dice & 0.939 & 0.942 & 0.939 & 0.937 & 0.937 & 0.936 & 0.928 & 0.943 & 0.937 & 0.940 & 0.934 & 0.936 & 0.936 & 0.936 & 0.937 & 0.939 & 0.931 & 0.941 & 0.937 & 0.915 & 0.916 & 0.902 & 0.909 & 0.912 & 0.910 & 0.891 & 0.920 & 0.909 & 0.909 & 0.905 & 0.905 & 0.905 & 0.909 & 0.901 & 0.906 & 0.891 & 0.914 & 0.909 \\
 &  &  & mIoU & 0.888 & 0.893 & 0.887 & 0.885 & 0.885 & 0.884 & 0.869 & 0.895 & 0.884 & 0.889 & 0.879 & 0.883 & 0.883 & 0.883 & 0.884 & 0.888 & 0.874 & 0.892 & 0.885 & 0.848 & 0.849 & 0.827 & 0.838 & 0.843 & 0.840 & 0.809 & 0.856 & 0.838 & 0.839 & 0.831 & 0.831 & 0.832 & 0.838 & 0.826 & 0.833 & 0.809 & 0.847 & 0.839 \\
 & right lung & right lung upper lobe & Dice & 0.916 & 0.919 & 0.911 & 0.914 & 0.913 & 0.913 & 0.902 & 0.921 & 0.913 & 0.917 & 0.910 & 0.913 & 0.911 & 0.913 & 0.911 & 0.915 & 0.904 & 0.918 & 0.913 & 0.897 & 0.898 & 0.887 & 0.889 & 0.895 & 0.892 & 0.878 & 0.902 & 0.889 & 0.894 & 0.888 & 0.888 & 0.889 & 0.889 & 0.887 & 0.890 & 0.879 & 0.898 & 0.890 \\
 &  &  & mIoU & 0.851 & 0.855 & 0.843 & 0.847 & 0.846 & 0.845 & 0.827 & 0.858 & 0.846 & 0.853 & 0.841 & 0.845 & 0.843 & 0.845 & 0.842 & 0.849 & 0.830 & 0.854 & 0.846 & 0.820 & 0.822 & 0.804 & 0.806 & 0.817 & 0.812 & 0.790 & 0.828 & 0.807 & 0.815 & 0.805 & 0.805 & 0.806 & 0.807 & 0.803 & 0.808 & 0.790 & 0.822 & 0.808 \\
 &  & right lung middle lobe & Dice & 0.888 & 0.895 & 0.875 & 0.882 & 0.886 & 0.883 & 0.857 & 0.896 & 0.883 & 0.884 & 0.879 & 0.870 & 0.878 & 0.882 & 0.871 & 0.879 & 0.843 & 0.892 & 0.882 & 0.875 & 0.879 & 0.857 & 0.869 & 0.874 & 0.871 & 0.842 & 0.882 & 0.869 & 0.869 & 0.862 & 0.852 & 0.865 & 0.868 & 0.851 & 0.864 & 0.823 & 0.879 & 0.869 \\
 &  &  & mIoU & 0.806 & 0.817 & 0.787 & 0.799 & 0.805 & 0.799 & 0.759 & 0.820 & 0.799 & 0.801 & 0.792 & 0.780 & 0.791 & 0.798 & 0.781 & 0.794 & 0.740 & 0.812 & 0.798 & 0.786 & 0.792 & 0.758 & 0.778 & 0.784 & 0.781 & 0.736 & 0.797 & 0.778 & 0.777 & 0.767 & 0.752 & 0.772 & 0.776 & 0.751 & 0.770 & 0.710 & 0.791 & 0.776 \\
 &  & right lung lower lobe & Dice & 0.939 & 0.942 & 0.936 & 0.937 & 0.937 & 0.937 & 0.926 & 0.943 & 0.936 & 0.938 & 0.933 & 0.936 & 0.936 & 0.937 & 0.936 & 0.939 & 0.928 & 0.941 & 0.936 & 0.919 & 0.922 & 0.908 & 0.915 & 0.918 & 0.917 & 0.898 & 0.926 & 0.916 & 0.916 & 0.911 & 0.913 & 0.911 & 0.915 & 0.911 & 0.915 & 0.901 & 0.921 & 0.915 \\
 &  &  & mIoU & 0.889 & 0.893 & 0.883 & 0.885 & 0.886 & 0.885 & 0.867 & 0.896 & 0.884 & 0.888 & 0.878 & 0.884 & 0.883 & 0.885 & 0.884 & 0.889 & 0.870 & 0.891 & 0.884 & 0.855 & 0.859 & 0.836 & 0.848 & 0.852 & 0.851 & 0.820 & 0.865 & 0.849 & 0.849 & 0.841 & 0.843 & 0.842 & 0.847 & 0.841 & 0.847 & 0.824 & 0.858 & 0.848 \\
\midrule
\multicolumn{42}{l}{\textit{Bones}} \\
vertebrae & cervical vertebrae & cervical vertebrae 1 (c1) & Dice & 0.270 & 0.308 & 0.068 & 0.223 & 0.186 & 0.282 & 0.122 & 0.312 & 0.205 & 0.277 & 0.156 & 0.158 & 0.221 & 0.316 & 0.203 & 0.237 & 0.128 & 0.338 & 0.195 & 0.126 & 0.198 & 0.007 & 0.117 & 0.109 & 0.047 & 0.087 & 0.169 & 0.098 & 0.130 & 0.111 & 0.118 & 0.024 & 0.124 & 0.136 & 0.095 & 0.008 & 0.223 & 0.127 \\
 &  &  & mIoU & 0.175 & 0.212 & 0.040 & 0.144 & 0.124 & 0.188 & 0.080 & 0.211 & 0.131 & 0.188 & 0.101 & 0.098 & 0.146 & 0.212 & 0.130 & 0.156 & 0.077 & 0.232 & 0.123 & 0.080 & 0.129 & 0.004 & 0.074 & 0.072 & 0.029 & 0.054 & 0.111 & 0.061 & 0.082 & 0.070 & 0.074 & 0.015 & 0.079 & 0.085 & 0.059 & 0.006 & 0.148 & 0.078 \\
 &  & cervical vertebrae 2 (c2) & Dice & 0.065 & 0.082 & 0.006 & 0.050 & 0.067 & 0.062 & 0.023 & 0.081 & 0.039 & 0.049 & 0.054 & 0.056 & 0.054 & 0.048 & 0.047 & 0.064 & 0.032 & 0.090 & 0.048 & 0.059 & 0.074 & 0.011 & 0.050 & 0.059 & 0.055 & 0.022 & 0.073 & 0.050 & 0.046 & 0.050 & 0.055 & 0.040 & 0.045 & 0.054 & 0.064 & 0.031 & 0.099 & 0.047 \\
 &  &  & mIoU & 0.052 & 0.062 & 0.003 & 0.037 & 0.053 & 0.046 & 0.015 & 0.063 & 0.029 & 0.037 & 0.041 & 0.044 & 0.041 & 0.034 & 0.034 & 0.050 & 0.023 & 0.068 & 0.035 & 0.046 & 0.058 & 0.006 & 0.038 & 0.044 & 0.042 & 0.014 & 0.059 & 0.037 & 0.033 & 0.038 & 0.042 & 0.029 & 0.033 & 0.042 & 0.049 & 0.021 & 0.076 & 0.034 \\
 &  & cervical vertebrae 3 (c3) & Dice & 0.284 & 0.283 & 0.157 & 0.251 & 0.264 & 0.258 & 0.114 & 0.282 & 0.201 & 0.242 & 0.181 & 0.218 & 0.195 & 0.212 & 0.204 & 0.233 & 0.131 & 0.259 & 0.213 & 0.246 & 0.266 & 0.088 & 0.232 & 0.214 & 0.194 & 0.052 & 0.265 & 0.177 & 0.213 & 0.153 & 0.156 & 0.194 & 0.175 & 0.174 & 0.202 & 0.099 & 0.268 & 0.178 \\
 &  &  & mIoU & 0.211 & 0.211 & 0.102 & 0.180 & 0.196 & 0.189 & 0.077 & 0.211 & 0.141 & 0.174 & 0.128 & 0.154 & 0.134 & 0.149 & 0.144 & 0.167 & 0.087 & 0.192 & 0.150 & 0.179 & 0.194 & 0.055 & 0.166 & 0.153 & 0.136 & 0.034 & 0.196 & 0.119 & 0.151 & 0.106 & 0.101 & 0.138 & 0.119 & 0.120 & 0.141 & 0.064 & 0.197 & 0.120 \\
 &  & cervical vertebrae 4 (c4) & Dice & 0.213 & 0.268 & 0.048 & 0.197 & 0.225 & 0.228 & 0.058 & 0.267 & 0.140 & 0.189 & 0.133 & 0.180 & 0.089 & 0.146 & 0.172 & 0.196 & 0.052 & 0.258 & 0.159 & 0.195 & 0.237 & 0.018 & 0.184 & 0.184 & 0.126 & 0.030 & 0.235 & 0.134 & 0.175 & 0.091 & 0.134 & 0.100 & 0.134 & 0.139 & 0.169 & 0.031 & 0.241 & 0.131 \\
 &  &  & mIoU & 0.152 & 0.195 & 0.029 & 0.137 & 0.163 & 0.164 & 0.039 & 0.197 & 0.095 & 0.133 & 0.090 & 0.125 & 0.058 & 0.101 & 0.118 & 0.136 & 0.034 & 0.190 & 0.111 & 0.140 & 0.172 & 0.011 & 0.131 & 0.132 & 0.088 & 0.019 & 0.170 & 0.090 & 0.125 & 0.061 & 0.093 & 0.066 & 0.092 & 0.094 & 0.119 & 0.018 & 0.174 & 0.088 \\
 &  & cervical vertebrae 5 (c5) & Dice & 0.299 & 0.336 & 0.236 & 0.289 & 0.298 & 0.300 & 0.187 & 0.333 & 0.275 & 0.282 & 0.260 & 0.270 & 0.267 & 0.272 & 0.288 & 0.304 & 0.239 & 0.339 & 0.276 & 0.201 & 0.212 & 0.059 & 0.175 & 0.200 & 0.145 & 0.051 & 0.220 & 0.174 & 0.171 & 0.136 & 0.161 & 0.100 & 0.162 & 0.160 & 0.166 & 0.038 & 0.239 & 0.168 \\
 &  &  & mIoU & 0.222 & 0.252 & 0.169 & 0.212 & 0.224 & 0.221 & 0.132 & 0.251 & 0.203 & 0.208 & 0.194 & 0.200 & 0.197 & 0.203 & 0.212 & 0.226 & 0.174 & 0.257 & 0.204 & 0.142 & 0.149 & 0.038 & 0.120 & 0.142 & 0.103 & 0.033 & 0.158 & 0.120 & 0.117 & 0.095 & 0.113 & 0.068 & 0.111 & 0.111 & 0.116 & 0.025 & 0.170 & 0.115 \\
 &  & cervical vertebrae 6 (c6) & Dice & 0.556 & 0.585 & 0.538 & 0.545 & 0.544 & 0.549 & 0.465 & 0.583 & 0.542 & 0.553 & 0.532 & 0.536 & 0.564 & 0.558 & 0.546 & 0.544 & 0.516 & 0.579 & 0.547 & 0.479 & 0.471 & 0.347 & 0.410 & 0.437 & 0.388 & 0.258 & 0.479 & 0.429 & 0.445 & 0.375 & 0.405 & 0.368 & 0.432 & 0.419 & 0.396 & 0.245 & 0.495 & 0.436 \\
 &  &  & mIoU & 0.443 & 0.470 & 0.421 & 0.431 & 0.436 & 0.437 & 0.357 & 0.468 & 0.428 & 0.438 & 0.417 & 0.420 & 0.448 & 0.444 & 0.432 & 0.431 & 0.401 & 0.465 & 0.431 & 0.361 & 0.357 & 0.246 & 0.300 & 0.331 & 0.289 & 0.180 & 0.366 & 0.318 & 0.329 & 0.270 & 0.298 & 0.266 & 0.317 & 0.304 & 0.288 & 0.172 & 0.377 & 0.322 \\
 &  & cervical vertebrae 7 (c7) & Dice & 0.798 & 0.803 & 0.782 & 0.786 & 0.795 & 0.793 & 0.743 & 0.807 & 0.787 & 0.789 & 0.772 & 0.788 & 0.784 & 0.790 & 0.790 & 0.794 & 0.768 & 0.804 & 0.787 & 0.749 & 0.755 & 0.707 & 0.728 & 0.747 & 0.706 & 0.604 & 0.755 & 0.733 & 0.731 & 0.690 & 0.719 & 0.725 & 0.735 & 0.730 & 0.738 & 0.660 & 0.751 & 0.733 \\
 &  &  & mIoU & 0.682 & 0.691 & 0.660 & 0.668 & 0.681 & 0.677 & 0.613 & 0.696 & 0.669 & 0.672 & 0.648 & 0.669 & 0.664 & 0.672 & 0.672 & 0.677 & 0.642 & 0.692 & 0.668 & 0.616 & 0.622 & 0.564 & 0.592 & 0.615 & 0.567 & 0.457 & 0.624 & 0.599 & 0.595 & 0.545 & 0.581 & 0.589 & 0.600 & 0.592 & 0.604 & 0.513 & 0.621 & 0.598 \\
 & thoracic vertebrae & thoracic vertebrae 1 (t1) & Dice & 0.850 & 0.857 & 0.838 & 0.841 & 0.852 & 0.845 & 0.788 & 0.859 & 0.845 & 0.844 & 0.829 & 0.842 & 0.837 & 0.845 & 0.843 & 0.842 & 0.821 & 0.858 & 0.844 & 0.814 & 0.821 & 0.789 & 0.805 & 0.818 & 0.778 & 0.712 & 0.821 & 0.801 & 0.799 & 0.767 & 0.805 & 0.790 & 0.802 & 0.798 & 0.806 & 0.750 & 0.822 & 0.802 \\
 &  &  & mIoU & 0.745 & 0.755 & 0.727 & 0.732 & 0.748 & 0.739 & 0.658 & 0.759 & 0.737 & 0.738 & 0.715 & 0.734 & 0.726 & 0.738 & 0.735 & 0.733 & 0.704 & 0.757 & 0.736 & 0.692 & 0.702 & 0.659 & 0.680 & 0.699 & 0.644 & 0.562 & 0.703 & 0.676 & 0.674 & 0.631 & 0.681 & 0.660 & 0.677 & 0.670 & 0.682 & 0.608 & 0.704 & 0.677 \\
 &  & thoracic vertebrae 2 (t2) & Dice & 0.835 & 0.844 & 0.821 & 0.824 & 0.835 & 0.825 & 0.767 & 0.845 & 0.826 & 0.830 & 0.810 & 0.826 & 0.815 & 0.825 & 0.817 & 0.823 & 0.788 & 0.845 & 0.826 & 0.804 & 0.813 & 0.767 & 0.795 & 0.807 & 0.777 & 0.697 & 0.814 & 0.787 & 0.789 & 0.763 & 0.780 & 0.772 & 0.786 & 0.785 & 0.796 & 0.714 & 0.813 & 0.789 \\
 &  &  & mIoU & 0.721 & 0.734 & 0.701 & 0.705 & 0.721 & 0.707 & 0.629 & 0.735 & 0.708 & 0.714 & 0.686 & 0.709 & 0.693 & 0.708 & 0.696 & 0.703 & 0.657 & 0.735 & 0.708 & 0.678 & 0.689 & 0.629 & 0.666 & 0.681 & 0.641 & 0.543 & 0.691 & 0.655 & 0.659 & 0.623 & 0.646 & 0.635 & 0.654 & 0.652 & 0.667 & 0.564 & 0.690 & 0.658 \\
 &  & thoracic vertebrae 3 (t3) & Dice & 0.829 & 0.836 & 0.805 & 0.812 & 0.822 & 0.809 & 0.744 & 0.837 & 0.812 & 0.820 & 0.795 & 0.811 & 0.804 & 0.817 & 0.805 & 0.820 & 0.764 & 0.837 & 0.807 & 0.795 & 0.807 & 0.749 & 0.781 & 0.797 & 0.772 & 0.686 & 0.806 & 0.773 & 0.777 & 0.745 & 0.770 & 0.769 & 0.778 & 0.776 & 0.786 & 0.711 & 0.807 & 0.776 \\
 &  &  & mIoU & 0.712 & 0.722 & 0.678 & 0.689 & 0.702 & 0.684 & 0.600 & 0.724 & 0.689 & 0.699 & 0.665 & 0.686 & 0.677 & 0.695 & 0.679 & 0.699 & 0.624 & 0.724 & 0.681 & 0.664 & 0.680 & 0.606 & 0.647 & 0.668 & 0.633 & 0.531 & 0.679 & 0.637 & 0.641 & 0.601 & 0.632 & 0.631 & 0.643 & 0.640 & 0.653 & 0.561 & 0.681 & 0.640 \\
 &  & thoracic vertebrae 4 (t4) & Dice & 0.823 & 0.834 & 0.798 & 0.805 & 0.813 & 0.806 & 0.737 & 0.834 & 0.806 & 0.814 & 0.790 & 0.805 & 0.798 & 0.810 & 0.794 & 0.817 & 0.773 & 0.834 & 0.809 & 0.797 & 0.808 & 0.756 & 0.775 & 0.793 & 0.773 & 0.686 & 0.806 & 0.781 & 0.778 & 0.752 & 0.774 & 0.771 & 0.775 & 0.779 & 0.789 & 0.716 & 0.807 & 0.780 \\
 &  &  & mIoU & 0.704 & 0.719 & 0.669 & 0.680 & 0.689 & 0.679 & 0.591 & 0.719 & 0.681 & 0.691 & 0.658 & 0.679 & 0.669 & 0.686 & 0.664 & 0.695 & 0.636 & 0.719 & 0.684 & 0.667 & 0.681 & 0.614 & 0.638 & 0.662 & 0.634 & 0.531 & 0.678 & 0.646 & 0.642 & 0.609 & 0.638 & 0.633 & 0.638 & 0.643 & 0.656 & 0.565 & 0.680 & 0.646 \\
 &  & thoracic vertebrae 5 (t5) & Dice & 0.808 & 0.821 & 0.787 & 0.793 & 0.801 & 0.789 & 0.717 & 0.821 & 0.796 & 0.801 & 0.769 & 0.787 & 0.788 & 0.789 & 0.788 & 0.800 & 0.747 & 0.822 & 0.797 & 0.788 & 0.798 & 0.745 & 0.774 & 0.785 & 0.761 & 0.676 & 0.798 & 0.771 & 0.768 & 0.744 & 0.753 & 0.761 & 0.764 & 0.766 & 0.780 & 0.710 & 0.798 & 0.772 \\
 &  &  & mIoU & 0.682 & 0.701 & 0.654 & 0.663 & 0.674 & 0.656 & 0.568 & 0.700 & 0.666 & 0.674 & 0.631 & 0.655 & 0.656 & 0.658 & 0.656 & 0.671 & 0.603 & 0.701 & 0.668 & 0.655 & 0.669 & 0.601 & 0.638 & 0.651 & 0.619 & 0.520 & 0.669 & 0.634 & 0.630 & 0.600 & 0.612 & 0.622 & 0.626 & 0.627 & 0.646 & 0.558 & 0.669 & 0.635 \\
 &  & thoracic vertebrae 6 (t6) & Dice & 0.787 & 0.804 & 0.762 & 0.772 & 0.782 & 0.764 & 0.700 & 0.800 & 0.767 & 0.780 & 0.743 & 0.760 & 0.767 & 0.760 & 0.762 & 0.778 & 0.731 & 0.802 & 0.769 & 0.771 & 0.784 & 0.727 & 0.761 & 0.767 & 0.739 & 0.647 & 0.783 & 0.750 & 0.745 & 0.717 & 0.712 & 0.741 & 0.739 & 0.745 & 0.750 & 0.682 & 0.784 & 0.747 \\
 &  &  & mIoU & 0.655 & 0.679 & 0.623 & 0.637 & 0.650 & 0.625 & 0.548 & 0.673 & 0.631 & 0.647 & 0.599 & 0.621 & 0.629 & 0.622 & 0.624 & 0.644 & 0.585 & 0.676 & 0.633 & 0.636 & 0.652 & 0.582 & 0.624 & 0.632 & 0.595 & 0.490 & 0.652 & 0.610 & 0.604 & 0.569 & 0.562 & 0.599 & 0.598 & 0.603 & 0.611 & 0.528 & 0.653 & 0.607 \\
 &  & thoracic vertebrae 7 (t7) & Dice & 0.780 & 0.798 & 0.752 & 0.767 & 0.772 & 0.751 & 0.684 & 0.795 & 0.755 & 0.770 & 0.726 & 0.743 & 0.752 & 0.755 & 0.756 & 0.773 & 0.718 & 0.793 & 0.758 & 0.761 & 0.774 & 0.716 & 0.753 & 0.756 & 0.732 & 0.617 & 0.774 & 0.737 & 0.729 & 0.698 & 0.729 & 0.728 & 0.731 & 0.736 & 0.750 & 0.655 & 0.774 & 0.735 \\
 &  &  & mIoU & 0.650 & 0.673 & 0.613 & 0.632 & 0.639 & 0.612 & 0.531 & 0.668 & 0.617 & 0.637 & 0.580 & 0.600 & 0.612 & 0.617 & 0.618 & 0.640 & 0.570 & 0.668 & 0.621 & 0.626 & 0.642 & 0.571 & 0.617 & 0.621 & 0.589 & 0.461 & 0.643 & 0.597 & 0.586 & 0.550 & 0.584 & 0.585 & 0.590 & 0.594 & 0.612 & 0.501 & 0.644 & 0.595 \\
 &  & thoracic vertebrae 8 (t8) & Dice & 0.781 & 0.799 & 0.747 & 0.770 & 0.766 & 0.754 & 0.673 & 0.794 & 0.752 & 0.767 & 0.721 & 0.732 & 0.752 & 0.745 & 0.757 & 0.776 & 0.698 & 0.792 & 0.747 & 0.760 & 0.769 & 0.704 & 0.745 & 0.749 & 0.733 & 0.599 & 0.775 & 0.727 & 0.722 & 0.688 & 0.735 & 0.725 & 0.717 & 0.735 & 0.750 & 0.640 & 0.771 & 0.727 \\
 &  &  & mIoU & 0.650 & 0.675 & 0.606 & 0.638 & 0.633 & 0.615 & 0.519 & 0.668 & 0.614 & 0.634 & 0.576 & 0.590 & 0.613 & 0.606 & 0.621 & 0.647 & 0.547 & 0.666 & 0.609 & 0.626 & 0.638 & 0.557 & 0.609 & 0.614 & 0.591 & 0.444 & 0.646 & 0.587 & 0.580 & 0.539 & 0.595 & 0.584 & 0.575 & 0.595 & 0.614 & 0.486 & 0.642 & 0.587 \\
 &  & thoracic vertebrae 9 (t9) & Dice & 0.761 & 0.779 & 0.722 & 0.749 & 0.743 & 0.737 & 0.644 & 0.778 & 0.730 & 0.742 & 0.699 & 0.732 & 0.729 & 0.722 & 0.742 & 0.754 & 0.673 & 0.773 & 0.721 & 0.740 & 0.748 & 0.662 & 0.715 & 0.721 & 0.706 & 0.546 & 0.754 & 0.703 & 0.689 & 0.650 & 0.711 & 0.704 & 0.688 & 0.706 & 0.724 & 0.599 & 0.747 & 0.703 \\
 &  &  & mIoU & 0.635 & 0.658 & 0.583 & 0.620 & 0.613 & 0.605 & 0.492 & 0.657 & 0.596 & 0.611 & 0.557 & 0.597 & 0.592 & 0.587 & 0.611 & 0.629 & 0.526 & 0.650 & 0.589 & 0.611 & 0.622 & 0.519 & 0.585 & 0.591 & 0.570 & 0.398 & 0.630 & 0.567 & 0.551 & 0.505 & 0.578 & 0.566 & 0.550 & 0.573 & 0.592 & 0.446 & 0.622 & 0.568 \\
 &  & thoracic vertebrae 10 (t10) & Dice & 0.783 & 0.802 & 0.744 & 0.772 & 0.769 & 0.769 & 0.653 & 0.803 & 0.757 & 0.760 & 0.725 & 0.757 & 0.752 & 0.750 & 0.762 & 0.767 & 0.689 & 0.794 & 0.755 & 0.762 & 0.772 & 0.671 & 0.741 & 0.751 & 0.730 & 0.531 & 0.774 & 0.733 & 0.707 & 0.676 & 0.715 & 0.721 & 0.718 & 0.732 & 0.733 & 0.623 & 0.770 & 0.727 \\
 &  &  & mIoU & 0.660 & 0.686 & 0.609 & 0.646 & 0.644 & 0.643 & 0.503 & 0.689 & 0.627 & 0.631 & 0.588 & 0.629 & 0.618 & 0.618 & 0.635 & 0.646 & 0.545 & 0.674 & 0.625 & 0.639 & 0.651 & 0.528 & 0.614 & 0.624 & 0.597 & 0.382 & 0.655 & 0.601 & 0.572 & 0.532 & 0.580 & 0.587 & 0.583 & 0.601 & 0.604 & 0.473 & 0.649 & 0.595 \\
 &  & thoracic vertebrae 11 (t11) & Dice & 0.761 & 0.785 & 0.729 & 0.758 & 0.750 & 0.756 & 0.644 & 0.785 & 0.747 & 0.742 & 0.707 & 0.746 & 0.732 & 0.742 & 0.750 & 0.749 & 0.700 & 0.773 & 0.746 & 0.730 & 0.745 & 0.634 & 0.723 & 0.715 & 0.701 & 0.497 & 0.756 & 0.708 & 0.681 & 0.637 & 0.698 & 0.684 & 0.696 & 0.700 & 0.703 & 0.568 & 0.747 & 0.704 \\
 &  &  & mIoU & 0.644 & 0.672 & 0.598 & 0.637 & 0.629 & 0.636 & 0.500 & 0.675 & 0.623 & 0.615 & 0.575 & 0.624 & 0.602 & 0.618 & 0.628 & 0.628 & 0.564 & 0.657 & 0.620 & 0.610 & 0.628 & 0.498 & 0.602 & 0.594 & 0.574 & 0.359 & 0.639 & 0.582 & 0.551 & 0.496 & 0.572 & 0.555 & 0.570 & 0.573 & 0.580 & 0.426 & 0.630 & 0.578 \\
 &  & thoracic vertebrae 12 (t12) & Dice & 0.762 & 0.781 & 0.731 & 0.760 & 0.750 & 0.757 & 0.647 & 0.788 & 0.743 & 0.738 & 0.706 & 0.747 & 0.738 & 0.750 & 0.757 & 0.759 & 0.705 & 0.776 & 0.744 & 0.727 & 0.743 & 0.649 & 0.714 & 0.713 & 0.695 & 0.487 & 0.751 & 0.703 & 0.679 & 0.630 & 0.710 & 0.693 & 0.700 & 0.705 & 0.713 & 0.587 & 0.746 & 0.705 \\
 &  &  & mIoU & 0.646 & 0.668 & 0.603 & 0.642 & 0.633 & 0.639 & 0.504 & 0.678 & 0.624 & 0.613 & 0.573 & 0.626 & 0.613 & 0.629 & 0.638 & 0.640 & 0.572 & 0.662 & 0.622 & 0.612 & 0.627 & 0.519 & 0.596 & 0.597 & 0.572 & 0.354 & 0.637 & 0.583 & 0.553 & 0.492 & 0.589 & 0.566 & 0.578 & 0.582 & 0.591 & 0.443 & 0.632 & 0.584 \\
 & lumbar vertebrae & lumbar vertebrae 1 (l1) & Dice & 0.761 & 0.781 & 0.721 & 0.761 & 0.756 & 0.757 & 0.639 & 0.787 & 0.749 & 0.736 & 0.692 & 0.745 & 0.737 & 0.751 & 0.756 & 0.764 & 0.706 & 0.774 & 0.745 & 0.765 & 0.775 & 0.709 & 0.757 & 0.757 & 0.723 & 0.549 & 0.780 & 0.746 & 0.716 & 0.661 & 0.755 & 0.725 & 0.747 & 0.743 & 0.759 & 0.655 & 0.774 & 0.741 \\
 &  &  & mIoU & 0.644 & 0.667 & 0.592 & 0.643 & 0.637 & 0.637 & 0.502 & 0.674 & 0.629 & 0.610 & 0.559 & 0.626 & 0.612 & 0.630 & 0.635 & 0.646 & 0.572 & 0.660 & 0.625 & 0.654 & 0.666 & 0.586 & 0.647 & 0.646 & 0.603 & 0.414 & 0.673 & 0.632 & 0.596 & 0.530 & 0.642 & 0.603 & 0.631 & 0.625 & 0.645 & 0.519 & 0.667 & 0.627 \\
 &  & lumbar vertebrae 2 (l2) & Dice & 0.582 & 0.604 & 0.526 & 0.569 & 0.575 & 0.582 & 0.436 & 0.616 & 0.545 & 0.541 & 0.488 & 0.544 & 0.531 & 0.551 & 0.544 & 0.561 & 0.504 & 0.600 & 0.550 & 0.585 & 0.611 & 0.516 & 0.574 & 0.586 & 0.535 & 0.378 & 0.623 & 0.527 & 0.539 & 0.469 & 0.538 & 0.503 & 0.559 & 0.525 & 0.537 & 0.439 & 0.615 & 0.548 \\
 &  &  & mIoU & 0.479 & 0.503 & 0.416 & 0.467 & 0.473 & 0.477 & 0.330 & 0.516 & 0.441 & 0.436 & 0.381 & 0.439 & 0.426 & 0.448 & 0.444 & 0.462 & 0.391 & 0.500 & 0.445 & 0.488 & 0.513 & 0.410 & 0.475 & 0.487 & 0.434 & 0.279 & 0.525 & 0.433 & 0.438 & 0.368 & 0.437 & 0.404 & 0.458 & 0.429 & 0.442 & 0.332 & 0.518 & 0.450 \\
 &  & lumbar vertebrae 3 (l3) & Dice & 0.339 & 0.400 & 0.225 & 0.342 & 0.324 & 0.365 & 0.192 & 0.406 & 0.315 & 0.301 & 0.235 & 0.301 & 0.274 & 0.313 & 0.324 & 0.353 & 0.252 & 0.397 & 0.329 & 0.339 & 0.400 & 0.248 & 0.338 & 0.326 & 0.251 & 0.167 & 0.391 & 0.284 & 0.306 & 0.224 & 0.275 & 0.228 & 0.302 & 0.292 & 0.332 & 0.211 & 0.411 & 0.307 \\
 &  &  & mIoU & 0.261 & 0.313 & 0.164 & 0.261 & 0.250 & 0.278 & 0.134 & 0.320 & 0.239 & 0.227 & 0.174 & 0.226 & 0.201 & 0.237 & 0.245 & 0.271 & 0.180 & 0.312 & 0.250 & 0.264 & 0.312 & 0.183 & 0.259 & 0.252 & 0.192 & 0.118 & 0.308 & 0.215 & 0.232 & 0.166 & 0.204 & 0.166 & 0.229 & 0.216 & 0.251 & 0.148 & 0.320 & 0.232 \\
 &  & lumbar vertebrae 4 (l4) & Dice & 0.284 & 0.330 & 0.185 & 0.256 & 0.289 & 0.306 & 0.152 & 0.321 & 0.248 & 0.266 & 0.175 & 0.164 & 0.221 & 0.266 & 0.264 & 0.265 & 0.201 & 0.350 & 0.248 & 0.192 & 0.224 & 0.050 & 0.175 & 0.203 & 0.141 & 0.050 & 0.231 & 0.151 & 0.178 & 0.083 & 0.094 & 0.093 & 0.179 & 0.126 & 0.121 & 0.018 & 0.276 & 0.160 \\
 &  &  & mIoU & 0.206 & 0.242 & 0.130 & 0.184 & 0.211 & 0.221 & 0.107 & 0.235 & 0.177 & 0.192 & 0.126 & 0.116 & 0.158 & 0.191 & 0.185 & 0.187 & 0.141 & 0.257 & 0.176 & 0.134 & 0.155 & 0.033 & 0.117 & 0.141 & 0.098 & 0.032 & 0.162 & 0.101 & 0.119 & 0.057 & 0.063 & 0.060 & 0.120 & 0.083 & 0.079 & 0.011 & 0.194 & 0.107 \\
 &  & lumbar vertebrae 5 (l5) & Dice & 0.427 & 0.482 & 0.374 & 0.420 & 0.401 & 0.450 & 0.339 & 0.476 & 0.399 & 0.412 & 0.417 & 0.404 & 0.416 & 0.448 & 0.432 & 0.430 & 0.316 & 0.493 & 0.398 & 0.365 & 0.438 & 0.191 & 0.327 & 0.378 & 0.262 & 0.227 & 0.414 & 0.306 & 0.343 & 0.308 & 0.319 & 0.271 & 0.350 & 0.355 & 0.320 & 0.110 & 0.438 & 0.319 \\
 &  &  & mIoU & 0.308 & 0.358 & 0.262 & 0.303 & 0.290 & 0.328 & 0.236 & 0.352 & 0.287 & 0.293 & 0.295 & 0.285 & 0.297 & 0.326 & 0.309 & 0.307 & 0.211 & 0.366 & 0.285 & 0.250 & 0.310 & 0.122 & 0.221 & 0.262 & 0.177 & 0.149 & 0.290 & 0.206 & 0.233 & 0.209 & 0.216 & 0.178 & 0.240 & 0.242 & 0.215 & 0.068 & 0.310 & 0.216 \\
 &  & lumbar vertebrae 6 (l6) & Dice & 0.030 & 0.049 & 0.000 & 0.018 & 0.031 & 0.035 & 0.000 & 0.060 & 0.000 & 0.014 & 0.013 & 0.001 & 0.000 & 0.000 & 0.004 & 0.023 & 0.000 & 0.071 & 0.000 & 0.031 & 0.092 & 0.000 & 0.000 & 0.050 & 0.031 & 0.000 & 0.077 & 0.000 & 0.009 & 0.001 & 0.000 & 0.000 & 0.000 & 0.004 & 0.008 & 0.000 & 0.092 & 0.000 \\
 &  &  & mIoU & 0.019 & 0.034 & 0.000 & 0.011 & 0.020 & 0.023 & 0.000 & 0.041 & 0.000 & 0.008 & 0.008 & 0.001 & 0.000 & 0.000 & 0.002 & 0.014 & 0.000 & 0.047 & 0.000 & 0.020 & 0.065 & 0.000 & 0.000 & 0.036 & 0.020 & 0.000 & 0.054 & 0.000 & 0.005 & 0.000 & 0.000 & 0.000 & 0.000 & 0.002 & 0.005 & 0.000 & 0.064 & 0.000 \\
 &  & sacral vertebrae 1 (s1) & Dice & 0.320 & 0.374 & 0.205 & 0.330 & 0.337 & 0.322 & 0.142 & 0.383 & 0.262 & 0.274 & 0.263 & 0.288 & 0.227 & 0.310 & 0.287 & 0.339 & 0.210 & 0.400 & 0.259 & 0.321 & 0.391 & 0.156 & 0.279 & 0.357 & 0.262 & 0.098 & 0.371 & 0.221 & 0.281 & 0.231 & 0.248 & 0.249 & 0.263 & 0.273 & 0.332 & 0.124 & 0.386 & 0.228 \\
 &  &  & mIoU & 0.229 & 0.275 & 0.130 & 0.231 & 0.242 & 0.230 & 0.090 & 0.281 & 0.176 & 0.190 & 0.180 & 0.197 & 0.150 & 0.212 & 0.198 & 0.241 & 0.136 & 0.296 & 0.174 & 0.224 & 0.286 & 0.096 & 0.187 & 0.255 & 0.180 & 0.062 & 0.269 & 0.147 & 0.192 & 0.154 & 0.165 & 0.166 & 0.177 & 0.184 & 0.231 & 0.077 & 0.280 & 0.151 \\
rib & left rib & left rib 1 & Dice & 0.773 & 0.784 & 0.755 & 0.770 & 0.776 & 0.760 & 0.673 & 0.782 & 0.775 & 0.774 & 0.744 & 0.774 & 0.770 & 0.773 & 0.773 & 0.782 & 0.737 & 0.783 & 0.776 & 0.703 & 0.715 & 0.645 & 0.663 & 0.700 & 0.651 & 0.587 & 0.720 & 0.672 & 0.682 & 0.643 & 0.669 & 0.667 & 0.666 & 0.661 & 0.683 & 0.623 & 0.719 & 0.672 \\
 &  &  & mIoU & 0.641 & 0.656 & 0.619 & 0.638 & 0.646 & 0.625 & 0.519 & 0.654 & 0.644 & 0.644 & 0.604 & 0.643 & 0.638 & 0.643 & 0.642 & 0.654 & 0.595 & 0.655 & 0.645 & 0.555 & 0.570 & 0.493 & 0.512 & 0.552 & 0.497 & 0.431 & 0.575 & 0.522 & 0.533 & 0.490 & 0.518 & 0.516 & 0.516 & 0.510 & 0.534 & 0.470 & 0.574 & 0.522 \\
 &  & left rib 2 & Dice & 0.731 & 0.753 & 0.705 & 0.731 & 0.738 & 0.701 & 0.521 & 0.746 & 0.729 & 0.725 & 0.688 & 0.737 & 0.718 & 0.731 & 0.730 & 0.747 & 0.665 & 0.743 & 0.734 & 0.675 & 0.683 & 0.566 & 0.618 & 0.667 & 0.614 & 0.452 & 0.693 & 0.614 & 0.617 & 0.571 & 0.607 & 0.592 & 0.609 & 0.587 & 0.638 & 0.539 & 0.685 & 0.612 \\
 &  &  & mIoU & 0.585 & 0.613 & 0.555 & 0.586 & 0.594 & 0.551 & 0.362 & 0.604 & 0.585 & 0.579 & 0.535 & 0.594 & 0.571 & 0.587 & 0.585 & 0.606 & 0.511 & 0.602 & 0.591 & 0.525 & 0.535 & 0.419 & 0.469 & 0.517 & 0.459 & 0.308 & 0.546 & 0.466 & 0.468 & 0.419 & 0.457 & 0.444 & 0.460 & 0.438 & 0.489 & 0.392 & 0.537 & 0.464 \\
 &  & left rib 3 & Dice & 0.716 & 0.743 & 0.681 & 0.707 & 0.721 & 0.667 & 0.425 & 0.731 & 0.712 & 0.704 & 0.647 & 0.719 & 0.695 & 0.709 & 0.715 & 0.730 & 0.640 & 0.729 & 0.711 & 0.655 & 0.667 & 0.542 & 0.588 & 0.648 & 0.606 & 0.385 & 0.676 & 0.588 & 0.591 & 0.533 & 0.585 & 0.579 & 0.593 & 0.579 & 0.605 & 0.509 & 0.669 & 0.582 \\
 &  &  & mIoU & 0.570 & 0.603 & 0.531 & 0.562 & 0.577 & 0.513 & 0.281 & 0.589 & 0.567 & 0.557 & 0.492 & 0.574 & 0.547 & 0.563 & 0.571 & 0.588 & 0.486 & 0.587 & 0.566 & 0.505 & 0.520 & 0.401 & 0.443 & 0.500 & 0.454 & 0.255 & 0.530 & 0.442 & 0.444 & 0.386 & 0.438 & 0.434 & 0.446 & 0.433 & 0.457 & 0.368 & 0.521 & 0.437 \\
 &  & left rib 4 & Dice & 0.721 & 0.749 & 0.707 & 0.719 & 0.726 & 0.672 & 0.448 & 0.739 & 0.720 & 0.711 & 0.650 & 0.719 & 0.709 & 0.711 & 0.724 & 0.739 & 0.647 & 0.738 & 0.720 & 0.644 & 0.665 & 0.548 & 0.574 & 0.647 & 0.607 & 0.392 & 0.676 & 0.589 & 0.588 & 0.531 & 0.577 & 0.586 & 0.576 & 0.579 & 0.618 & 0.514 & 0.665 & 0.575 \\
 &  &  & mIoU & 0.577 & 0.611 & 0.561 & 0.576 & 0.582 & 0.519 & 0.298 & 0.598 & 0.578 & 0.566 & 0.496 & 0.576 & 0.564 & 0.566 & 0.582 & 0.600 & 0.494 & 0.598 & 0.577 & 0.497 & 0.521 & 0.404 & 0.429 & 0.501 & 0.457 & 0.260 & 0.531 & 0.443 & 0.442 & 0.385 & 0.432 & 0.439 & 0.430 & 0.433 & 0.469 & 0.373 & 0.519 & 0.429 \\
 &  & left rib 5 & Dice & 0.721 & 0.746 & 0.697 & 0.718 & 0.716 & 0.665 & 0.446 & 0.736 & 0.716 & 0.708 & 0.637 & 0.722 & 0.709 & 0.702 & 0.718 & 0.736 & 0.644 & 0.736 & 0.713 & 0.637 & 0.660 & 0.538 & 0.573 & 0.638 & 0.599 & 0.394 & 0.670 & 0.571 & 0.580 & 0.522 & 0.573 & 0.580 & 0.552 & 0.558 & 0.599 & 0.496 & 0.659 & 0.574 \\
 &  &  & mIoU & 0.576 & 0.607 & 0.549 & 0.574 & 0.572 & 0.511 & 0.295 & 0.595 & 0.572 & 0.562 & 0.481 & 0.579 & 0.563 & 0.555 & 0.574 & 0.596 & 0.491 & 0.595 & 0.569 & 0.488 & 0.515 & 0.395 & 0.427 & 0.492 & 0.448 & 0.261 & 0.525 & 0.425 & 0.435 & 0.376 & 0.427 & 0.433 & 0.406 & 0.412 & 0.451 & 0.353 & 0.512 & 0.428 \\
 &  & left rib 6 & Dice & 0.705 & 0.728 & 0.676 & 0.701 & 0.698 & 0.650 & 0.419 & 0.723 & 0.697 & 0.686 & 0.613 & 0.704 & 0.675 & 0.676 & 0.708 & 0.711 & 0.617 & 0.723 & 0.694 & 0.615 & 0.640 & 0.520 & 0.552 & 0.610 & 0.569 & 0.366 & 0.649 & 0.554 & 0.551 & 0.495 & 0.559 & 0.517 & 0.529 & 0.545 & 0.562 & 0.486 & 0.637 & 0.546 \\
 &  &  & mIoU & 0.563 & 0.591 & 0.531 & 0.560 & 0.556 & 0.500 & 0.276 & 0.584 & 0.555 & 0.541 & 0.463 & 0.563 & 0.531 & 0.532 & 0.566 & 0.572 & 0.469 & 0.585 & 0.553 & 0.470 & 0.501 & 0.383 & 0.412 & 0.469 & 0.425 & 0.241 & 0.508 & 0.412 & 0.411 & 0.356 & 0.418 & 0.380 & 0.389 & 0.405 & 0.422 & 0.350 & 0.494 & 0.405 \\
 &  & left rib 7 & Dice & 0.709 & 0.738 & 0.688 & 0.709 & 0.704 & 0.659 & 0.448 & 0.730 & 0.699 & 0.694 & 0.613 & 0.710 & 0.691 & 0.677 & 0.712 & 0.723 & 0.617 & 0.731 & 0.697 & 0.632 & 0.673 & 0.553 & 0.588 & 0.645 & 0.596 & 0.403 & 0.672 & 0.583 & 0.570 & 0.526 & 0.592 & 0.572 & 0.555 & 0.589 & 0.608 & 0.528 & 0.659 & 0.579 \\
 &  &  & mIoU & 0.565 & 0.599 & 0.540 & 0.565 & 0.560 & 0.508 & 0.299 & 0.589 & 0.555 & 0.547 & 0.461 & 0.567 & 0.546 & 0.531 & 0.569 & 0.582 & 0.465 & 0.590 & 0.554 & 0.485 & 0.531 & 0.412 & 0.443 & 0.502 & 0.449 & 0.270 & 0.530 & 0.437 & 0.427 & 0.384 & 0.447 & 0.429 & 0.412 & 0.443 & 0.464 & 0.386 & 0.514 & 0.435 \\
 &  & left rib 8 & Dice & 0.698 & 0.725 & 0.664 & 0.699 & 0.685 & 0.655 & 0.399 & 0.721 & 0.688 & 0.682 & 0.602 & 0.700 & 0.683 & 0.652 & 0.699 & 0.712 & 0.590 & 0.719 & 0.684 & 0.620 & 0.666 & 0.546 & 0.569 & 0.629 & 0.582 & 0.394 & 0.666 & 0.566 & 0.556 & 0.522 & 0.573 & 0.555 & 0.546 & 0.577 & 0.589 & 0.515 & 0.652 & 0.563 \\
 &  &  & mIoU & 0.554 & 0.586 & 0.517 & 0.555 & 0.540 & 0.506 & 0.259 & 0.581 & 0.543 & 0.536 & 0.450 & 0.557 & 0.537 & 0.506 & 0.555 & 0.571 & 0.439 & 0.577 & 0.540 & 0.476 & 0.526 & 0.408 & 0.429 & 0.487 & 0.437 & 0.264 & 0.526 & 0.424 & 0.417 & 0.383 & 0.430 & 0.416 & 0.406 & 0.434 & 0.447 & 0.376 & 0.510 & 0.423 \\
 &  & left rib 9 & Dice & 0.698 & 0.728 & 0.666 & 0.696 & 0.694 & 0.654 & 0.356 & 0.724 & 0.686 & 0.679 & 0.594 & 0.699 & 0.682 & 0.649 & 0.700 & 0.713 & 0.588 & 0.718 & 0.686 & 0.620 & 0.665 & 0.551 & 0.580 & 0.638 & 0.581 & 0.398 & 0.668 & 0.583 & 0.557 & 0.529 & 0.574 & 0.578 & 0.548 & 0.588 & 0.594 & 0.508 & 0.652 & 0.567 \\
 &  &  & mIoU & 0.555 & 0.591 & 0.519 & 0.554 & 0.551 & 0.508 & 0.230 & 0.586 & 0.543 & 0.534 & 0.445 & 0.557 & 0.538 & 0.505 & 0.557 & 0.575 & 0.438 & 0.579 & 0.543 & 0.475 & 0.524 & 0.410 & 0.437 & 0.496 & 0.435 & 0.267 & 0.526 & 0.439 & 0.418 & 0.389 & 0.430 & 0.434 & 0.406 & 0.444 & 0.451 & 0.370 & 0.510 & 0.425 \\
 &  & left rib 10 & Dice & 0.655 & 0.697 & 0.620 & 0.654 & 0.658 & 0.628 & 0.255 & 0.693 & 0.643 & 0.643 & 0.539 & 0.667 & 0.638 & 0.631 & 0.665 & 0.683 & 0.544 & 0.685 & 0.643 & 0.585 & 0.627 & 0.501 & 0.569 & 0.592 & 0.535 & 0.313 & 0.632 & 0.557 & 0.523 & 0.482 & 0.547 & 0.545 & 0.524 & 0.551 & 0.569 & 0.461 & 0.625 & 0.527 \\
 &  &  & mIoU & 0.519 & 0.564 & 0.485 & 0.521 & 0.523 & 0.490 & 0.161 & 0.560 & 0.509 & 0.508 & 0.405 & 0.532 & 0.503 & 0.495 & 0.529 & 0.550 & 0.409 & 0.551 & 0.507 & 0.443 & 0.488 & 0.368 & 0.428 & 0.452 & 0.395 & 0.203 & 0.492 & 0.417 & 0.388 & 0.349 & 0.408 & 0.406 & 0.388 & 0.413 & 0.428 & 0.332 & 0.484 & 0.392 \\
 &  & left rib 11 & Dice & 0.593 & 0.655 & 0.558 & 0.621 & 0.610 & 0.581 & 0.226 & 0.650 & 0.589 & 0.588 & 0.479 & 0.618 & 0.581 & 0.587 & 0.614 & 0.636 & 0.497 & 0.644 & 0.594 & 0.539 & 0.578 & 0.457 & 0.540 & 0.537 & 0.478 & 0.268 & 0.580 & 0.527 & 0.486 & 0.408 & 0.520 & 0.500 & 0.511 & 0.527 & 0.540 & 0.409 & 0.585 & 0.520 \\
 &  &  & mIoU & 0.464 & 0.524 & 0.431 & 0.492 & 0.481 & 0.449 & 0.144 & 0.519 & 0.459 & 0.458 & 0.355 & 0.487 & 0.453 & 0.456 & 0.484 & 0.506 & 0.373 & 0.513 & 0.463 & 0.400 & 0.438 & 0.325 & 0.398 & 0.400 & 0.342 & 0.168 & 0.441 & 0.387 & 0.351 & 0.283 & 0.379 & 0.363 & 0.373 & 0.386 & 0.401 & 0.281 & 0.444 & 0.382 \\
 &  & left rib 12 & Dice & 0.447 & 0.509 & 0.393 & 0.471 & 0.459 & 0.436 & 0.123 & 0.507 & 0.439 & 0.415 & 0.330 & 0.458 & 0.433 & 0.449 & 0.460 & 0.496 & 0.351 & 0.486 & 0.447 & 0.390 & 0.420 & 0.289 & 0.389 & 0.388 & 0.311 & 0.099 & 0.442 & 0.375 & 0.344 & 0.283 & 0.359 & 0.367 & 0.385 & 0.363 & 0.416 & 0.213 & 0.462 & 0.399 \\
 &  &  & mIoU & 0.333 & 0.388 & 0.288 & 0.352 & 0.342 & 0.320 & 0.075 & 0.386 & 0.326 & 0.306 & 0.233 & 0.340 & 0.320 & 0.333 & 0.344 & 0.374 & 0.250 & 0.366 & 0.333 & 0.278 & 0.302 & 0.196 & 0.276 & 0.273 & 0.210 & 0.059 & 0.321 & 0.265 & 0.237 & 0.189 & 0.253 & 0.255 & 0.273 & 0.255 & 0.299 & 0.138 & 0.338 & 0.283 \\
 & right rib & right rib 1 & Dice & 0.769 & 0.778 & 0.750 & 0.767 & 0.770 & 0.745 & 0.670 & 0.776 & 0.766 & 0.765 & 0.734 & 0.764 & 0.763 & 0.760 & 0.764 & 0.775 & 0.729 & 0.779 & 0.767 & 0.693 & 0.701 & 0.630 & 0.643 & 0.686 & 0.632 & 0.561 & 0.709 & 0.647 & 0.662 & 0.612 & 0.653 & 0.646 & 0.642 & 0.652 & 0.667 & 0.600 & 0.707 & 0.649 \\
 &  &  & mIoU & 0.637 & 0.650 & 0.614 & 0.636 & 0.639 & 0.607 & 0.518 & 0.647 & 0.634 & 0.634 & 0.594 & 0.632 & 0.630 & 0.628 & 0.633 & 0.646 & 0.587 & 0.651 & 0.635 & 0.545 & 0.555 & 0.479 & 0.493 & 0.538 & 0.477 & 0.406 & 0.564 & 0.498 & 0.512 & 0.461 & 0.502 & 0.497 & 0.492 & 0.502 & 0.518 & 0.448 & 0.562 & 0.501 \\
 &  & right rib 2 & Dice & 0.720 & 0.740 & 0.691 & 0.718 & 0.723 & 0.668 & 0.517 & 0.733 & 0.713 & 0.707 & 0.667 & 0.720 & 0.699 & 0.714 & 0.717 & 0.732 & 0.648 & 0.732 & 0.710 & 0.660 & 0.665 & 0.549 & 0.597 & 0.649 & 0.597 & 0.442 & 0.679 & 0.587 & 0.588 & 0.537 & 0.592 & 0.563 & 0.592 & 0.580 & 0.619 & 0.521 & 0.670 & 0.579 \\
 &  &  & mIoU & 0.574 & 0.599 & 0.539 & 0.573 & 0.578 & 0.514 & 0.359 & 0.590 & 0.567 & 0.559 & 0.512 & 0.575 & 0.550 & 0.568 & 0.570 & 0.590 & 0.493 & 0.590 & 0.564 & 0.510 & 0.519 & 0.403 & 0.451 & 0.500 & 0.444 & 0.299 & 0.531 & 0.441 & 0.442 & 0.390 & 0.444 & 0.418 & 0.445 & 0.433 & 0.471 & 0.376 & 0.522 & 0.435 \\
 &  & right rib 3 & Dice & 0.701 & 0.728 & 0.657 & 0.698 & 0.706 & 0.639 & 0.408 & 0.717 & 0.692 & 0.680 & 0.619 & 0.698 & 0.682 & 0.691 & 0.702 & 0.712 & 0.611 & 0.717 & 0.686 & 0.638 & 0.649 & 0.527 & 0.555 & 0.627 & 0.584 & 0.351 & 0.661 & 0.567 & 0.567 & 0.505 & 0.574 & 0.560 & 0.576 & 0.575 & 0.588 & 0.494 & 0.652 & 0.559 \\
 &  &  & mIoU & 0.556 & 0.586 & 0.507 & 0.552 & 0.561 & 0.486 & 0.267 & 0.574 & 0.547 & 0.532 & 0.464 & 0.553 & 0.535 & 0.546 & 0.557 & 0.570 & 0.460 & 0.574 & 0.540 & 0.490 & 0.505 & 0.387 & 0.414 & 0.481 & 0.435 & 0.228 & 0.516 & 0.424 & 0.424 & 0.362 & 0.431 & 0.417 & 0.431 & 0.431 & 0.444 & 0.357 & 0.507 & 0.417 \\
 &  & right rib 4 & Dice & 0.715 & 0.737 & 0.681 & 0.710 & 0.708 & 0.657 & 0.433 & 0.728 & 0.705 & 0.694 & 0.625 & 0.710 & 0.698 & 0.692 & 0.709 & 0.726 & 0.627 & 0.728 & 0.701 & 0.637 & 0.655 & 0.541 & 0.555 & 0.628 & 0.592 & 0.361 & 0.666 & 0.576 & 0.576 & 0.510 & 0.577 & 0.569 & 0.563 & 0.584 & 0.604 & 0.511 & 0.658 & 0.559 \\
 &  &  & mIoU & 0.571 & 0.598 & 0.534 & 0.567 & 0.564 & 0.505 & 0.287 & 0.588 & 0.562 & 0.548 & 0.471 & 0.567 & 0.553 & 0.549 & 0.566 & 0.586 & 0.475 & 0.588 & 0.558 & 0.490 & 0.512 & 0.399 & 0.414 & 0.483 & 0.443 & 0.236 & 0.522 & 0.432 & 0.432 & 0.368 & 0.433 & 0.426 & 0.420 & 0.439 & 0.458 & 0.371 & 0.513 & 0.416 \\
 &  & right rib 5 & Dice & 0.724 & 0.746 & 0.696 & 0.720 & 0.715 & 0.664 & 0.457 & 0.741 & 0.711 & 0.705 & 0.640 & 0.722 & 0.702 & 0.706 & 0.716 & 0.736 & 0.653 & 0.738 & 0.716 & 0.646 & 0.668 & 0.558 & 0.577 & 0.648 & 0.604 & 0.397 & 0.681 & 0.584 & 0.591 & 0.535 & 0.592 & 0.583 & 0.571 & 0.591 & 0.615 & 0.517 & 0.672 & 0.582 \\
 &  &  & mIoU & 0.580 & 0.606 & 0.548 & 0.576 & 0.570 & 0.510 & 0.305 & 0.600 & 0.566 & 0.558 & 0.484 & 0.578 & 0.556 & 0.560 & 0.572 & 0.595 & 0.499 & 0.597 & 0.571 & 0.497 & 0.523 & 0.414 & 0.431 & 0.502 & 0.452 & 0.263 & 0.535 & 0.437 & 0.446 & 0.389 & 0.444 & 0.437 & 0.425 & 0.443 & 0.467 & 0.373 & 0.525 & 0.437 \\
 &  & right rib 6 & Dice & 0.726 & 0.746 & 0.703 & 0.724 & 0.713 & 0.664 & 0.456 & 0.743 & 0.713 & 0.706 & 0.638 & 0.722 & 0.692 & 0.704 & 0.723 & 0.736 & 0.657 & 0.742 & 0.716 & 0.642 & 0.663 & 0.561 & 0.580 & 0.639 & 0.598 & 0.412 & 0.676 & 0.591 & 0.580 & 0.527 & 0.598 & 0.552 & 0.559 & 0.587 & 0.599 & 0.524 & 0.666 & 0.575 \\
 &  &  & mIoU & 0.584 & 0.609 & 0.557 & 0.582 & 0.570 & 0.512 & 0.306 & 0.604 & 0.571 & 0.562 & 0.483 & 0.580 & 0.547 & 0.559 & 0.581 & 0.597 & 0.505 & 0.603 & 0.573 & 0.496 & 0.520 & 0.419 & 0.437 & 0.496 & 0.449 & 0.276 & 0.532 & 0.446 & 0.438 & 0.385 & 0.452 & 0.411 & 0.415 & 0.442 & 0.454 & 0.383 & 0.522 & 0.431 \\
 &  & right rib 7 & Dice & 0.722 & 0.746 & 0.702 & 0.715 & 0.714 & 0.661 & 0.450 & 0.741 & 0.712 & 0.704 & 0.625 & 0.719 & 0.697 & 0.690 & 0.717 & 0.735 & 0.642 & 0.740 & 0.713 & 0.638 & 0.667 & 0.560 & 0.589 & 0.647 & 0.594 & 0.399 & 0.675 & 0.582 & 0.570 & 0.528 & 0.597 & 0.569 & 0.559 & 0.588 & 0.613 & 0.536 & 0.661 & 0.569 \\
 &  &  & mIoU & 0.579 & 0.607 & 0.555 & 0.572 & 0.570 & 0.509 & 0.300 & 0.601 & 0.568 & 0.559 & 0.470 & 0.576 & 0.551 & 0.543 & 0.574 & 0.595 & 0.488 & 0.600 & 0.570 & 0.491 & 0.525 & 0.417 & 0.444 & 0.504 & 0.446 & 0.266 & 0.531 & 0.437 & 0.428 & 0.385 & 0.451 & 0.426 & 0.415 & 0.442 & 0.467 & 0.393 & 0.517 & 0.425 \\
 &  & right rib 8 & Dice & 0.712 & 0.739 & 0.688 & 0.710 & 0.706 & 0.664 & 0.437 & 0.739 & 0.702 & 0.699 & 0.631 & 0.708 & 0.682 & 0.682 & 0.711 & 0.735 & 0.620 & 0.736 & 0.703 & 0.629 & 0.670 & 0.547 & 0.564 & 0.644 & 0.588 & 0.402 & 0.675 & 0.570 & 0.563 & 0.522 & 0.581 & 0.550 & 0.547 & 0.580 & 0.601 & 0.523 & 0.663 & 0.562 \\
 &  &  & mIoU & 0.566 & 0.600 & 0.539 & 0.566 & 0.561 & 0.513 & 0.290 & 0.598 & 0.557 & 0.553 & 0.477 & 0.563 & 0.536 & 0.534 & 0.566 & 0.594 & 0.465 & 0.595 & 0.558 & 0.483 & 0.528 & 0.405 & 0.422 & 0.500 & 0.441 & 0.268 & 0.532 & 0.426 & 0.421 & 0.381 & 0.436 & 0.409 & 0.404 & 0.436 & 0.457 & 0.380 & 0.519 & 0.419 \\
 &  & right rib 9 & Dice & 0.697 & 0.727 & 0.666 & 0.696 & 0.694 & 0.648 & 0.398 & 0.721 & 0.685 & 0.681 & 0.606 & 0.685 & 0.675 & 0.652 & 0.696 & 0.709 & 0.582 & 0.717 & 0.678 & 0.602 & 0.646 & 0.513 & 0.547 & 0.617 & 0.560 & 0.357 & 0.648 & 0.552 & 0.534 & 0.494 & 0.551 & 0.550 & 0.518 & 0.562 & 0.564 & 0.479 & 0.640 & 0.540 \\
 &  &  & mIoU & 0.555 & 0.589 & 0.521 & 0.556 & 0.552 & 0.503 & 0.262 & 0.584 & 0.544 & 0.539 & 0.460 & 0.542 & 0.532 & 0.509 & 0.556 & 0.571 & 0.434 & 0.579 & 0.537 & 0.459 & 0.506 & 0.377 & 0.408 & 0.475 & 0.417 & 0.236 & 0.507 & 0.412 & 0.396 & 0.359 & 0.410 & 0.409 & 0.380 & 0.420 & 0.424 & 0.346 & 0.497 & 0.401 \\
 &  & right rib 10 & Dice & 0.682 & 0.710 & 0.658 & 0.677 & 0.676 & 0.639 & 0.422 & 0.708 & 0.664 & 0.675 & 0.602 & 0.674 & 0.659 & 0.651 & 0.685 & 0.695 & 0.590 & 0.700 & 0.657 & 0.577 & 0.626 & 0.490 & 0.558 & 0.585 & 0.530 & 0.310 & 0.621 & 0.539 & 0.512 & 0.462 & 0.536 & 0.542 & 0.508 & 0.542 & 0.560 & 0.446 & 0.620 & 0.506 \\
 &  &  & mIoU & 0.548 & 0.578 & 0.524 & 0.546 & 0.543 & 0.503 & 0.292 & 0.577 & 0.530 & 0.541 & 0.466 & 0.540 & 0.525 & 0.517 & 0.549 & 0.562 & 0.452 & 0.568 & 0.524 & 0.436 & 0.485 & 0.356 & 0.417 & 0.445 & 0.390 & 0.201 & 0.482 & 0.400 & 0.377 & 0.332 & 0.398 & 0.403 & 0.373 & 0.404 & 0.421 & 0.319 & 0.479 & 0.372 \\
 &  & right rib 11 & Dice & 0.648 & 0.672 & 0.615 & 0.634 & 0.641 & 0.590 & 0.415 & 0.669 & 0.627 & 0.631 & 0.561 & 0.625 & 0.606 & 0.611 & 0.637 & 0.651 & 0.551 & 0.659 & 0.626 & 0.521 & 0.559 & 0.443 & 0.515 & 0.514 & 0.460 & 0.279 & 0.573 & 0.506 & 0.464 & 0.400 & 0.521 & 0.475 & 0.495 & 0.514 & 0.549 & 0.403 & 0.569 & 0.507 \\
 &  &  & mIoU & 0.517 & 0.541 & 0.485 & 0.505 & 0.512 & 0.459 & 0.296 & 0.540 & 0.496 & 0.501 & 0.432 & 0.497 & 0.477 & 0.482 & 0.506 & 0.520 & 0.423 & 0.529 & 0.494 & 0.386 & 0.422 & 0.314 & 0.377 & 0.378 & 0.327 & 0.177 & 0.435 & 0.368 & 0.334 & 0.277 & 0.380 & 0.344 & 0.358 & 0.375 & 0.408 & 0.277 & 0.430 & 0.370 \\
 &  & right rib 12 & Dice & 0.471 & 0.533 & 0.432 & 0.473 & 0.475 & 0.437 & 0.247 & 0.515 & 0.461 & 0.454 & 0.370 & 0.429 & 0.440 & 0.455 & 0.463 & 0.497 & 0.380 & 0.507 & 0.468 & 0.360 & 0.398 & 0.297 & 0.359 & 0.363 & 0.289 & 0.100 & 0.424 & 0.351 & 0.317 & 0.261 & 0.343 & 0.335 & 0.353 & 0.348 & 0.391 & 0.210 & 0.439 & 0.377 \\
 &  &  & mIoU & 0.360 & 0.412 & 0.327 & 0.360 & 0.365 & 0.324 & 0.170 & 0.399 & 0.350 & 0.346 & 0.270 & 0.322 & 0.333 & 0.345 & 0.352 & 0.381 & 0.280 & 0.389 & 0.355 & 0.254 & 0.284 & 0.202 & 0.251 & 0.255 & 0.194 & 0.060 & 0.308 & 0.245 & 0.216 & 0.172 & 0.239 & 0.232 & 0.247 & 0.244 & 0.277 & 0.136 & 0.320 & 0.266 \\
\midrule
\multicolumn{42}{l}{\textit{Vasculatory}} \\
heart atrium & left heart atrium & left auricle of heart & Dice & 0.661 & 0.672 & 0.570 & 0.634 & 0.653 & 0.639 & 0.456 & 0.680 & 0.610 & 0.649 & 0.608 & 0.509 & 0.604 & 0.605 & 0.585 & 0.595 & 0.506 & 0.668 & 0.602 & 0.641 & 0.645 & 0.556 & 0.605 & 0.626 & 0.610 & 0.401 & 0.656 & 0.600 & 0.614 & 0.554 & 0.505 & 0.567 & 0.585 & 0.544 & 0.560 & 0.398 & 0.647 & 0.583 \\
 &  &  & mIoU & 0.511 & 0.523 & 0.423 & 0.484 & 0.503 & 0.487 & 0.328 & 0.532 & 0.462 & 0.499 & 0.458 & 0.369 & 0.456 & 0.456 & 0.435 & 0.446 & 0.366 & 0.519 & 0.454 & 0.489 & 0.495 & 0.407 & 0.453 & 0.474 & 0.460 & 0.283 & 0.506 & 0.451 & 0.462 & 0.408 & 0.361 & 0.422 & 0.435 & 0.393 & 0.416 & 0.276 & 0.496 & 0.435 \\
carotid artery & left carotid artery & left common carotid artery & Dice & 0.673 & 0.679 & 0.639 & 0.665 & 0.659 & 0.646 & 0.515 & 0.680 & 0.661 & 0.662 & 0.644 & 0.662 & 0.656 & 0.657 & 0.650 & 0.668 & 0.626 & 0.674 & 0.662 & 0.633 & 0.642 & 0.544 & 0.622 & 0.628 & 0.558 & 0.408 & 0.643 & 0.597 & 0.600 & 0.548 & 0.596 & 0.599 & 0.604 & 0.588 & 0.619 & 0.528 & 0.643 & 0.606 \\
 &  &  & mIoU & 0.520 & 0.528 & 0.486 & 0.510 & 0.505 & 0.491 & 0.368 & 0.528 & 0.508 & 0.509 & 0.490 & 0.508 & 0.503 & 0.503 & 0.497 & 0.516 & 0.471 & 0.521 & 0.509 & 0.477 & 0.488 & 0.393 & 0.466 & 0.472 & 0.405 & 0.277 & 0.490 & 0.443 & 0.446 & 0.397 & 0.443 & 0.446 & 0.449 & 0.435 & 0.465 & 0.376 & 0.489 & 0.453 \\
 &  & left internal carotid artery & Dice & 0.008 & 0.011 & 0.000 & 0.000 & 0.011 & 0.005 & 0.000 & 0.015 & 0.000 & 0.001 & 0.000 & 0.000 & 0.000 & 0.000 & 0.000 & 0.000 & 0.000 & 0.016 & 0.000 & 0.004 & 0.011 & 0.000 & 0.002 & 0.006 & 0.007 & 0.000 & 0.012 & 0.000 & 0.000 & 0.002 & 0.000 & 0.000 & 0.000 & 0.000 & 0.001 & 0.000 & 0.012 & 0.000 \\
 &  &  & mIoU & 0.005 & 0.007 & 0.000 & 0.000 & 0.007 & 0.003 & 0.000 & 0.009 & 0.000 & 0.000 & 0.000 & 0.000 & 0.000 & 0.000 & 0.000 & 0.000 & 0.000 & 0.011 & 0.000 & 0.003 & 0.007 & 0.000 & 0.001 & 0.004 & 0.005 & 0.000 & 0.007 & 0.000 & 0.000 & 0.001 & 0.000 & 0.000 & 0.000 & 0.000 & 0.001 & 0.000 & 0.008 & 0.000 \\
 & right carotid artery & right common carotid artery & Dice & 0.623 & 0.636 & 0.571 & 0.614 & 0.615 & 0.569 & 0.404 & 0.631 & 0.601 & 0.602 & 0.573 & 0.592 & 0.595 & 0.593 & 0.599 & 0.628 & 0.544 & 0.630 & 0.611 & 0.599 & 0.607 & 0.509 & 0.575 & 0.582 & 0.503 & 0.324 & 0.610 & 0.551 & 0.567 & 0.498 & 0.531 & 0.557 & 0.550 & 0.565 & 0.570 & 0.477 & 0.613 & 0.559 \\
 &  &  & mIoU & 0.466 & 0.481 & 0.416 & 0.458 & 0.458 & 0.415 & 0.276 & 0.476 & 0.446 & 0.447 & 0.420 & 0.437 & 0.440 & 0.438 & 0.442 & 0.472 & 0.391 & 0.473 & 0.456 & 0.444 & 0.453 & 0.362 & 0.422 & 0.428 & 0.358 & 0.216 & 0.456 & 0.400 & 0.415 & 0.353 & 0.380 & 0.405 & 0.400 & 0.412 & 0.418 & 0.334 & 0.458 & 0.408 \\
 &  & right internal carotid artery & Dice & 0.009 & 0.012 & 0.000 & 0.003 & 0.011 & 0.008 & 0.000 & 0.012 & 0.000 & 0.002 & 0.002 & 0.002 & 0.002 & 0.000 & 0.000 & 0.001 & 0.000 & 0.027 & 0.000 & 0.009 & 0.016 & 0.000 & 0.007 & 0.011 & 0.011 & 0.000 & 0.017 & 0.000 & 0.002 & 0.003 & 0.001 & 0.001 & 0.003 & 0.000 & 0.004 & 0.000 & 0.027 & 0.000 \\
 &  &  & mIoU & 0.006 & 0.008 & 0.000 & 0.002 & 0.007 & 0.005 & 0.000 & 0.008 & 0.000 & 0.001 & 0.001 & 0.001 & 0.001 & 0.000 & 0.000 & 0.000 & 0.000 & 0.018 & 0.000 & 0.006 & 0.011 & 0.000 & 0.005 & 0.008 & 0.007 & 0.000 & 0.011 & 0.000 & 0.001 & 0.002 & 0.000 & 0.001 & 0.002 & 0.000 & 0.003 & 0.000 & 0.018 & 0.000 \\
\midrule
\multicolumn{42}{l}{\textit{Abdominal}} \\
thyroid & thyroid gland & left thyroid & Dice & 0.684 & 0.696 & 0.661 & 0.669 & 0.673 & 0.669 & 0.538 & 0.698 & 0.671 & 0.673 & 0.655 & 0.672 & 0.673 & 0.674 & 0.672 & 0.679 & 0.622 & 0.686 & 0.672 & 0.684 & 0.691 & 0.647 & 0.686 & 0.687 & 0.659 & 0.591 & 0.696 & 0.673 & 0.666 & 0.658 & 0.670 & 0.677 & 0.673 & 0.667 & 0.672 & 0.635 & 0.685 & 0.672 \\
 &  &  & mIoU & 0.543 & 0.557 & 0.517 & 0.526 & 0.532 & 0.528 & 0.398 & 0.560 & 0.530 & 0.532 & 0.513 & 0.530 & 0.530 & 0.531 & 0.529 & 0.539 & 0.476 & 0.547 & 0.532 & 0.546 & 0.555 & 0.508 & 0.551 & 0.551 & 0.521 & 0.452 & 0.561 & 0.535 & 0.529 & 0.518 & 0.532 & 0.539 & 0.537 & 0.529 & 0.535 & 0.495 & 0.549 & 0.534 \\
 &  & right thyroid & Dice & 0.704 & 0.712 & 0.676 & 0.672 & 0.700 & 0.684 & 0.594 & 0.713 & 0.686 & 0.692 & 0.673 & 0.683 & 0.689 & 0.697 & 0.687 & 0.697 & 0.643 & 0.706 & 0.686 & 0.715 & 0.718 & 0.693 & 0.716 & 0.716 & 0.691 & 0.637 & 0.721 & 0.705 & 0.700 & 0.693 & 0.702 & 0.703 & 0.711 & 0.707 & 0.707 & 0.679 & 0.711 & 0.703 \\
 &  &  & mIoU & 0.568 & 0.577 & 0.539 & 0.536 & 0.565 & 0.548 & 0.455 & 0.579 & 0.549 & 0.557 & 0.536 & 0.547 & 0.552 & 0.561 & 0.550 & 0.562 & 0.501 & 0.571 & 0.549 & 0.585 & 0.590 & 0.560 & 0.587 & 0.589 & 0.559 & 0.500 & 0.595 & 0.574 & 0.569 & 0.560 & 0.572 & 0.571 & 0.581 & 0.577 & 0.577 & 0.545 & 0.582 & 0.574 \\
liver & left lobe of liver & left lateral superior segment of liver & Dice & 0.655 & 0.672 & 0.626 & 0.642 & 0.656 & 0.644 & 0.568 & 0.673 & 0.645 & 0.646 & 0.634 & 0.617 & 0.633 & 0.634 & 0.628 & 0.641 & 0.593 & 0.663 & 0.643 & 0.702 & 0.716 & 0.674 & 0.690 & 0.703 & 0.690 & 0.621 & 0.716 & 0.691 & 0.688 & 0.672 & 0.669 & 0.683 & 0.685 & 0.683 & 0.669 & 0.604 & 0.707 & 0.689 \\
 &  &  & mIoU & 0.502 & 0.520 & 0.472 & 0.487 & 0.503 & 0.490 & 0.416 & 0.522 & 0.490 & 0.492 & 0.479 & 0.461 & 0.477 & 0.479 & 0.471 & 0.485 & 0.436 & 0.511 & 0.488 & 0.554 & 0.571 & 0.524 & 0.541 & 0.557 & 0.542 & 0.468 & 0.571 & 0.543 & 0.540 & 0.521 & 0.519 & 0.534 & 0.536 & 0.533 & 0.518 & 0.450 & 0.561 & 0.540 \\
 &  & left lateral inferior segment of liver & Dice & 0.626 & 0.631 & 0.564 & 0.612 & 0.618 & 0.615 & 0.486 & 0.636 & 0.582 & 0.609 & 0.590 & 0.547 & 0.588 & 0.590 & 0.542 & 0.566 & 0.489 & 0.624 & 0.583 & 0.613 & 0.623 & 0.533 & 0.588 & 0.585 & 0.583 & 0.482 & 0.626 & 0.573 & 0.586 & 0.564 & 0.549 & 0.580 & 0.543 & 0.563 & 0.559 & 0.486 & 0.617 & 0.583 \\
 &  &  & mIoU & 0.483 & 0.490 & 0.422 & 0.469 & 0.475 & 0.472 & 0.352 & 0.495 & 0.439 & 0.466 & 0.447 & 0.407 & 0.445 & 0.446 & 0.400 & 0.425 & 0.353 & 0.482 & 0.440 & 0.469 & 0.482 & 0.395 & 0.447 & 0.444 & 0.440 & 0.346 & 0.485 & 0.431 & 0.444 & 0.421 & 0.410 & 0.437 & 0.405 & 0.423 & 0.420 & 0.352 & 0.475 & 0.440 \\
 &  & left medial segment of liver & Dice & 0.215 & 0.249 & 0.093 & 0.209 & 0.235 & 0.225 & 0.045 & 0.257 & 0.162 & 0.196 & 0.126 & 0.116 & 0.143 & 0.142 & 0.149 & 0.100 & 0.043 & 0.250 & 0.165 & 0.225 & 0.255 & 0.111 & 0.181 & 0.229 & 0.225 & 0.078 & 0.260 & 0.180 & 0.191 & 0.188 & 0.189 & 0.162 & 0.191 & 0.207 & 0.165 & 0.087 & 0.271 & 0.196 \\
 &  &  & mIoU & 0.147 & 0.174 & 0.056 & 0.140 & 0.162 & 0.154 & 0.028 & 0.180 & 0.106 & 0.132 & 0.081 & 0.073 & 0.094 & 0.091 & 0.098 & 0.064 & 0.025 & 0.173 & 0.110 & 0.158 & 0.181 & 0.071 & 0.122 & 0.160 & 0.157 & 0.052 & 0.185 & 0.120 & 0.132 & 0.129 & 0.126 & 0.110 & 0.129 & 0.141 & 0.111 & 0.057 & 0.191 & 0.132 \\
 & right lobe of liver & right anterior superior segment of liver & Dice & 0.357 & 0.394 & 0.227 & 0.309 & 0.364 & 0.357 & 0.204 & 0.390 & 0.331 & 0.345 & 0.294 & 0.208 & 0.296 & 0.336 & 0.221 & 0.188 & 0.149 & 0.387 & 0.299 & 0.362 & 0.365 & 0.234 & 0.296 & 0.380 & 0.348 & 0.185 & 0.381 & 0.298 & 0.323 & 0.296 & 0.190 & 0.305 & 0.316 & 0.207 & 0.216 & 0.157 & 0.394 & 0.305 \\
 &  &  & mIoU & 0.235 & 0.264 & 0.145 & 0.200 & 0.242 & 0.235 & 0.131 & 0.262 & 0.216 & 0.227 & 0.191 & 0.129 & 0.189 & 0.217 & 0.139 & 0.121 & 0.095 & 0.258 & 0.193 & 0.239 & 0.243 & 0.151 & 0.188 & 0.252 & 0.228 & 0.116 & 0.254 & 0.191 & 0.211 & 0.190 & 0.112 & 0.193 & 0.202 & 0.126 & 0.135 & 0.096 & 0.262 & 0.196 \\
 &  & right anterior inferior segment of liver & Dice & 0.554 & 0.578 & 0.487 & 0.534 & 0.541 & 0.539 & 0.427 & 0.572 & 0.522 & 0.539 & 0.503 & 0.503 & 0.518 & 0.511 & 0.514 & 0.530 & 0.419 & 0.564 & 0.537 & 0.545 & 0.560 & 0.479 & 0.513 & 0.546 & 0.526 & 0.436 & 0.564 & 0.510 & 0.531 & 0.514 & 0.501 & 0.486 & 0.503 & 0.492 & 0.504 & 0.435 & 0.556 & 0.521 \\
 &  &  & mIoU & 0.402 & 0.424 & 0.339 & 0.381 & 0.388 & 0.387 & 0.289 & 0.419 & 0.372 & 0.388 & 0.355 & 0.354 & 0.367 & 0.361 & 0.365 & 0.378 & 0.284 & 0.412 & 0.384 & 0.394 & 0.408 & 0.332 & 0.364 & 0.394 & 0.377 & 0.300 & 0.412 & 0.361 & 0.380 & 0.364 & 0.354 & 0.340 & 0.354 & 0.347 & 0.358 & 0.299 & 0.404 & 0.371 \\
 &  & right posterior superior segment of liver & Dice & 0.761 & 0.768 & 0.742 & 0.746 & 0.756 & 0.750 & 0.720 & 0.776 & 0.757 & 0.759 & 0.746 & 0.742 & 0.738 & 0.757 & 0.743 & 0.748 & 0.720 & 0.766 & 0.749 & 0.741 & 0.749 & 0.712 & 0.729 & 0.738 & 0.726 & 0.673 & 0.756 & 0.729 & 0.733 & 0.716 & 0.697 & 0.717 & 0.723 & 0.711 & 0.713 & 0.679 & 0.746 & 0.721 \\
 &  &  & mIoU & 0.627 & 0.636 & 0.605 & 0.607 & 0.621 & 0.614 & 0.578 & 0.645 & 0.621 & 0.625 & 0.608 & 0.604 & 0.598 & 0.622 & 0.604 & 0.612 & 0.578 & 0.632 & 0.612 & 0.598 & 0.608 & 0.564 & 0.584 & 0.595 & 0.580 & 0.518 & 0.616 & 0.583 & 0.589 & 0.568 & 0.547 & 0.569 & 0.577 & 0.563 & 0.565 & 0.526 & 0.604 & 0.574 \\
 &  & right posterior inferior segment of liver & Dice & 0.773 & 0.782 & 0.745 & 0.757 & 0.759 & 0.760 & 0.723 & 0.780 & 0.758 & 0.762 & 0.751 & 0.756 & 0.746 & 0.752 & 0.752 & 0.761 & 0.733 & 0.768 & 0.752 & 0.736 & 0.746 & 0.702 & 0.722 & 0.729 & 0.725 & 0.683 & 0.750 & 0.717 & 0.729 & 0.719 & 0.701 & 0.700 & 0.709 & 0.704 & 0.708 & 0.690 & 0.739 & 0.710 \\
 &  &  & mIoU & 0.644 & 0.656 & 0.610 & 0.624 & 0.629 & 0.628 & 0.584 & 0.654 & 0.626 & 0.632 & 0.619 & 0.623 & 0.611 & 0.619 & 0.619 & 0.630 & 0.596 & 0.640 & 0.619 & 0.597 & 0.609 & 0.557 & 0.580 & 0.590 & 0.584 & 0.537 & 0.615 & 0.576 & 0.589 & 0.578 & 0.557 & 0.555 & 0.566 & 0.560 & 0.563 & 0.543 & 0.601 & 0.568 \\
\bottomrule
\end{tabular}%
}
\end{table*}



\end{document}